\documentclass[twoside,11pt]{article}

 \usepackage[abbrvbib, preprint]{jmlr2e}

\usepackage{lastpage}
\jmlrheading{23}{2022}{1-\pageref{LastPage}}{1/21; Revised
	5/22}{8/22}{21-0028}{Zhize Li and Jian Li}

\ShortHeadings{Nonsmooth Nonconvex Optimization and Finding Local Minima}{Li and Li}
\firstpageno{1}

\usepackage{tabularx}
\usepackage{booktabs}
\usepackage{setspace}
\usepackage{multirow}
\usepackage{makecell}
\usepackage{graphicx}
\usepackage{delarray}
\usepackage{xcolor}
\usepackage{colortbl}
\usepackage{amssymb}
\usepackage{amsmath}
\usepackage{mathrsfs}
\usepackage{mathtools}
\usepackage{nicefrac}       
\usepackage{tikz}
\usepackage{xspace}
\xspaceaddexceptions{]\}}
\usepackage[linesnumbered,boxed,ruled]{algorithm2e}  
\usepackage{algorithmic}
\usepackage{mdframed}
\usepackage{bm}
\usepackage{tablefootnote}
\usepackage{afterpage}

\usepackage{times}
\usepackage{enumerate}
\usepackage{fmtcount}
\usepackage{thmtools}
\usepackage{thm-restate}

\allowdisplaybreaks

\newcommand{\cD}{{\mathcal D}}

\newcommand{\calG}{{\mathcal G}}

\newtheorem{assumption}{Assumption}
{\hspace*{\fill}$\Box$\par\vspace{4mm}}
\newenvironment{proofof}[1]{\smallskip\noindent{\bf Proof of #1.}}%
{\hspace*{\fill}$\Box$\par}

\let\oldnl\nl 
\newcommand{\nonl}{\renewcommand{\nl}{\let\nl\oldnl}}
\usetikzlibrary{fit,calc}
\newcommand*{\tikzmk}[1]{\tikz[remember picture,overlay,] \node (#1) {};\ignorespaces}
\newcommand{\boxalg}[1]{\tikz[remember picture,overlay]{\node[yshift=3pt,fill=#1,opacity=.25,fit={(A)($(B)+(.9\linewidth,.8\baselineskip)$)}] {};}\ignorespaces}
\newcommand{\boxalgone}[1]{\tikz[remember picture,overlay]{\node[yshift=3pt,fill=#1,opacity=.25,fit={(A)($(B)+(.345\linewidth,.2\baselineskip)$)}] {};}\ignorespaces}
\colorlet{algbgcolor}{gray!80}

\newcommand{\topic}[1]{\vspace{2mm}\noindent{{\bf #1:}}}
\newcommand{\red}[1]{\textcolor{red}{#1}}
\newcommand{\blue}[1]{\textcolor{blue}{#1}}
\definecolor{bgcolor}{rgb}{0.66,0.88,1.00}
\definecolor{bgcolor2}{rgb}{0.66,0.88,0.50}

\newcommand{\E}{{\mathbb{E}}}
\newcommand{\R}{{\mathbb R}}
\newcommand{\prox}{\mathrm{prox}}
\newcommand{\inner}[2]{\langle #1,#2 \rangle}

\newcommand{\ns}[1]{\| #1 \|^2}
\newcommand{\n}[1]{\| #1 \|}

\newcommand{\hx}{\widehat{x}}

\newcommand{\tx}{\widetilde{x}}

\newcommand{\bx}{\bar{x}}

\newcommand{\tdo}{\widetilde{O}}
\newcommand{\hess}{\mathcal{H}}

\newcommand{\mathG}{\epsilon}
\newcommand{\mathf}{f_{\mathrm{thres}}}
\newcommand{\mathT}{t_{\mathrm{thres}}}

\newcommand{\base}{{1+\eta\gamma}}

\newcommand{\Dtop}{{\frac{\delta}{C_1\rho}+r}}
\newcommand{\spe}{{super\_epoch}}
\newcommand{\pr}{{\mathbb P}}
\newcommand{\proj}{\bm{\text{Proj}}}

\newcommand{\eat}[1]{}
\usepackage{todonotes}

\begin{document}

\title{Simple and Optimal Stochastic Gradient Methods for Nonsmooth Nonconvex Optimization\thanks{Some preliminary results of this paper appear in two conference papers NeurIPS'18~\citep{li2018simple} and NeurIPS'19~\citep{li2019ssrgd}. 
This paper further simplifies some of the proofs, improves
the bounds and extends various results to more general settings. The detailed differences between the present paper
and the preliminary conference papers are summarized in Section~\ref{subsec:comparison}.}
}

\author{\name Zhize Li \email zhizeli@cmu.edu \\
	\addr Department of Electrical and Computer Engineering\\
	Carnegie Mellon University\\
	Pittsburgh, PA 15213, USA
	\AND
	\name Jian Li \email lijian83@mail.tsinghua.edu.cn \\
	\addr Institute for Interdisciplinary Information Sciences\\
	Tsinghua University\\
	Beijing 100084, China}

\maketitle

\begin{abstract}
We propose and analyze several stochastic gradient algorithms for 
finding stationary points or local minimum in nonconvex, possibly with nonsmooth regularizer, finite-sum and online optimization problems.
First, we propose a simple proximal stochastic gradient algorithm based on variance reduction called ProxSVRG+.  
We provide a clean and tight analysis of ProxSVRG+, which shows that it outperforms the deterministic proximal gradient descent (ProxGD) for a wide range of minibatch sizes, hence solves an open problem proposed in~\citet{reddi2016proximal}. Also, ProxSVRG+ uses much less proximal oracle calls than ProxSVRG~\citep{reddi2016proximal} and extends to the online setting by avoiding full gradient computations.
Then, we further propose an optimal algorithm, called SSRGD, based on SARAH~\citep{nguyen2017sarah} and show that
SSRGD further improves the gradient complexity of ProxSVRG+ and achieves the optimal upper bound, matching the known lower bound of \citep{fang2018spider,li2021page}.
Moreover, we show that both ProxSVRG+ and SSRGD enjoy automatic adaptation with local structure of the objective function such as the Polyak-\L{}ojasiewicz (PL) condition for nonconvex functions in the finite-sum case, i.e., we prove that both of them can automatically switch to faster global linear convergence without any restart performed in prior work ProxSVRG~\citep{reddi2016proximal}.
Finally, we focus on the more challenging problem of finding an $(\epsilon, \delta)$-local minimum
instead of just finding an $\epsilon$-approximate (first-order) stationary point 
(which may be some bad unstable saddle points).
We show that SSRGD can find an $(\epsilon, \delta)$-local minimum 
by simply adding some random perturbations. 
Our algorithm is almost as simple as its counterpart for finding stationary points, and achieves similar
optimal rates.
\end{abstract}


\section{Introduction}
\label{sec:intro}

Nonconvex optimization is ubiquitous in machine learning problems,
especially in training deep neural networks.
In this paper, we consider the nonsmooth (composite) nonconvex optimization problems of the form
\begin{equation}\label{eq:problem}
\min_{x\in \R^d} \left\{     \Phi(x) := f(x)+h(x)   \right\},
\end{equation}
where $f: \R^d\to \R$ is a differentiable and possibly nonconvex function, while $h: \R^d\to \R$ is nonsmooth but convex (e.g., $\ell_1$ norm $\| x \|_1$ or indicator function $I_C(x)$ for some convex set $C$).
In particular, we are interested in functions of $f$ having the \emph{finite-sum} form
\begin{align}\label{prob:finite}
f(x) := \frac{1}{n}\sum_{i=1}^n{f_i(x)},
\end{align}
where functions $f_i$s are also possibly nonconvex.
The finite-sum form captures the standard empirical risk minimization problems and thus is fundamental to many machine learning problems, ranging from
convex optimization ($f_i$s are convex functions) such as logistic regression, SVM to highly nonconvex problem such as optimizing deep neural networks.
Moreover, if the number of data samples $n$ is very large or even infinite, e.g., in the online/streaming case, then function $f$ usually is modeled via the \emph{online} form
\begin{align}\label{prob:online}
f(x) := \E_{\zeta\sim \cD}[F(x,\zeta)].
\end{align}
For notational convenience, we adopt the notation of the finite-sum form \eqref{prob:finite} in the descriptions and algorithms in the rest of this paper. However, our results apply to the online form \eqref{prob:online} as well by letting $f_i(x) := F(x, \zeta_i)$ and treating $n$ as a very large number or even infinite.

There is a large body of literature for solving the standard problem \eqref{eq:problem} with finite-sum form \eqref{prob:finite} or online form \eqref{prob:online}.
The convex setting (i.e., $f_i$s are convex) are well-understood 
(see e.g., \citealp{xiao2014proximal,lin2015universal,lan2015optimal,woodworth2016tight,lan2018random,allen2017katyusha,zhize2019unified,li2021anita}). 
Due to the increasing popularity of deep learning, the nonconvex case has attracted significant attention in recent years. 
In search of the optimal algorithms, 
a large family of \emph{variance-reduced} methods plays an important role.
In particular, SVRG~\citep{johnson2013accelerating}, SAGA~\citep{schmidt2013minimizing,defazio2014saga} and SARAH~\citep{nguyen2017sarah} are representative variance-reduced methods which were originally designed to solve convex
optimization problems. They were extended to solve nonconvex problems in subsequent works, such as
SCSG~\citep{lei2017less,lei2017non}, SVRG+~\citep{li2018simple}, L-SVRG~\citep{kovalev2020don}, SNVRG~\citep{zhou2018stochastic}, SPIDER~\citep{fang2018spider}, SpiderBoost~\citep{wang2019spiderboost}, SSRGD~\citep{li2019ssrgd}, PAGE~\citep{li2021page}.
Particularly, \cite{li2020unified} provided a unified analysis for a large family of stochastic gradient methods in nonconvex optimization such as SGD, SGD with arbitrary sampling, SGD with compressed gradient, variance-reduced methods such as SVRG and SAGA, and their distributed variants. 
There are also many advanced variants in the distributed/federated settings such as \citep{karimireddy2020scaffold, li2020acceleration, zhao2021fedpage, li2021canita, gorbunov2021marina, richtarik2021ef21, fatkhullin2021ef21, li2021zerosarah, zhao2021faster, richtarik20223pc, li2022destress, zhao2022beer, li2022soteriafl}.

While much prior work focused on the smooth convex/nonconvex case (i.e., $h(x)\equiv 0$ in \eqref{eq:problem}),
relatively less work studied the more general \emph{nonsmooth} nonconvex case. Here we briefly survey previous work
that are directly related to our work.
\cite{ghadimi2016mini} analyzed the deterministic proximal gradient method (i.e., computing the full-gradient in every iteration) for this nonsmooth nonconvex setting. Here we denote it as ProxGD.
\cite{ghadimi2016mini} also considered the stochastic variant (here we denote it as ProxSGD). However, ProxSGD requires the minibatch size being a very large number (i.e., $b=O(\sigma^2/\epsilon^2)$) for showing the convergence.
Later, \cite{reddi2016proximal} provided two algorithms called ProxSVRG and ProxSAGA, which are based on 
SVRG ~\citep{johnson2013accelerating} and SAGA~\citep{defazio2014saga}.
However, their convergence results (using constant or moderate size minibatches) are still worse than the deterministic ProxGD in terms of {\em proximal oracle complexity} (see Definition~\ref{def:oracle} for the formal definition).
Note that their algorithms (i.e., ProxSVRG/SAGA) outperform the ProxGD only if they use a quite large minibatch size $b = O(n^{2/3})$, where $n$ is the number of training samples.
Note that from the perspectives of both computational efficiency and statistical generalization,
always computing full-gradient (GD or ProxGD) may not be desirable for large-scale machine learning problems. A
reasonable minibatch size is desirable in practice, since the computation of minibatch stochastic gradients can be much cheaper and also
implemented in parallel. In fact, practitioners typically use moderate minibatch sizes, often ranging from something
like 16 or 32 to a few hundreds (sometimes to a few thousands). Hence, it is important
to study the convergence in moderate and constant minibatch size regime.
In light of this consideration, \cite{reddi2016proximal} presented an important open problem of
\emph{developing stochastic methods with provably better performance than ProxGD with constant minibatch size}. 
In this paper, we provide algorithms for solving this open problem and also achieving optimal results.
See Table~\ref{tab:1} and \ref{tab:3} for more related works and their detailed convergence results.

Besides, we show that better convergence can be
achieved if the objective/loss function satisfies the Polyak-\L ojasiewicz (PL) condition (Assumption \ref{asp:pl}). 
Note that under the PL condition, one can obtain a faster linear convergence $O(\cdot \log \frac{1}{\epsilon})$ rather than the sublinear convergence $O(\cdot\frac{1}{\epsilon^2})$. 
In many cases, although the objective function is globally nonconvex, some local regions (e.g., large gradient regions) may satisfy the PL condition. 
Thus,  we also prove that our algorithms can achieve faster linear convergence rates under the PL condition.
In particular, the parameter settings of our algorithm remain the same for the finite-sum case, i.e., our algorithms can automatically switch to the faster linear convergence rate in these regions where the PL condition is satisfied.

For nonconvex problems, the point with zero gradient $\nabla f(x)=0$ can be a local minimum, a local maximum or a saddle point.
To avoid stucking in bad saddle points (or local maxima), we want to further find a local minimum, i.e., $\nabla f(x)=0$ and $\nabla^2 f(x) \succ 0$ (this is a sufficient condition for $x$ being a local minimum).
We note that although finding the global minimum for nonconvex problems is NP-hard in general, it is known that  for some special nonconvex problems all local minima are also global minima, such as matrix sensing \citep{bhojanapalli2016global}, matrix completion \citep{ge2016matrix}, and some special neural networks \citep{ge2017learning}.
In our paper, we also consider the goal of finding an approximate $(\epsilon,\delta)$-local minimum (i.e., $\|\nabla f(x)\|\leq \epsilon$ and $\lambda_{\min}(\nabla^2 f(x))\geq -\delta$) instead of just finding the $\epsilon$-approximate first-order solution (i.e., $\n{\nabla f(x)} \leq \epsilon$).
For this purpose, \cite{xu2018first} and \cite{allen2018neon2} independently proposed generic reductions Neon/Neon2, that can be combined with algorithms that finds $\epsilon$-approximate (first-order) solutions 
in order to find an $(\epsilon,\delta)$-local minimum. 
However, algorithms obtained via such reduction are quite complicated and rarely used in practice. In particular, the reduction needs to extract negative curvature directions from the Hessian to escape saddle points by using a negative curvature search subroutine: given a point $x$, find an approximate eigenvector corresponding to the  smallest eigenvalue of $\nabla^2 f(x)$. This also makes the analysis more complicated.
In practice, standard stochastic gradient algorithms can often work well in nonconvex setting (they can escape bad saddle points) without a negative curvature search subroutine.
Intuitively, the saddle points are not very stable, and some stochasticity can escape such saddle points.
This raises the following natural theoretical question ``Is there any simple modification to the standard first-order gradient method, that can achieve second-order optimality guarantee (local minimum)?".
For gradient descent (GD), \cite{jin2017escape} showed that a simple perturbation step (by injecting Gaussian noises) is enough to escape saddle points for finding an $(\epsilon,\delta)$-local minimum, and this is necessary \citep{du2017gradient}.
\cite{jin2018accelerated} showed that an accelerated GD version can achieve faster convergence rate.  Note that both \citep{jin2017escape, jin2018accelerated} require computing the full gradients. \cite{ge2015escaping}, \cite{daneshmand2018escaping},
\cite{jin2019stochastic}, and \cite{fang2019sharp} analyzed stochastic gradients can also find approximate local minimum if some Gaussian noises are injected.\footnote{\cite{daneshmand2018escaping} and \cite{fang2019sharp} also show that the plain SGD can find approximate local minimum under certain assumptions.}
Recently, \cite{ge2019stable} showed that a simple perturbation step is also enough to find an $(\epsilon,\delta)$-local minimum for SVRG algorithm \citep{johnson2013accelerating,li2018simple}. Moreover, \cite{ge2019stable} also developed a stabilized trick to further improve the dependency of Hessian Lipschitz parameter.
See also Table~\ref{tab:localmin-finite} for the convergence rates of the aforementioned works.

In the next section, we present our contributions and provide more discussions and details of related work.

\section{Our Contributions}
\label{sec:contribution}
In this section, we review previous related work and present our contributions.
Concretely, in Section~\ref{sec:con-standard}, we compare the convergence results of our ProxSVRG+ and SSRGD with previous work, and show that SSRGD further improves on ProxSVRG+ and achieves the optimal convergence results.
In Section~\ref{sec:con-pl}, we present the convergence results of ProxSVRG+ and SSRGD under the PL condition. In this PL setting, SSRGD achieves new state-of-the-art results. Note that both ProxSVRG+ and SSRGD can automatically switch to the faster global linear convergence in the finite-sum case under the PL condition.
In Section~\ref{sec:con-localmin}, we further present the convergence results of SSRGD for finding an \emph{approximate local minimum} which is a more challenging guarantee compared with just finding an approximate first-order stationary point.

\subsection{Nonsmooth nonconvex optimization}
\label{sec:con-standard}

We list the convergence results of ProxGD, ProxSGD, ProxSVRG/SAGA, ProxSVRG+ and SSRGD in Table~\ref{tab:1}.
Our goal in this section is to find an $\epsilon$-approximate solution of \eqref{eq:problem} (see Definition~\ref{def:eps}).
The convergence results are stated in terms of the number of stochastic first-order oracle (SFO) calls and proximal oracle (PO) calls (see Definition~\ref{def:oracle}).
Although the algorithm of ProxSVRG+ is the same as in the conference version~\citep{li2018simple}, our convergence analysis is notably different.
In this paper, we further simplify our original proofs of ProxSVRG+ provided in \citet{li2018simple}, which 
allows for using larger step size and also leads to better constant in the convergence results.

The original version of SSRGD (Simple Stochastic Recursive Gradient Descent) in \citet{li2019ssrgd} was designed to solve the smooth nonconvex problems (i.e., $h(x)\equiv 0$ in \eqref{eq:problem}).
In this paper, we extend it to solve the \emph{nonsmooth} nonconvex problems \eqref{eq:problem}.
Compared with the SFO complexity of ProxSVRG+, SSRGD improves the factor $\sqrt{b}$ to $b$, e.g., $O(\frac{n}{\epsilon^2\sqrt{b}}+\frac{b}{\epsilon^2})$ to $O(\frac{n}{\epsilon^2 b}+\frac{b}{\epsilon^2})$ in the finite-sum case, where $b$ is the minibatch size (See Table \ref{tab:1}).
Although SSRGD yields better convergence results than ProxSVRG+,
in our opinion, the analysis of ProxSVRG+ is quite simple and clean, and useful for understanding the analysis of SSRGD. Hence, we choose to present the details of ProxSVRG+ as well.


\begin{figure}[!t]
	\centering
	\begin{minipage}[htb]{0.47\textwidth}
		\centering
		\includegraphics[width=0.95\textwidth]{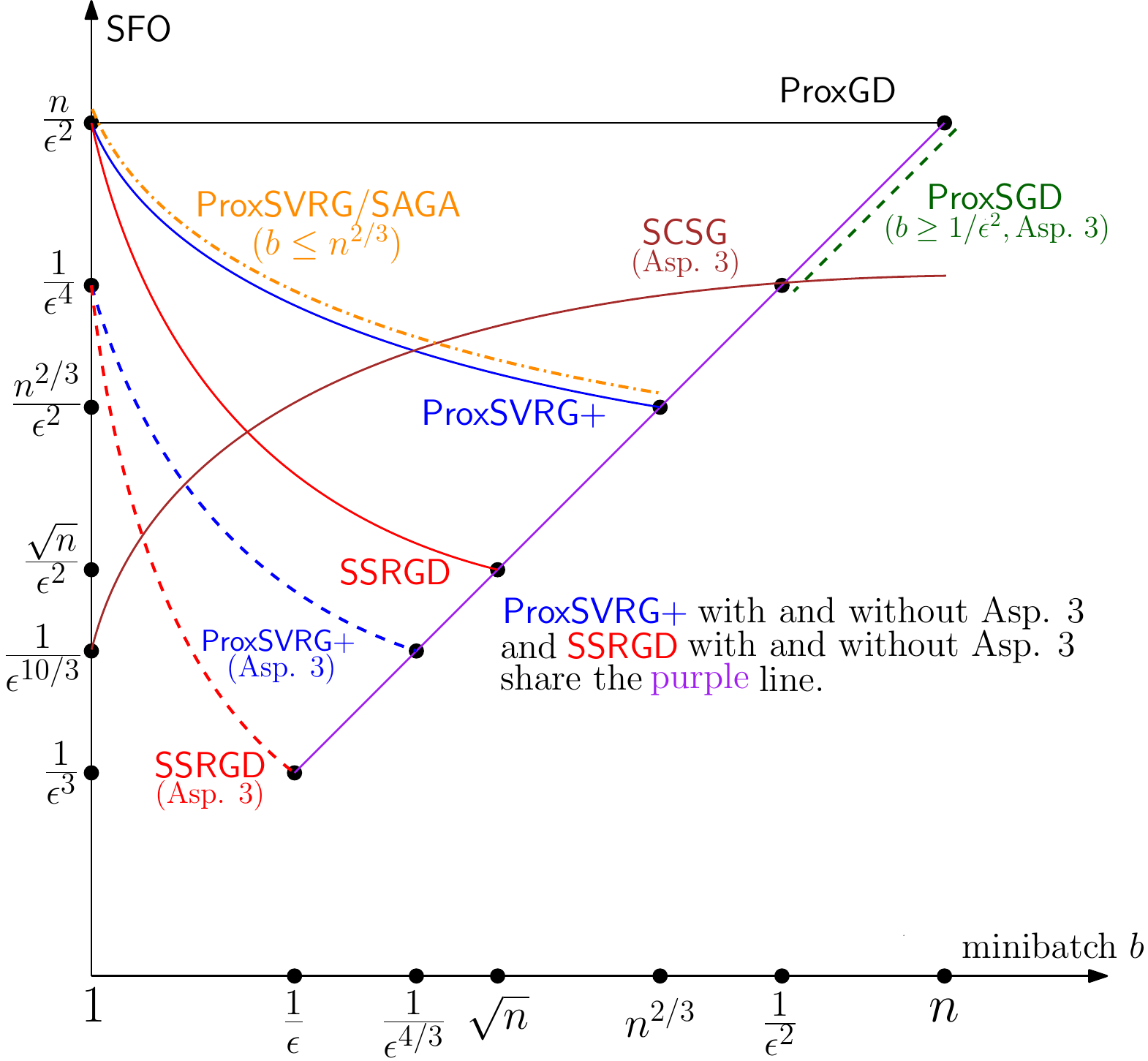}
		\caption[fig1]{\small SFO complexity w.r.t. minibatch $b$. \footnotemark
		}
		\label{fig:1}
	\end{minipage}\hspace{5pt}
	\begin{minipage}[htb]{0.47\textwidth}
		\centering
		\includegraphics[width=0.95\textwidth]{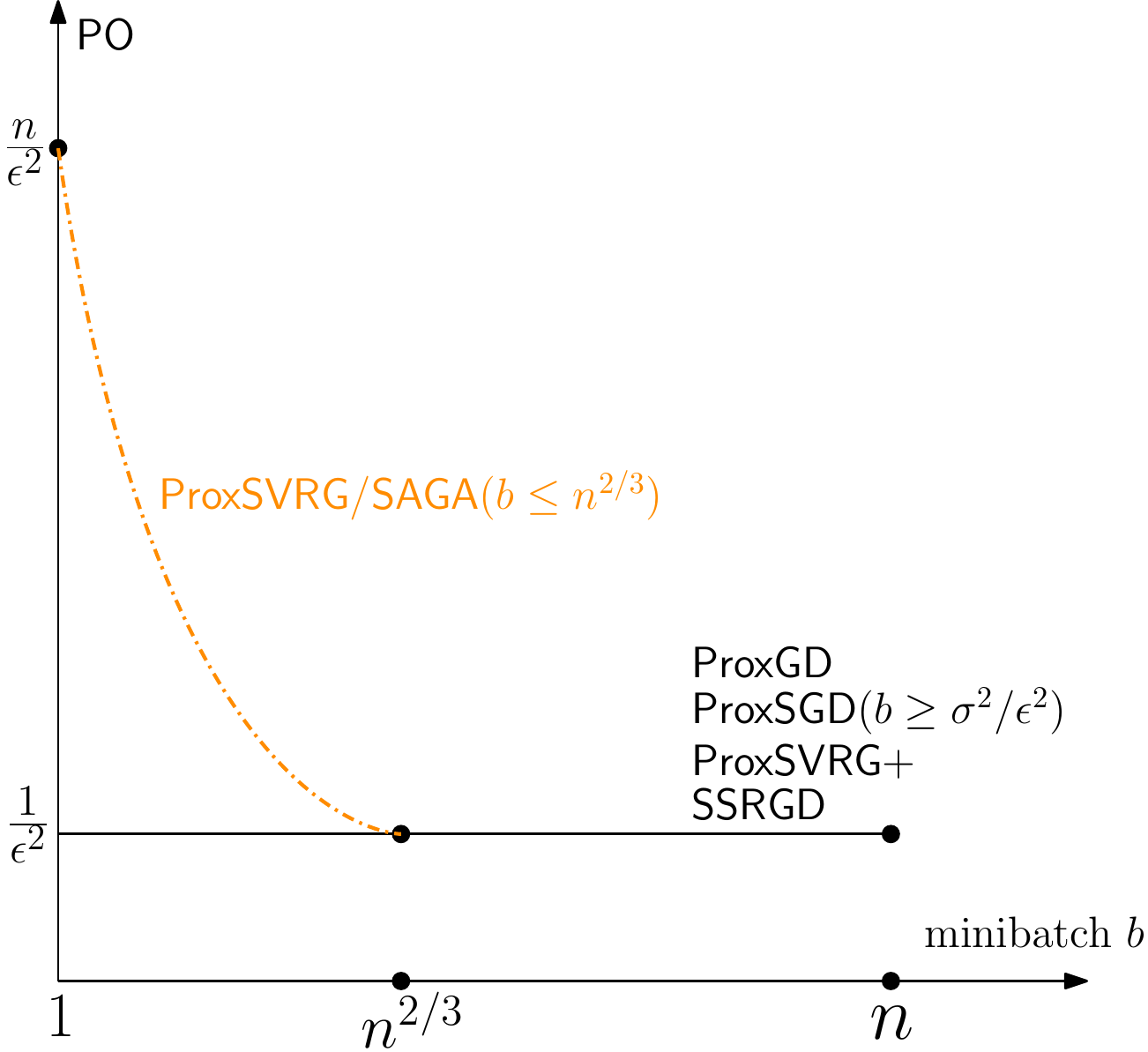}
		\caption{\small PO complexity w.r.t. minibatch $b$.
			\newline
		}
		\label{fig:2}
	\end{minipage}
\end{figure}
\footnotetext{In this figure, we assume that %
	$\sigma^2/\epsilon^2\leq n$, i.e., $B:=\min\{n,\sigma^2/\epsilon^2\}=\sigma^2/\epsilon^2$.
	Otherwise there is no difference from the finite-sum case if $B=n$.
	We also omit $\sigma$ for simplicity of presentation.}

\begin{table}[!t]
	\vspace{-3mm}
	\centering
	\caption{SFO and PO complexity for finding an $\epsilon$-approximate solution of problem (\ref{eq:problem})}
	\label{tab:1}
	\small
	\vspace{1mm}
	\begin{tabular}{|c|c|c|c|}
		\hline
		Algorithms & \makecell{Stochastic first-order \\ oracle (SFO)} & \makecell{Proximal oracle \\ (PO)}
		& Assumptions\\ \hline
		
		\makecell{ProxGD\\ \citep{ghadimi2016mini}}
		& $O(\frac{n}{\epsilon^2})$  & $O(\frac{1}{\epsilon^2})$
		& \makecell{Asp \ref{asp:smooth} \\ (finite-sum)} \\\hline
		
		\makecell{ProxSGD\\ \citep{ghadimi2016mini}}
		& \makecell{$O(\frac{b}{\epsilon^2}),$ where $b\geq \frac{\sigma^2}{\epsilon^2}$}  & $O(\frac{1}{\epsilon^2})$
		& \makecell{Asp \ref{asp:avgsmooth}, \ref{asp:bv} \\ (finite-sum or online 
			\tablefootnote{Note that we refer to the finite-sum problem \eqref{prob:finite} with large or infinite $n$ as the online problem \eqref{prob:online}, as discussed in Section \ref{sec:intro}. 
				In the online problem, computing the full gradient may be very expensive or simply impossible (e.g., if $n$ is infinite), so the bounded variance assumption of stochastic gradient (Assumption \ref{asp:bv}) 
				is needed.\label{tabfn:online}})} \\\hline

		\makecell{ProxSVRG/SAGA\\ \citep{reddi2016proximal}}
		&\makecell{$O\big(\frac{n}{\epsilon^2\sqrt{b}}+n\big),$\\ where $b\le n^{2/3}$}
		& $O\big(\frac{n}{b^{3/2}\epsilon^2}\big)$
		&  \makecell{Asp \ref{asp:avgsmooth} \\ (finite-sum)}  \\\hline
		
		\makecell{SCSG~\citep{lei2017non} \\
			(smooth case $h(x)\equiv 0$ in \eqref{eq:problem})}
		&\makecell{$O\big(\frac{b^{1/3}B^{2/3}}{\epsilon^2} +B \big)$ ~\tablefootnote{$B:=\min\{n,\sigma^2/\epsilon^2\}$.\label{tabfn:B}}} &NA \tablefootnote{SCSG~\citep{lei2017non} only considered the smooth case, i.e., $h(x)\equiv 0$ in problem \eqref{eq:problem}. The proximal oracle is required only for the nonsmooth setting.} & \makecell{Asp \ref{asp:avgsmooth}, \ref{asp:bv} \\ (finite-sum or online \textsuperscript{\getrefnumber{tabfn:online}})} \\\hline

		\blue{\makecell{ProxSVRG+ \\
				(this paper, Theorem~\ref{thm:1})}}
		& $O\big(\frac{n}{\epsilon^2\sqrt{b}}+\frac{b}{\epsilon^2} +n\big)$
		& $O(\frac{1}{\epsilon^2})$  &
		\makecell{Asp \ref{asp:avgsmooth} \\ (finite-sum)}
		\\ \hline
		\blue{\makecell{ProxSVRG+ \\
				(this paper, Theorem~\ref{thm:1})}}
		&$O\big(\frac{B}{\epsilon^2\sqrt{b}}
		+\frac{b}{\epsilon^2} +B\big)$  ~\textsuperscript{\getrefnumber{tabfn:B}}
		& $O(\frac{1}{\epsilon^2})$ &  \makecell{Asp \ref{asp:avgsmooth}, \ref{asp:bv} \\ (finite-sum or online \textsuperscript{\getrefnumber{tabfn:online}})} \\\hline
		
		\red{\makecell{
				SSRGD \\ (this paper, Theorem~\ref{thm:ssrgd})}}
		& $O\big(\frac{n}{\epsilon^2b}+\frac{b}{\epsilon^2}+n\big)$
		& $O(\frac{1}{\epsilon^2})$  &
		\makecell{Asp \ref{asp:avgsmooth} \\ (finite-sum)}
		\\ \hline
		\red{\makecell{SSRGD \\
				(this paper, Theorem~\ref{thm:ssrgd})}}
		&$O\big(\frac{B}{\epsilon^2b}
		+\frac{b}{\epsilon^2}+B\big)$  ~\textsuperscript{\getrefnumber{tabfn:B}}
		& $O(\frac{1}{\epsilon^2})$ &  \makecell{Asp \ref{asp:avgsmooth}, \ref{asp:bv} \\ (finite-sum or online
			\textsuperscript{\getrefnumber{tabfn:online}})} \\\hline

	\end{tabular}
			\vspace{-2mm}
\end{table}

We highlight the following results yielded by ProxSVRG+ and SSRGD:
\begin{enumerate}[1)]

	\item 	
	\cite{reddi2016proximal} proposed the following open question: developing stochastic methods with provably better performance than ProxGD with constant minibatch size $b$.    Note that \#PO of ProxSVRG~\citep{reddi2016proximal} is $n/b^{2/3}$ times larger than ProxGD.	
	Our ProxSVRG+ is $\sqrt{b}$ (resp. $\sqrt{b}n/B$ in the online case) times faster than
	ProxGD in terms of \#SFO when $b\leq n^{2/3}$
	(resp. $b\leq B^{2/3}$),
	and $n/b$ times faster than	ProxGD when $b>n^{2/3}$ (resp. $b>B^{2/3}$), where $B:=\min\{n,\frac{\sigma^2}{\epsilon^2}\}$.
	SSRGD is $b$ (resp. $bn/B$ in the online case) times faster than
	ProxGD in terms of \#SFO when $b\leq n^{1/2}$
	(resp. $b\leq B^{1/2}$),
	and $n/b$ times faster than	ProxGD when $b>n^{1/2}$ 	
	(resp. $b>B^{1/2}$). 
	Note that the number of proximal oracle (PO) calls for ProxGD, ProxSVRG+ and SSRGD are the same, i.e., \#PO $=O(1/\epsilon^2)$. 
	Hence, both results answers the open question posed by \citet{reddi2016proximal}.
	Also see Figure \ref{fig:1} and \ref{fig:2} for an overview.

	\begin{table}[t]
		\centering
		\caption{SFO and PO complexity of recent algorithms for solving problem (\ref{eq:problem}) \tablefootnote{Similar results hold for these recent algorithms and our SSRGD by replacing $n$ with $B:=\min\{n,\frac{\sigma^2}{\epsilon^2}\}$ in finite-sum or online setting (under Asp \ref{asp:avgsmooth} and \ref{asp:bv}) similar to Table \ref{tab:1}. Note that SSRGD also matches the lower bound $\Omega(B+\frac{\sqrt{B}}{\epsilon^2})$~\citep{li2021page} in the online setting (i.e., it achieves optimal results in both finite-sum and online settings).}}
		\label{tab:3}
		\small
		\vspace{1mm}
		\begin{tabular}{|c|c|c|c|}
			\hline
			Algorithms & \makecell{Stochastic first-order \\ oracle (SFO)} & \makecell{Proximal oracle \\ (PO)}   
			& Assumptions\\ \hline
			
			SNVRG~\citep{zhou2018stochastic}
			&  $\widetilde{O}(n+\frac{\sqrt{n}}{\epsilon^2})$ \tablefootnote{They only analyzed a fixed choice of minibatch size $b$.\label{tabfn:fixb}}
			& NA &\makecell{Asp \ref{asp:avgsmooth} \\ (finite-sum)}
			\\ \hline
			
			SPIDER~\citep{fang2018spider}
			&  $O(n+\frac{\sqrt{n}}{\epsilon^2})$
			\textsuperscript{\getrefnumber{tabfn:fixb}}
			& NA &\makecell{Asp \ref{asp:avgsmooth} \\ (finite-sum)}
			\\ \hline
			
			SpiderBoost~\citep{wang2019spiderboost}
			&  $O(n+\frac{\sqrt{n}}{\epsilon^2})$
			\textsuperscript{\getrefnumber{tabfn:fixb}}
			& $O(\frac{1}{\epsilon^2})$ &\makecell{Asp \ref{asp:avgsmooth} \\ (finite-sum)}
			\\ \hline
			
			ProxSARAH~\citep{pham2019proxsarah}
			& $O(n+\frac{\sqrt{n}}{\epsilon^2})$
			& \makecell{$O(\frac{\sqrt{n}}{b\epsilon^2}),$\\ where $b\le \sqrt{n}$} &\makecell{Asp \ref{asp:avgsmooth} \\ (finite-sum)}
			\\ \hline
			
			PAGE~\citep{li2021page}
			& $O(n+\frac{\sqrt{n}+b}{\epsilon^2})$
			& NA &\makecell{Asp \ref{asp:avgsmooth} \\ (finite-sum)}
			\\ \hline
			
			\red{\makecell{SSRGD
					(this paper, Theorem~\ref{thm:ssrgd})}}
			& \makecell{$O(n+\frac{\sqrt{n}}{\epsilon^2})$}
			& $O(\frac{1}{\epsilon^2})$  &  
			\makecell{Asp \ref{asp:avgsmooth} \\ (finite-sum)}
			\\ \hline
			\makecell{Lower bound~\citep{fang2018spider}}
			&\makecell{$\Omega(\frac{\sqrt{n}}{\epsilon^2}),$\\ where $n\leq O(\frac{1}{\epsilon^4})$}
			& NA &\makecell{Asp \ref{asp:avgsmooth} \\ (finite-sum)}
			\\ \hline
			\makecell{Lower bound \citep{li2021page}}
			&$\Omega(n+\frac{\sqrt{n}}{\epsilon^2})$
			& NA &\makecell{Asp \ref{asp:avgsmooth} \\ (finite-sum)}
			\\ \hline
		\end{tabular}
	\end{table}

	\item		
	For the online case (which needs an extra bounded variance Assumption \ref{asp:bv} since the full gradient may not be available),
	ProxSVRG+ and SSRGD generalize and improve the result achieved by SCSG~\citep{lei2017non} for the smooth nonconvex case ($h(x)\equiv 0$ in form (\ref{eq:problem})) to this nonsmooth setting.
	We note that ProxSVRG+ is also more straightforward than SCSG and the proof is also simpler.
	Also note that SCSG \citep{lei2017non} achieves its best convergence result with minibatch size $b=1$ (see Figure \ref{fig:1}),
	while ProxSVRG+ and SSRGD achieves their best results with moderate minibatch sizes and thus can also enjoy speed up with parallelism/vectorization.

	\item By choosing the best minibatch size $b=\sqrt{n}$ ($b=\sqrt{B}$ in online case), SSRGD achieves the 
	{\em optimal} results in both finite-sum and online settings, which match the lower bounds given in \citep{fang2018spider,li2021page}.
	Note that the optimal SFO complexity
	$O(n+\frac{\sqrt{n}}{\epsilon^2})$
	have already been achieved by several recent works such as SNVRG \citep{zhou2018finding}, 
	SPIDER \citep{fang2018spider}, SpiderBoost \citep{wang2019spiderboost}, ProxSARAH \citep{pham2019proxsarah}
	and PAGE \citep{li2021page}. 
	We highlight some differences between SSRGD and previous results. 
	For SNVRG, SPIDER and PAGE, they only considered the smooth case ($h(x)\equiv 0$ in form (\ref{eq:problem})). SpiderBoost only analyzed a fixed choice of minibtach size $b$ and ProxSARAH requires much more \#PO calls if minibatch size $b$ is small. SSRGD provides the results for all minibatch size $b\in[1,n]$ and the number of \#PO calls is always the same as ProxGD.
	We also note that ProxSARAH with $\gamma_t\equiv 1$ (Algorithm 1 in \citep{pham2019proxsarah}) is the same as SSRGD. 
	However, the convergence analysis in \citet{pham2019proxsarah} (Theorem 6 in their paper) requires $\gamma_t\equiv \gamma = \frac{1}{L\sqrt{\omega m}}$ (where $\omega = \frac{3(n-b)}{2b(n-1)}$), hence does not cover the case that $\gamma_t\equiv 1$.
	Our main technical contribution is a simple and clean analysis (arguably simpler than that in the previous optimal algorithms) that is inspired by our analysis of ProxSVRG+.
	See Table \ref{tab:3} for these recent results. Note that these results
	were not stated in terms of minibatch size $b$, so we use a separate table for them.
	
\end{enumerate}

\subsection{PL setting}
\label{sec:con-pl}

Note that under the PL condition (Assumption \ref{asp:pl}), one can obtain the faster linear convergence rates $O(\cdot\log \frac{1}{\epsilon})$ (see Theorem~\ref{thm:pl1} and \ref{thm:pl-ssrgd}) rather than the sublinear convergence rates $O(\cdot\frac{1}{\epsilon^2})$ (see Theorem~\ref{thm:1} and \ref{thm:ssrgd}).

\begin{table}[t]
	\centering
	\caption{SFO and PO complexity of algorithms under PL condition for solving problem (\ref{eq:problem})}
	\label{tab:1pl}
	\small
	\vspace{1mm}
	\begin{tabular}{|c|c|c|c|}
		\hline
		Algorithms & \makecell{Stochastic first-order \\ oracle (SFO)} & \makecell{Proximal oracle \\ (PO)}
		& Assumptions\\ \hline
		
		\makecell{ProxGD\\ 	\citep{karimi2016linear}}
		& $O(n\kappa\log\frac{1}{\epsilon})$
		& $O(\kappa\log\frac{1}{\epsilon})$
		& \makecell{Asp \ref{asp:smooth}, \ref{asp:pl} \\ (finite-sum)} \\\hline
		
		\makecell{ProxSVRG/SAGA\\ \citep{reddi2016proximal}}
		&\makecell{$O\big(\frac{n\kappa}{\sqrt{b}}\log\frac{1}{\epsilon}+n\log\frac{1}{\epsilon}\big),$\\ where $b\le n^{2/3}$}
		& $O\big(\frac{n\kappa}{b^{3/2}}\log\frac{1}{\epsilon}\big)$
		&  \makecell{Asp \ref{asp:avgsmooth}, \ref{asp:pl} \\ (finite-sum)}  \\\hline
		
		\makecell{SCSG~\citep{lei2017non} \\
			(smooth case $h(x)\equiv 0$ in \eqref{eq:problem})}
		&\makecell{$O\big(b^{1/3}B^{2/3}\kappa\log\frac{1}{\epsilon}\big)$ \tablefootnote{$B:=\min\{n,\frac{\sigma^2}{\mu\epsilon}\}$.\label{tabfn:Bpl}}} &NA \tablefootnote{SCSG~\citep{lei2017non} also only considered the smooth case (i.e., $h(x)\equiv 0$ in problem \eqref{eq:problem}) in the PL setting. The proximal oracle is only required for the nonsmooth setting.\label{tabfn:na}} & \makecell{Asp \ref{asp:avgsmooth}, \ref{asp:bv}, \ref{asp:pl} \\ (finite-sum or online)} \\\hline

		\blue{\makecell{ProxSVRG+ \\
				(this paper, Theorem~\ref{thm:pl1})}}
		&$O\big(\frac{n\kappa}{\sqrt{b}}\log\frac{1}{\epsilon}+b\kappa\log\frac{1}{\epsilon}\big)$
		&  $O(\kappa\log\frac{1}{\epsilon})$
		& \makecell{Asp \ref{asp:avgsmooth}, \ref{asp:pl} \\ (finite-sum)}
		\\ \hline
		\blue{\makecell{ProxSVRG+ \\
				(this paper, Theorem~\ref{thm:pl1})}}
		&$O\big(\frac{B\kappa}{\sqrt{b}}\log\frac{1}{\epsilon}+b\kappa\log\frac{1}{\epsilon}\big)$   \textsuperscript{\getrefnumber{tabfn:Bpl}}
		& $O(\kappa\log\frac{1}{\epsilon})$ &  \makecell{Asp \ref{asp:avgsmooth}, \ref{asp:bv}, \ref{asp:pl} \\ (finite-sum or online)} \\\hline
		
		\red{\makecell{	SSRGD \\
				(this paper, Theorem~\ref{thm:pl-ssrgd})}}
		& $O\big(\frac{n\kappa}{b}\log\frac{1}{\epsilon}+b\kappa\log\frac{1}{\epsilon}\big)$
		& $O(\kappa\log\frac{1}{\epsilon})$  &
		\makecell{Asp \ref{asp:avgsmooth}, \ref{asp:pl} \\ (finite-sum)}
		\\ \hline
		\red{\makecell{SSRGD\\
				(this paper, Theorem~\ref{thm:pl-ssrgd})}}
		&$O\big(\frac{B\kappa}{b}\log\frac{1}{\epsilon}+b\kappa\log\frac{1}{\epsilon}\big)$ \textsuperscript{\getrefnumber{tabfn:Bpl}}
		& $O(\kappa\log\frac{1}{\epsilon})$ &  \makecell{Asp \ref{asp:avgsmooth}, \ref{asp:bv}, \ref{asp:pl} \\ (finite-sum or online)} \\\hline
		
	\end{tabular}
\end{table}

Now we summarize the convergence results of prior work, ProxSVRG+ and SSRGD under PL condition (Assumption \ref{asp:pl}) in Table~\ref{tab:1pl}.
The convergence result of these algorithms are very similar to Table~\ref{tab:1} by replacing $\frac{1}{\epsilon^2}$ with $\kappa\log\frac{1}{\epsilon}$.
Similarly, under PL condition, ProxSVRG+ also improves ProxSVRG by using less PO calls and extends the choice of minibatch size to all $b\in[1,n]$.
SSRGD further improves ProxSVRG+ by a factor of $\sqrt{b}$, e.g., from $O(\frac{n\kappa}{\sqrt{b}}\log\frac{1}{\epsilon}+b\kappa\log\frac{1}{\epsilon})$ to $O(\frac{n\kappa}{b}\log\frac{1}{\epsilon}+b\kappa\log\frac{1}{\epsilon})$ in the finite-sum case (See Table \ref{tab:1pl}).
In particular, the best result for SSRGD is $\tdo(\sqrt{n}\kappa)$ while the best results for ProxSVRG and ProxSVRG+ are $\tdo(n^{2/3}\kappa)$.
For the online case, the best result for SSRGD is  $\tdo(\sqrt{B}\kappa)$ while the best results for SCSG and ProxSVRG+ are $\tdo(B^{2/3}\kappa)$, where $B:=\min\{n,\frac{\sigma^2}{\mu\epsilon}\}$.
See Table \ref{tab:1pl} for more details.

By choosing the best minibatch size, SSRGD achieves new state-of-the-art results in the PL setting. 
See Table~\ref{tab:3pl} for convergence results of SSRGD (with best minibatch $b$) and some prior results. 
Note that we are mainly interested in the case where the condition number $\kappa> \sqrt{n}$. Hence, one can see
that SSRGD is better than Prox-SpiderBoost-PL \citep{wang2019spiderboost} in term of the number of SFO calls.
If the condition number $\kappa\leq \sqrt{n}$,  the SFO complexity of both Prox-SpiderBoost-PL and SSRGD can be bounded by $O(n\log\frac{1}{\epsilon})$.

Note that we do not combine Table \ref{tab:1pl} and \ref{tab:3pl}
since all prior results in Table~\ref{tab:3pl} were not stated in terms of the minibatch size $b$. Hence, we use a separate table to list the best results they achieved.
We emphasize that our analysis of SSRGD in this PL setting is new and its convergence result also improves over all prior results (see Table \ref{tab:1pl} and \ref{tab:3pl}).

\begin{table}[t]
	\centering
	\caption{SFO and PO complexity of recent algorithms under PL condition for solving problem (\ref{eq:problem})}
	\label{tab:3pl}
	\small
	\vspace{1mm}
	\begin{tabular}{|c|c|c|c|}
		\hline
		Algorithms & \makecell{Stochastic first-order \\ oracle (SFO)} & \makecell{Proximal oracle \\ (PO)}
		& Assumptions\\ \hline
		
		\makecell{SNVRG \\ \citep{zhou2018stochastic}}
		& $O\big(\big((n+\sqrt{n}\kappa)\log^3n\big)\log\frac{1}{\epsilon}\big)$
		& NA \tablefootnote{`NA' in the PO column means that these algorithms only considered the smooth case (i.e., $h(x)\equiv 0$ in problem \eqref{eq:problem}) in the PL setting. The proximal oracle is only required for the nonsmooth setting.\label{tabfn:na-pl}}
		& \makecell{Asp \ref{asp:avgsmooth}, \ref{asp:pl} \\ (finite-sum)} \\\hline
		
		\makecell{SNVRG \\ \citep{zhou2018stochastic}}
		& $O\big(\big((B+\sqrt{B}\kappa)\log^3B\big)\log\frac{1}{\epsilon}\big)$ \tablefootnote{$B:=\min\{n,\frac{\sigma^2}{\mu\epsilon}\}$.\label{tabfn:Bpl2}}
		& NA \textsuperscript{\getrefnumber{tabfn:na-pl}}
		& \makecell{Asp \ref{asp:avgsmooth}, \ref{asp:bv}, \ref{asp:pl} \\ (finite-sum or online)} \\\hline
		
		\makecell{Prox-SpiderBoost-PL\\ \citep{wang2019spiderboost}}
		& $O\big((n+\kappa^2)\log\frac{1}{\epsilon}\big)$
		& $O(\kappa\log\frac{1}{\epsilon})$
		&\makecell{Asp \ref{asp:avgsmooth}, \ref{asp:pl} \\ (finite-sum)} \\\hline
		
		\makecell{PAGE \\ \citep{li2021page}}
		& $O\big((n+\sqrt{n}\kappa)\log\frac{1}{\epsilon}\big)$
		& NA \textsuperscript{\getrefnumber{tabfn:na-pl}}
		& \makecell{Asp \ref{asp:avgsmooth}, \ref{asp:pl} \\ (finite-sum)} \\\hline
		
		\makecell{PAGE \\ \citep{li2021page}}
		& $O\big((B+\sqrt{B}\kappa)\log\frac{1}{\epsilon}\big)$ \textsuperscript{\getrefnumber{tabfn:Bpl2}}
		& NA \textsuperscript{\getrefnumber{tabfn:na-pl}}
		& \makecell{Asp \ref{asp:avgsmooth}, \ref{asp:bv}, \ref{asp:pl} \\ (finite-sum or online)} \\\hline
		
		\red{\makecell{SSRGD\\
				(this paper, Theorem~\ref{thm:pl-ssrgd})}}
		&\makecell{$O\big((n+\sqrt{n}\kappa)\log\frac{1}{\epsilon}\big)$}
		& $O(\kappa\log\frac{1}{\epsilon})$  &
		\makecell{Asp \ref{asp:avgsmooth}, \ref{asp:pl}\\ (finite-sum)}
		\\ \hline
		\red{\makecell{SSRGD\\
				(this paper, Theorem~\ref{thm:pl-ssrgd})}}
		& $O\big((B+\sqrt{B}\kappa)\log\frac{1}{\epsilon}\big)$ \textsuperscript{\getrefnumber{tabfn:Bpl2}}
		& $O(\kappa\log\frac{1}{\epsilon})$
		&  \makecell{Asp \ref{asp:avgsmooth}, \ref{asp:bv}, \ref{asp:pl} \\ (finite-sum or online)} \\\hline
	\end{tabular}
\end{table}

\subsection{Finding local minimum}
\label{sec:con-localmin}

Now, we consider the problem of finding the approximate $(\epsilon,\delta)$-local minimum (i.e., $\n{\nabla f(\hx)}\leq \epsilon$ and $\lambda_{\min}(\nabla^2 f(\hx))\geq -\delta$) in nonconvex optimization problems.
We compare our solution with several other recent theoretical results on finding approximate local minimum.
This includes those that adopt Neon/Neon2~\citep{xu2018first,allen2018neon2} (which involve some negative curvature searching procedure), such as \citep{agarwal2016finding,carmon2016accelerated,allen2018neon2,zhou2018finding,fang2018spider} and those by adding simple pertubations to fairly standard gradient methods,
such as PGD~\citep{jin2017escape}, PAGD~\citep{jin2018accelerated}, CNC-SGD~\citep{daneshmand2018escaping} and Stabilized SVRG~\citep{ge2019stable}.

We show that our SSRGD can find an $(\epsilon,\delta)$-local minimum and further improve the convergence result of Stabilized SVRG~\citep{ge2019stable} from $n^{2/3}/\epsilon^2$ to $n^{1/2}/\epsilon^2$ (see Table \ref{tab:localmin-finite}).
Similar to \citet{ge2019stable}, SSRGD for finding a local minimum is as simple as its counterpart for finding a first-order stationary point. This is done by just adding a random perturbation in each superepoch, and it does not require a negative curvature (NC) search subroutine (such as Neon/Neon2) or computing Hessian-vector products (such as FastCubic/CDHS).
Thus SSRGD (only uses stochastic gradients and random perturbations) can be easily applied in practice.
We note that the convergence rate of SSRGD can be better than Neon2+SPIDER~\citep{fang2018spider} if $\delta$ is very small (i.e., higher accuracy for second-order guarantee $\lambda_{\min}(\nabla^2 f(\hx))\geq -\delta$).
Also Neon2+SPIDER~\citep{fang2018spider} requires a negative curvature (NC) search subroutine (such as Neon/Neon2) and thus is more complicated than SSRGD. 
Our convergence analysis is also arguably simpler.
The previous results and our new results are summarized in Table \ref{tab:localmin-finite} (finite-sum case) and \ref{tab:localmin-online} (online case).
Also note that the first term of the convergence result of SSRGD (i.e., $\frac{\sqrt{n}}{\epsilon^2}$ or $\frac{1}{\epsilon^3}$) matches the corresponding result for finding the first-order optimal solution (See previous Table \ref{tab:1} or Figure \ref{fig:1}) and hence is optimal.

Finally, if we further assume that $f$ has $L_3$-Lipschitz continuous third-order derivative (i.e., Assumption \ref{asp:smooth-3rd}), we show that better convergence rate can be achieved, by replacing the super epoch part of SSRGD (Algorithm~\ref{alg:ssrgd}) by a negative-curvature search step (e.g., Neon2~\citep{allen2018neon2})). 
Currently, the best known result under this setting is achieved in \citet{zhou2018finding}, which also uses a negative-curvature search procedure. Our approach is similar to theirs and we obtain the same convergence rate (see Table~\ref{tab:localmin-online-3rd}).

\begin{table}[t]
	\centering
	\caption{Gradient complexity of algorithms for nonconvex finite-sum problem \eqref{prob:finite} under Asp \ref{asp:smoothgandh}}
	\label{tab:localmin-finite}
	\small
	\vspace{1mm}
	\begin{spacing}{1.4}
		\begin{tabular}{|c|c|c|c|}
			\hline
			Algorithms
			& \makecell{Stochastic gradient\\ complexity}
			& Guarantee & NC
			\tablefootnote{Negative Curvature search subroutine.\label{tabfn:nc-search}}
			 \\
			\hline
			PGD~\citep{jin2017escape}
			& $\tdo(\frac{n}{\epsilon^2} + \frac{n}{\delta^4})$
			& $(\epsilon,\delta)$-local min & No\\
			\hline
			PAGD~\citep{jin2018accelerated}
			& $\tdo(\frac{n}{\epsilon^{1.75}} + \frac{n}{\delta^{3.5}})$
			& $(\epsilon,\delta)$-local min & No\\
			\hline
			\makecell{Neon2+FastCubic/CDHS \\
				\citep{agarwal2016finding,carmon2016accelerated}}
			& $\tdo(\frac{n}{\epsilon^{1.5}}
			+\frac{n^{3/4}}{\epsilon^{1.75}}
			+\frac{n^{3/4}}{\delta^{3.5}}
			+\frac{n}{\delta^{3}})$
			& $(\epsilon,\delta)$-local min & Needed\\
			\hline
			\makecell{Neon2+SVRG \citep{allen2018neon2}}
			& $\tdo(\frac{n^{2/3}}{\epsilon^2}
			+\frac{n^{3/4}}{\delta^{3.5}}+\frac{n}{\delta^{3}})$
			& $(\epsilon,\delta)$-local min & Needed\\
			\hline
			\makecell{Neon2+SNVRG \citep{zhou2018finding}} &$\tdo(\frac{n^{1/2}}{\epsilon^2} +\frac{n^{3/4}}{\delta^{3.5}}+\frac{n}{\delta^{3}})$
			& $(\epsilon,\delta)$-local min & Needed\\
			\hline
			\makecell{Neon2+SPIDER \citep{fang2018spider}} &$\tdo(\frac{n^{1/2}}{\epsilon^2} +\frac{n^{1/2}}{\epsilon\delta^2}+\frac{1}{\epsilon\delta^3}+\frac{1}{\delta^5})$
			& $(\epsilon,\delta)$-local min & Needed\\
			\hline
			\makecell{Stabilized SVRG \citep{ge2019stable}}
			& $\tdo(\frac{n^{2/3}}{\epsilon^2}
			+\frac{n^{2/3}}{\delta^4}+\frac{n}{\delta^{3}})$
			& $(\epsilon,\delta)$-local min & No\\
			\hline
			\red{\makecell{SSRGD  (this paper, Theorem~\ref{thm:ssrgd-lm})}}
			& $\tdo(\frac{n^{1/2}}{\epsilon^2} +\frac{n^{1/2}}{\delta^4} + \frac{n}{\delta^3})$
			& $(\epsilon,\delta)$-local min & No\\
			\hline
		\end{tabular}
	\end{spacing}
\end{table}

\begin{table}[htb]
	\centering
	\caption{Gradient complexity of algorithms for nonconvex online  problem \eqref{prob:online} under Asp \ref{asp:smoothgandh} and \ref{asp:var2}}
	\label{tab:localmin-online}
	\small
	\vspace{1mm}
	\begin{spacing}{1.4}
	\begin{tabular}{|c|c|c|c|}
		\hline
		Algorithms & \makecell{Stochastic gradient\\ complexity} & Guarantee & NC \textsuperscript{\getrefnumber{tabfn:nc-search}}\\
		\hline
		Noisy SGD~\citep{ge2015escaping} & poly$(d,\frac{1}{\epsilon},\frac{1}{\delta})$
		& $(\epsilon,\delta)$-local min & No\\ \hline
		CNC-SGD~\citep{daneshmand2018escaping}& $\tdo(\frac{1}{\epsilon^{4}} +\frac{1}{\delta^{10}})$
		& $(\epsilon,\delta)$-local min & No\\ \hline
		Perturbed SGD~\citep{jin2019stochastic} & $\tdo(\frac{1}{\epsilon^{4}}+\frac{1}{\delta^8})$
		& $(\epsilon,\delta)$-local min & No\\ \hline
		SGD with averaging~\citep{fang2019sharp}& $\tdo(\frac{1}{\epsilon^{3.5}} +\frac{1}{\delta^{7}})$
		& $(\epsilon,\delta)$-local min & No\\ \hline
		Neon2+SCSG~\citep{allen2018neon2} & $\tdo(\frac{1}{\epsilon^{10/3}} +\frac{1}{\epsilon^2\delta^3}+\frac{1}{\delta^5})$
		& $(\epsilon,\delta)$-local min & Needed\\  \hline
		Neon2+Natasha2~\citep{allen2018natasha} & $\tdo(\frac{1}{\epsilon^{3.25}}+\frac{1}{\epsilon^3\delta}+\frac{1}{\delta^5})$
		& $(\epsilon,\delta)$-local min & Needed\\  \hline
		Neon2+SNVRG~\citep{zhou2018finding} &$\tdo(\frac{1}{\epsilon^{3}}+\frac{1}{\epsilon^2\delta^3}+\frac{1}{\delta^5})$
		& $(\epsilon,\delta)$-local min & Needed\\  \hline
		Neon2+SPIDER~\citep{fang2018spider} &$\tdo(\frac{1}{\epsilon^{3}}+\frac{1}{\epsilon^2\delta^2}+\frac{1}{\delta^5})$
		& $(\epsilon,\delta)$-local min & Needed\\  \hline
		\red{SSRGD (this paper, Theorem~\ref{thm:ssrgd-lm})} & $\tdo(\frac{1}{\epsilon^{3}}+\frac{1}{\epsilon^2\delta^3}+\frac{1}{\epsilon\delta^4})$ & $(\epsilon,\delta)$-local min & No\\  \hline
	\end{tabular}
\end{spacing}
\end{table}

\begin{table}[htb]
	\centering
	\caption{Gradient complexity for nonconvex online problem \eqref{prob:online} under Asp \ref{asp:smoothgandh}, \ref{asp:var2} and \ref{asp:smooth-3rd}}
	\label{tab:localmin-online-3rd}
	\small
	\vspace{1mm}
	\begin{spacing}{1.4}
		\begin{tabular}{|c|c|c|c|}
			\hline
			Algorithms & \makecell{Stochastic gradient\\ complexity} & Guarantee & NC \textsuperscript{\getrefnumber{tabfn:nc-search}} \\
			\hline
			FLASH~\citep{yu2017third} 
			&$\tdo(\frac{1}{\epsilon^{10/3}}+\frac{1}{\epsilon^2\delta^2}+\frac{1}{\delta^4})$
			& $(\epsilon,\delta)$-local min & Needed 
			\\  \hline
			SNVRG~\citep{zhou2018finding} &$\tdo(\frac{1}{\epsilon^{3}}+\frac{1}{\epsilon^2\delta^2}+\frac{1}{\delta^4})$
			& $(\epsilon,\delta)$-local min & Needed 
			\\  \hline
			\red{SSRGD (this paper, Theorem~\ref{thm:ssrgd-lm-3rd})} & $\tdo(\frac{1}{\epsilon^{3}}+\frac{1}{\epsilon^2\delta^2}+\frac{1}{\delta^4})$ & $(\epsilon,\delta)$-local min & Needed
			\\  \hline
		\end{tabular}
	\end{spacing}
\end{table}

\subsection{Comparison with the preliminary conference papers}
\label{subsec:comparison}
The present paper significantly extends the preliminary two conference papers \citep{li2018simple,li2019ssrgd}. The major differences between the present paper and the conference papers are summarized as follows.
(1)	We further simplify the proof of ProxSVRG+ in \citep{li2018simple}. 
See the proof of Theorem \ref{thm:1} in Appendix~\ref{app:proof-svrg}.
(2) We extend the original SSRGD in \citep{li2019ssrgd}, 
which can only handle smooth functions, to a proximal version that can handle nonsmooth functions as well. See Algorithm~\ref{alg:ssrgd_hl} and Theorem \ref{thm:ssrgd}.
(3)	We show SSRGD can achieve linear convergence rate if PL condition is satisfied. Moreover, SSRGD obtains new state-of-the-art results in this classical PL setting. This part is not published elsewhere. See Theorem \ref{thm:pl-ssrgd}.
(4)	We provide more details and intuitions in the analysis of SSRGD for escaping saddle point. See the proof of Theorem \ref{thm:ssrgd-lm} in Appendix~\ref{app:proof-localmin}.
(5) We briefly note that SSRGD, when combined with Neon2, can achieve better convergence rate under an additional third order smoothness assumption (Assumption~\ref{asp:smooth-3rd}). See Theorem~\ref{thm:ssrgd-lm-3rd}. This result is not published elsewhere.

\subsection{Organization}
The remaining paper is organized as follows.
Section~\ref{sec:pre} introduces the notations, standard assumptions and definitions in nonconvex 
optimization.
Section~\ref{sec:svrg} presents the ProxSVRG+ algorithm and its convergence results.
Section~\ref{sec:ssrgd} present the SSRGD algorithm and its convergence results.
Then, Section~\ref{sec:pl} present the results of ProxSVRG+ and SSRGD in the PL setting, where faster linear convergence can be obtained.
Finally, we show how to find an approximate local minimum instead of first-order stationary point via SSRGD and present the corresponding convergence results in Section~\ref{sec:localmin}. 
All proofs are deferred to the appendix.

\section{Preliminaries}
\label{sec:pre}
Let $[n]$ denote the set $\{1,2,\cdots,n\}$ and $\n{\cdot}$ the Euclidean norm for a vector or the spectral norm for a matrix.
Let $\inner{u}{v}$ denote the inner product of two vectors $u$ and $v$.
Let $\lambda_{\min}(A)$ denote the smallest eigenvalue of a symmetric matrix $A$.
Let $\mathbb{B}_x(r)$ denote a Euclidean ball with center $x$ and radius $r$.
We use $O(\cdot)$ and $\Omega(\cdot)$ to hide the absolute constant, and $\widetilde{O}(\cdot)$ to hide the logarithmic factor.

We assume that the nonsmooth function $h(x)$ in problem \eqref{eq:problem} is well structured such that the following proximal operator on $h$ can be computed efficiently:
\begin{equation}\label{eq:prox}
\prox_{\eta h}(x) := \arg\min_{y\in\R^d}{\Big(h(y)+\frac{1}{2\eta}\ns{y-x}\Big)}.
\end{equation}
For convex problems, one typically uses the optimality gap $\Phi(x)-\Phi(x^*)$ as the convergence criterion for problem \eqref{eq:problem} (see e.g., \citep{nesterov2014introductory}).
But for general nonconvex problems, one typically uses the gradient norm as the convergence criterion.
E.g., for smooth nonconvex problems (i.e., $h(x)\equiv 0$), \cite{ghadimi2013stochastic}, \cite{reddi2016stochastic} and \cite{lei2017non} used $\n{\nabla f(x)}$ to measure the convergence.
In order to analyze the convergence results for \emph{nonsmooth} nonconvex problems, following with \citet{ghadimi2016mini,reddi2016proximal}, we use the \emph{gradient mapping}:
\begin{equation}\label{eq:gradmap}
\calG_\eta(x) := \frac{1}{\eta}\Big(x - \prox_{\eta h}\big(x-\eta\nabla f(x)\big)\Big).
\end{equation}
Note that if $h(x)$ is a constant function (in particular, zero),
this gradient mapping reduces to  the ordinary gradient:
$\calG_\eta(x) = \nabla\Phi(x) = \nabla f(x)$.
Thus we use the norm of the gradient mapping $\calG_\eta(x)$ as the convergence criterion for problem \eqref{eq:problem} 
in the same way as in \citet{ghadimi2016mini,reddi2016proximal}.
\begin{definition}
	\label{def:eps}
	$\hx$ is called an $\epsilon$-approximate solution for problem \eqref{eq:problem}
	if $\E[\n{\calG_\eta(\hx)}]\leq \epsilon$.
	In particular, if $h(x)\equiv 0$ in \eqref{eq:problem}, this is equivalent to
	$\E[\n{\nabla f(\hx)}]\leq \epsilon$.
\end{definition}
Note that $\calG_\eta(x)$ has been already normalized by the step-size $\eta$, i.e., it is independent of different algorithms.
Let $x^+:=\prox_{\eta h}\big(x-\eta\nabla f(x)\big)$.
Then one can see that 
$\calG_\eta(x) := \frac{1}{\eta}\big(x - x^+\big)=\nabla f(x) + \partial h(x^+)$.
Moreover, if $\calG_\eta(x^*)=0$, then $x^*$ indeed is a first-order stationary point for problem \eqref{eq:problem}, i.e., $\partial \Phi(x^*) = 0$.

To measure the efficiency of a stochastic algorithm for solving problem \eqref{eq:problem}, we use the following SFO and PO oracle complexities.
\begin{definition}\label{def:oracle}
	\begin{enumerate}
		\item Stochastic first-order oracle (SFO): given a point $x$, SFO outputs a stochastic gradient $\nabla f_i(x)$ (i.e., gradient of one component/data in \eqref{prob:finite}) such that $\E_{i}[\nabla f_i(x)]=\nabla f(x)$.
		\item Proximal oracle (PO): given a point $x$, PO outputs
		the result of the proximal projection $\prox_{\eta h}(x)$ (see (\ref{eq:prox})).
	\end{enumerate}
\end{definition}

Moreover, in order to prove convergence results, we usually need the following standard smoothness assumptions. Besides, for stochastic/online problems \eqref{prob:online}, we also usually need the extra bounded variance assumption.
These assumptions are very standard in the optimization literature (see e.g., \citealp{nesterov2014introductory, ghadimi2016mini, lei2017non, li2018simple, allen2018natasha, zhou2018stochastic, fang2018spider, pham2019proxsarah, li2020unified}).

\begin{assumption}[$L$-smoothness]\label{asp:smooth}
	A function $f:\R^d\to \R$ is $L$-smooth if
	\begin{equation}\label{eq:smooth}
	\exists L >0, ~\mathrm{such~that} ~~  \n{\nabla f(x) - \nabla f(y)}\leq L \n{x-y}, \quad \forall x,y \in \R^d.
	\end{equation}
\end{assumption}

\begin{assumption}[Average $L$-smoothness]\label{asp:avgsmooth}
	A function $f(x):=\frac{1}{n}\sum_{i=1}^{n}f_i(x)$ is average $L$-smooth if
	\begin{equation}\label{eq:avgsmooth}
	\exists L >0, ~\mathrm{such~that} ~~  \E_i[\ns{\nabla f_i(x) - \nabla f_i(y)}]\leq L^2 \ns{x-y}, \quad \forall x,y \in \R^d.
	\end{equation}
\end{assumption}
It is not hard to see that Assumption \ref{asp:avgsmooth} implies Assumption \ref{asp:smooth}.

\begin{assumption}[Bounded variance]\label{asp:bv}
	The stochastic gradient has bounded variance if
	\begin{equation}\label{eq:bv}
	\exists \sigma >0, ~\mathrm{such~that} ~~ \E_{i}[\ns{\nabla f_i(x)-\nabla f(x)}]\leq \sigma^2, \quad \forall x \in \R^d.
	\end{equation}
\end{assumption}

\topic{PL setting} We also prove faster linear convergence rates for nonconvex functions under the Polyak-\L{}ojasiewicz (PL) condition \citep{polyak1963gradient}, i.e., $\ns{\nabla f(x)} \geq 2\mu (f(x)-f^*)$.
Similar to Definition \ref{def:eps}, due to the nonsmooth term $h(x)$ in problem \eqref{eq:problem},
we use the gradient mapping $\calG_\eta(x)$ (see \eqref{eq:gradmap}) to define a more general form of PL condition as follows:
\begin{assumption}[PL condition] \label{asp:pl}
	A function $\Phi:\R^d\to \R$ satisfies PL condition~\footnote{It is worth noting that the PL condition does not imply convexity of the function. For example, $f(x) = x^2 + 3\sin^2 x$ is a nonconvex function but $f$ satisfies PL condition with $\mu=1/32$.} if
	\begin{align}
	\exists \mu>0, ~\mathrm{such~that} ~& \ns{\calG_\eta(x)} \geq 2\mu (\Phi(x)-\Phi^*),~ \forall x\in \R^d    \label{eq:pl} \\
	& (\ns{\nabla f(x)} \geq 2\mu (f(x)-f^*) \mathrm{~if~}  h(x)\equiv 0 \mathrm{~in~} \eqref{eq:problem}). \notag
	\end{align}
\end{assumption}

When Assumption~\ref{asp:pl} holds, we say that it is the PL setting.
In the PL setting, we can show linear convergence to the global minimum.
Here, we directly use the optimality gap $\Phi(x)-\Phi^*$ as the convergence criterion (see e.g., \citealp{reddi2016proximal,lei2017non,li2018simple,zhou2018stochastic}), i.e., we use the following Definition \ref{def:eps_pl} in place of Definition \ref{def:eps} for the PL setting.
\begin{definition}
	\label{def:eps_pl}
	$\hx$ is called an $\epsilon$-approximate solution for problem \eqref{eq:problem} under PL condition (Assumption \ref{asp:pl})
	if $\E[\Phi(\hx)-\Phi^*]\leq \epsilon$.
\end{definition}

\topic{Local minima}
Finally, we define the approximate local minimum. Note that in this setting, we do not consider the nonsmooth term, i.e., $h(x)\equiv 0$ in \eqref{eq:problem}.
Otherwise the second-order guarantee in Definition \ref{def:localmin} is not well-defined for the nonsmooth term.
\begin{definition}
	\label{def:localmin}
	$\hx$ is called an $(\epsilon,\delta)$-local minimum for a twice-differentiable function $f$ if
	\begin{equation}\label{eq:localmin}
	\n{\nabla f(\hx)}\leq \epsilon  \mathrm{~~and~~}  \lambda_{\min}(\nabla^2 f(\hx))\geq -\delta.
	\end{equation}
\end{definition}
For finding an approximate local minimum instead of finding an approximate first-order stationary point, we usually need the extra smoothness assumption for the Hessians of $f_i$s.
\begin{assumption}[Gradient and Hessian Lipschitz]\label{asp:smoothgandh}
	A function $f_i:\R^d\to \R$ has an $L$-Lipschitz continuous gradient if
	\vspace{-1mm}
	\begin{equation}\label{smoothgrad}
	\vspace{-1mm}
	\exists L>0, ~\mathrm{such~that} ~~  \n{\nabla f_i(x)-\nabla f_i(y)}\leq L\n{x-y}, \quad \forall x, y \in \R^d,
	\end{equation}
	and has a $\rho$-Lipschitz continuous Hessian if
	\vspace{-1mm}
	\begin{equation}\label{eq:smoothhess}
	\vspace{-1mm}
	\exists \rho >0, ~\mathrm{such~that} ~~  \n{\nabla^2 f_i(x) - \nabla^2 f_i(y)}\leq \rho \n{x-y}, \quad \forall x,y \in \R^d.
	\end{equation}
\end{assumption}
Definition \ref{def:localmin} and Assumption \ref{asp:smoothgandh} are also standard in the literature
for finding local minima (see e.g., \citealp{ge2015escaping,jin2017escape,xu2018first,allen2018neon2,zhou2018finding,fang2018spider,ge2019stable,li2019ssrgd}).

For achieving a high probability result of finding the $(\epsilon,\delta)$-local minimum in the online case (i.e., Case 2 in Theorem \ref{thm:ssrgd-lm}), we need a slightly stronger version of bounded variance Assumption \ref{asp:var2}
in place of Assumption \ref{asp:bv}.
\begin{assumption}[Bounded Variance]
	\label{asp:var2}
	$\exists \sigma >0$, such that $\ns{\nabla f_i(x)-\nabla f(x)} \leq \sigma^2$,  $\forall i, x$.
\end{assumption}
We want to point out that Assumption \ref{asp:var2} can be relaxed such that $\n{\nabla f_i(x)-\nabla f(x)}$ has sub-Gaussian tail. Then it is sufficient for us to get a high probability bound by using Hoeffding bound on these sub-Gaussian variables.
Again, Assumption \ref{asp:var2} (or the relaxed sub-Gaussian version) is also standard in the online case for finding approximate local minima (see e.g., \citealp{allen2018neon2,zhou2018finding,fang2018spider,jin2019stochastic,fang2019sharp,li2019ssrgd}).

If we further assume that $f$ has $L_3$-Lipschitz continuous third-order derivative, 
it is possible to achieve even better convergence rate.

\begin{assumption}[Third-order  Derivative Lipschitz]\label{asp:smooth-3rd}
	A function $f:\R^d\to \R$ has an $L_3$-Lipschitz continuous third-order derivative if
	\begin{equation}\label{eq:smooth-3rd}
		\exists L_3>0, ~\mathrm{such~that} ~~  \n{\nabla^3 f(x)-\nabla^3 f(y)}_F\leq L_3\n{x-y}, \quad \forall x, y \in \R^d.
	\end{equation}
\end{assumption}

We note such smoothness assumption has already been used in other previous works such as \citep{anandkumar2016efficient, carmon2017convex,yu2017third} for escaping higher order saddle points or for achieving better results.

\section{ProxSVRG+}
\label{sec:svrg}
In this section, we propose a proximal stochastic gradient algorithm called ProxSVRG+ \citep{li2018simple}.
The details of ProxSVRG+ are described in Algorithm~\ref{alg:1}.
We call $B$ the batch size and $b$ the minibatch size.

\begin{algorithm}[!b]
	\caption{ProxSVRG+}
	\label{alg:1}
	\textbf{Input:}
	initial point $x_0$, batch size $B$, minibatch size $b$, epoch length $m$, step size $\eta$
	
	$\tx^0 = x_0$ \\
	\For{$s=0, 1, 2, \ldots $}{
		
		$g^s=\frac{1}{B}\sum_{j\in I_B}\nabla f_j(\tx^{s})$\,\,\footnotemark
		\label{line:firstB}
		
		\For{$k=1, 2, \ldots, m$}{
			\label{line:innerm}
			$t= sm+k$
			
			$v_{t-1} =\frac{1}{b}\sum_{i\in I_b}\big(\nabla f_i(x_{t-1})-\nabla f_i(\tx^s)\big) + g^s$
			\label{line:minibatch_svrg+}
			
			$x_t = \prox_{\eta h}(x_{t-1} - \eta v_{t-1})$
		}
		$\tx^{s+1} = x_{(s+1)m}$ \label{line:txs}
	}
\end{algorithm} 
\footnotetext{If $B=n$, ProxSVRG+ is almost the same as ProxSVRG~\citep{reddi2016proximal} (i.e., $g^s=\frac{1}{n}\sum_{j=1}^n\nabla f_j(\tx^{s-1})=\nabla f(\tx^{s-1})$) except some detailed parameter settings (e.g., step size, epoch length).} 

We note that our algorithm is similar to nonconvex ProxSVRG \citep{reddi2016proximal} and convex Prox-SVRG \citep{xiao2014proximal}.
Prox-SVRG \citep{xiao2014proximal} only focused on convex problems, while
nonconvex ProxSVRG \citep{reddi2016proximal} analyzed nonconvex problems.
The major difference of our ProxSVRG+ from Prox-SVRG and nonconvex ProxSVRG is that we avoid the computation of the full gradient at the beginning of each epoch, i.e., $B$ may not equal to $n$ (see Line 4 of Algorithm~\ref{alg:1}) while Prox-SVRG and nonconvex ProxSVRG used $B=n$. 
Our contribution mainly lies in the analysis, which is tighter. 
Note that even if we choose $B=n$, our analysis is stronger than ProxSVRG \citep{reddi2016proximal}
(see Table~\ref{tab:1}).
Also, our ProxSVRG+ shows that the ``stochastically controlled'' trick of SCSG \citep{lei2017non} (i.e., the epoch length is a geometrically distributed random variable) is not really necessary for achieving the desired convergence bound.\footnote{A similar observation was also made in Natasha1.5 \citep{allen2018natasha}. However, Natasha1.5 divides each epoch into multiple sub-epochs and randomly chooses the iteration point at the end of each sub-epoch. In our ProxSVRG+ (Algorithm~\ref{alg:1}), the epoch length is deterministic and it directly uses the point in the last iteration at the end of each epoch.
}
As a result, our ProxSVRG+ generalizes the result of SCSG to the more general nonsmooth nonconvex setting and yields simpler analysis.

\subsection{Convergence results of ProxSVRG+}
\label{sec:result-svrg}
Now, we present the main convergence results for ProxSVRG+.

\begin{theorem}\label{thm:1}
	Let the step size $\eta\leq\frac{1}{(1+2m/\sqrt{b})L}$, where $b$ denotes the minibatch size (Line~\ref{line:minibatch_svrg+} of Algorithm~\ref{alg:1}) and $m$ denotes the epoch length (Line~\ref{line:innerm} of Algorithm~\ref{alg:1}). Then Algorithm~\ref{alg:1} can find an $\epsilon$-approximate solution for problem \eqref{eq:problem}, i.e., $\E[\n{\calG_\eta(\hx)}]\leq \epsilon$ (see Definition \ref{def:eps}).
	We distinguish the following two cases:
	\begin{enumerate}
		\item
		(Finite-sum) Suppose Assumption \ref{asp:avgsmooth} holds. Let batch size $B=n$ and $m=\sqrt{b}$.
		Then the number of SFO calls is at most
		\begin{equation*}
		B + 12L(\Phi(x_0)-\Phi^*) \Big(\frac{B}{\epsilon^2\sqrt{b}}+\frac{b}{\epsilon^2}\Big) =n+ O\Big(\frac{n}{\epsilon^2\sqrt{b}}+\frac{b}{\epsilon^2}\Big). \footnote{In case the number of SFO calls is less than $B$ (i.e., if the total number of epochs $S<1$), we may add an explicit term $B$ to the number of SFO calls since the algorithm uses $B$ SFO calls at the beginning of the first epoch $s=0$ at Line \ref{line:firstB} of Algorithm~\ref{alg:1}. In this situation, ProxSVRG+ (Algorithm~\ref{alg:1})  terminates within the first epoch $s=0$, and the first term $B$ is dominating.}
		\end{equation*}
		\item
		(Finite-sum or online) Suppose Assumptions \ref{asp:avgsmooth} and \ref{asp:bv} hold.
		Let batch size $B=\min\{n, \frac{2\sigma^2}{\epsilon^2}\}$ and $m=\sqrt{b}$.
		Then the number of SFO calls is at most
		\begin{equation*}
		B + 12L(\Phi(x_0)-\Phi^*)\Big(\frac{B}{\epsilon^2\sqrt{b}}+\frac{b}{\epsilon^2}\Big) =\min\Big\{n, \frac{2\sigma^2}{\epsilon^2}\Big\} +O\Big(\min\Big\{n,\frac{\sigma^2}{\epsilon^2}\Big\}\frac{1}{\epsilon^2\sqrt{b}}+\frac{b}{\epsilon^2}\Big).
		\end{equation*}
	\end{enumerate}
	In both cases, the number of PO calls equals to the total number of iterations $T=Sm$, which is at most
	$$\frac{12L(\Phi(x_0)-\Phi^*)}{\epsilon^2}=O\left(\frac{1}{\epsilon^2}\right).$$
	\vspace{2mm}
\end{theorem}

\topic{Remark}
For simplicity of presentation and better comparison with previous bounds,
the bounds in Theorem~\ref{thm:1} are stated under condition $m=\sqrt{b}$.
In fact, our convergence analysis allows for more general values of $m$ and $b$,
and the bounds would depend on both $m$ and $b$. Please see the proof of Theorem~\ref{thm:1}
for the details.

The proof for Theorem \ref{thm:1} is notably different from that of ProxSVRG \citep{reddi2016proximal}. \cite{reddi2016proximal} used a Lyapunov function $R_t^{s}=\Phi(x_t)+c_t\ns{x_t-\tx^s}$ and showed that it decreases by the accumulated gradient mapping $\sum_{t=sm}^{sm+m-1}\ns{\calG_\eta(x_t)}$ in epoch $s$ (i.e., $R_{(s+1)m}^s\leq R_{sm}^s-\sum_{t=sm}^{(s+1)m-1}\ns{\calG_\eta(x_t)}$).
In our proof, we \emph{directly} show that $\Phi(x_t)$ decreases by the accumulated gradient mapping (i.e., $\Phi(x_{(s+1)m})\leq \Phi(x_{sm})-\sum_{t=sm}^{(s+1)m-1}\ns{\calG_\eta(x_t)}$) using a different analysis.
This is made possible by tightening the inequalities using Young's inequality and the relation between the variance of stochastic gradient estimator and the inner product of the gradient difference and point difference.
Also, our convergence result holds for any minibatch size $b\in[1,n]$ unlike ProxSVRG which requires $b\leq n^{2/3}$. Moreover, our ProxSVRG+ uses much less proximal oracle calls than ProxSVRG (see Table \ref{tab:1}).

For the online/stochastic Case 2,  we avoid the computation of the full gradient at the beginning of each epoch, i.e., $B$ may be less than $n$. Then, we use the similar idea in SCSG \citep{lei2017non} to bound the variance term, but we do not need the ``stochastically controlled'' trick of SCSG (as we discussed before) to achieve the desired convergence bound which yields a much simpler analysis for our ProxSVRG+.

We defer the proof of Theorem \ref{thm:1} to Appendix \ref{app:proof-svrg}. We want to mention that the proof in this paper simplifies our previous proof provided in \citep{li2018simple} and allows for a larger step size and leads to a better constant in the convergence result (i.e., 12 vs. 36).

\section{SSRGD}
\label{sec:ssrgd}
Now, we present our new SSRGD algorithm to solve the \emph{nonsmooth} nonconvex problems \eqref{eq:problem}.
The orginal version of SSRGD in \citep{li2019ssrgd} was designed to solve the smooth nonconvex problems (i.e., $h(x)\equiv 0$ in \eqref{eq:problem}) and to find the approximate local minima by escaping saddle points.
The new SSRGD algorithm in this paper can be seen as a proximal version of the original SSRGD algorithm.

\begin{algorithm}[!t]
	\caption{SSRGD}
	\label{alg:ssrgd_hl}
	\textbf{Input:} initial point $x_0$, batch size $B$, minibatch size $b$, epoch length $m$, step size $\eta$
	\\ 
	\For{$s=0,1,2,\ldots$}{
		\tikzmk{A}\If{$\n{\nabla f(x_{sm})}\leq \mathG \mathrm{~and ~not ~currently ~in ~a ~super ~epoch}$ }{
			\label{line:epochbegin}
			$x_{sm}\leftarrow x_{sm}+\xi,$ where $\xi$ uniformly $\sim \mathbb{B}_0(r)$, start a super epoch  \\ \nonl
			{\footnotesize // we use super epoch to avoid adding the perturbation steps too often near a saddle point}}\tikzmk{B} \boxalg{algbgcolor}
		$v_{sm}\leftarrow \frac{1}{B}\sum_{j\in I_B}\nabla f_j(x_{sm})$ \label{line:full} \\
		\For{$k=1, 2, \ldots, m$}{
			\label{line:innerm2}
			$t \leftarrow sm+k$ \\
			$x_{t} \leftarrow \blue{\prox_{\eta h}(x_{t-1} - \eta v_{t-1})}$ \label{line:proxgrad}\\
			$v_{t}\leftarrow \frac{1}{b}\sum_{i\in I_b}\big(\nabla f_i(x_{t})-\nabla f_i(x_{t-1})\big) + v_{t-1}$ \label{line:v} \\
			\tikzmk{A}\algorithmicif~meet stop condition
			\algorithmicthen~stop super epoch  \label{line:cond}
			\tikzmk{B} \boxalgone{algbgcolor}
		}\label{line:epochend}
	}
\end{algorithm}

In this section, we first focus on finding an $\epsilon$-approximate solution.
Hence, we ignore the super epoch part (Line 3--5 and Line 11 of Algorithm~\ref{alg:ssrgd_hl}) which is used for escaping saddle points, and add the proximal operator (Line 9 of Algorithm~\ref{alg:ssrgd_hl}) for dealing with this nonsmooth setting.
Line 3--5 and Line 11 will be useful in the next Section \ref{sec:localmin} when we aim to find an $(\epsilon,\delta)$-local minimum (see Definition \ref{def:localmin}).
Here, we show that SSRGD (Algorithm~\ref{alg:ssrgd_hl}) achieves the optimal convergence results for finding the $\epsilon$-approximate (first-order) solution for the nonsmooth nonconvex problems \eqref{eq:problem}. 
The main update step (Line 10) adopts the recursive formula which was
originally proposed in \citep{nguyen2017sarah}, and also used in several previous papers
on nonconvex problems such as 
SPIDER \citep{fang2018spider}, SpiderBoost \citep{wang2019spiderboost}, ProxSARAH \citep{pham2019proxsarah}.
Our main contribution in this section is a simple and clean analysis that is inspired by our analysis of ProxSVRG+.

\subsection{Convergence results of SSRGD}
\label{sec:result-ssrgd}

Now, we present the main theorem for SSRGD which can lead to the optimal convergence results.

\begin{theorem}\label{thm:ssrgd}
	Let the step size $\eta\leq\frac{1}{(1+\sqrt{(m-1)/b})L}$, where $b$ denotes the minibatch size (Line~\ref{line:v} of Algorithm~\ref{alg:ssrgd_hl}) and $m$ denotes the epoch length (Line~\ref{line:innerm2} of Algorithm~\ref{alg:ssrgd_hl}). 
	Then Algorithm~\ref{alg:ssrgd_hl} can find an $\epsilon$-approximate solution for problem \eqref{eq:problem}, i.e., $\E[\n{\calG_\eta(\hx)}]\leq \epsilon$ (see Definition \ref{def:eps}).
	We distinguish the following two cases:
	\begin{enumerate}
		\item
		(Finite-sum) Suppose Assumption \ref{asp:avgsmooth} holds. We let batch size $B=n$ and $m=b$.
		Then the number of SFO calls is at most
		\begin{equation*}
		B+ 8L(\Phi(x_0)-\Phi^*)\Big(\frac{B}{\epsilon^2 b}+\frac{b}{\epsilon^2}\Big) = n+O\Big(\frac{n}{\epsilon^2 b}+\frac{b}{\epsilon^2}\Big).
		\end{equation*}
		\item
		(Finite-sum or online) Suppose Assumptions \ref{asp:avgsmooth} and \ref{asp:bv} hold.
		We let batch size $B=\min\{n, \frac{2\sigma^2}{\epsilon^2}\}$ and $m=b$.
		Then the number of SFO calls is at most
		\begin{equation*}
		B+ 8L(\Phi(x_0)-\Phi^*)\Big(\frac{B}{\epsilon^2 b}+\frac{b}{\epsilon^2}\Big) =\min\Big\{n, \frac{2\sigma^2}{\epsilon^2}\Big\} +O\Big(\min\Big\{n,\frac{\sigma^2}{\epsilon^2}\Big\}\frac{1}{\epsilon^2 b}+\frac{b}{\epsilon^2}\Big).
		\end{equation*}
	\end{enumerate}
	In both cases, the number of PO calls equals to the total number of iterations $T=Sm$, which is at most
	$$\frac{8L(\Phi(x_0)-\Phi^*)}{\epsilon^2}=O\left(\frac{1}{\epsilon^2}\right).$$
\end{theorem}

\topic{Remark}
Similar to Theorem~\ref{thm:1}, our analysis allows for more general value of $m$ and $b$.
Compared with the convergence results of ProxSVRG+ (Theorem \ref{thm:1}), SSRGD improves the factor $\sqrt{b}$ to $b$, i.e., $O(\frac{B}{\epsilon^2\sqrt{b}}+\frac{b}{\epsilon^2})$ in Theorem \ref{thm:1} to $O(\frac{B}{\epsilon^2 b}+\frac{b}{\epsilon^2})$ in Theorem \ref{thm:ssrgd}.
In particular, in the finite-sum Case 1, the best result for ProxSVRG+ is $\frac{n^{2/3}}{\epsilon^2}$ where minibatch $b=n^{2/3}$, while the best result for SSRGD is $\frac{\sqrt{n}}{\epsilon^2}$ where minibatch $b=\sqrt{n}$.
Moreover, SSRGD can achieve the optimal upper bounds, 
matching lower bounds $\Omega(n+\frac{\sqrt{n}}{\epsilon^2})$ for the finite-sum case and $\Omega(B+\frac{\sqrt{B}}{\epsilon^2})$ for the online case, shown in \citep{fang2018spider,li2021page}.
We defer the proof of Theorem \ref{thm:ssrgd} to Appendix \ref{app:proof-ssrgd}.

\section{Faster Linear Convergence under PL Condition}
\label{sec:pl}
In this section, we show that better convergence can be achieved if the objective function $\Phi(x)$ satisfies the PL condition (Assumption \ref{asp:pl}).
$$\exists \mu>0, ~\mathrm{such~that} ~ \ns{\calG_\eta(x)} \geq 2\mu (\Phi(x)-\Phi^*),~ \forall x\in \R^d. $$
\cite{karimi2016linear} showed that PL condition is weaker than many conditions
(e.g., strong convexity (SC), restricted strong convexity (RSC) and weak strong convexity (WSC)
\citep{necoara2015linear}).
Also, if $\Phi$ is convex, PL condition is equivalent to the error bounds (EB) and quadratic growth (QG) condition \citep{luo1993error,anitescu2000degenerate}.

Note that under the PL condition, one can obtain a faster linear convergence $O(\cdot\log \frac{1}{\epsilon})$ (see Theorem~\ref{thm:pl1} and \ref{thm:pl-ssrgd}) rather than the sublinear convergence $O(\cdot\frac{1}{\epsilon^2})$ (see Theorem~\ref{thm:1} and \ref{thm:ssrgd}).
See Tables \ref{tab:1pl}--\ref{tab:3pl} for an overview of convergence results in this PL setting.
In many cases, although the objective function is globally nonconvex, some local regions (e.g., large gradient regions) may satisfy the PL condition. We prove that ProxSVRG+ (Algorithm~\ref{alg:1}) and SSRGD (Algorithm~\ref{alg:ssrgd_hl}) with same parameter settings for the finite-sum case can {automatically} switch to the faster linear convergence rate in these regions where PL condition is satisfied.
Also note that under the PL condition, we can use the optimality gap $\Phi(x)-\Phi^*$ as the convergence criterion (see Definition \ref{def:eps_pl}) instead of $\n{\calG_\eta(x)}$ (see Definition \ref{def:eps}).
Besides, we can directly use the final iteration $x_{Sm}$ as the output point in this PL setting instead of the randomly chosen one $\hx$. Similar to \citep{reddi2016proximal,li2018simple}, we mainly consider the case where the condition number $\kappa \geq \sqrt{n}$ in the following subsections.
Note that if $\kappa< \sqrt{n}$,  the SFO complexity of both ProxSVRG+ and SSRGD can be bounded by $O(n\log\frac{1}{\epsilon})$, i.e., independent with $\kappa$.
The detailed proofs of Theorem \ref{thm:pl1}--\ref{thm:pl-ssrgd} are deferred to Appendix~\ref{app:proof-pl}.

\subsection{ProxSVRG+ under PL Condition}
\label{sec:pl-svrg}
Similar to Theorem \ref{thm:1}, we provide the convergence result of ProxSVRG+ (Algorithm~\ref{alg:1}) under PL condition in the following theorem.

\begin{theorem}\label{thm:pl1}
	Let the step size $\eta\leq\frac{1}{(1+2m/\sqrt{b})L}$, where $b$ denotes the minibatch size (Line~\ref{line:minibatch_svrg+} of Algorithm~\ref{alg:1}) and $m$ denotes the epoch length (Line~\ref{line:innerm} of Algorithm~\ref{alg:1}). 
	Then the final iteration point $x_{Sm}$ in Algorithm~\ref{alg:1} satisfies $\E[\Phi(x_{Sm})-\Phi^*]\leq \epsilon$ under PL condition.
	We distinguish the following two cases:
	\begin{enumerate}
		\item (Finite-sum) Suppose Assumptions \ref{asp:avgsmooth} and \ref{asp:pl} hold.
		We let batch size $B=n$ and $m=\sqrt{b}$.
		Then the number of SFO calls can be bounded by
		\begin{equation*}
		\Big(\frac{B}{\sqrt{b}}+b\Big)\frac{3L}{\mu}\log\frac{2(\Phi(x_0)-\Phi^*)}{\epsilon}=O\Big(\big(\frac{n }{\sqrt{b}}+b\big)\kappa\log\frac{1}{\epsilon}\Big).
		\end{equation*}
		\item (Finite-sum or online) Suppose Assumptions \ref{asp:avgsmooth}, \ref{asp:bv} and \ref{asp:pl} hold. We let batch size  $B=\min\{n,\frac{\sigma^2}{\mu\epsilon}\}$ $m=\sqrt{b}$. Then
		the number of SFO calls can be bounded by
		\begin{equation*}
			\Big(\frac{B}{\sqrt{b}}+b\Big)\frac{3L}{\mu}\log\frac{2(\Phi(x_0)-\Phi^*)}{\epsilon}=O\Big(\big(\frac{\min\{n,\frac{\sigma^2}{\mu\epsilon}\} }{\sqrt{b}}+b\big)\kappa\log\frac{1}{\epsilon}\Big).
		\end{equation*} 
	\end{enumerate}
	In both cases, the number of PO calls equals to the total number of iterations $T=Sm$ which is bounded by $$\frac{3L}{\mu}\log\frac{2(\Phi(x_0)-\Phi^*)}{\epsilon}=O\Big(\kappa\log\frac{1}{\epsilon}\Big),$$
	where $\kappa:= \frac{L}{\mu}$.
\end{theorem}

\topic{Remark}
From the above theorem, we can see that under the PL condition, ProxSVRG+ (Algorithm~\ref{alg:1}) can achieve a faster linear convergence $O(\cdot\log \frac{1}{\epsilon})$ rather than the sublinear convergence $O(\cdot\frac{1}{\epsilon^2})$ (see Theorem~\ref{thm:1}).
We would like to mention that Theorem \ref{thm:pl1} uses exactly the same parameter setting as in Theorem \ref{thm:1} for the finite-sum case.
Hence, ProxSVRG+ can automatically switch to this faster linear convergence rate instead of the previous sublinear convergence as long as the objective function $\Phi(x)$ satisfies the PL condition in these regions.

\subsection{SSRGD under PL Condition}
\label{sec:pl-ssrgd}
Similar to Theorem \ref{thm:ssrgd}, we provide the convergence result of SSRGD (Algorithm~\ref{alg:ssrgd_hl}) under PL condition in the following theorem.

\begin{theorem}\label{thm:pl-ssrgd}
	Let the step size $\eta\leq\frac{1}{(1+\sqrt{(m-1)/b})L}$, where $b$ denotes the minibatch size (Line~\ref{line:v} of Algorithm~\ref{alg:ssrgd_hl}) and $m$ denotes the epoch length (Line~\ref{line:innerm2} of Algorithm~\ref{alg:ssrgd_hl}). 
	Then the final iteration point $x_{Sm}$ in Algorithm~\ref{alg:ssrgd_hl} satisfies $\E[\Phi(x_{Sm})-\Phi^*]\leq \epsilon$ under PL condition.
	We distinguish the following two cases:
	\begin{enumerate}
		\item (Finite-sum) Suppose Assumptions \ref{asp:avgsmooth} and \ref{asp:pl} hold.  We let batch size $B=n$ and $m=b$. Then 
		the number of SFO calls can be bounded by
		\begin{equation*}
		\Big(\frac{B}{b}+b\Big)\frac{2L}{\mu}\log\frac{2(\Phi(x_0)-\Phi^*)}{\epsilon}=O\Big(\big(\frac{n }{b}+b\big)\kappa\log\frac{1}{\epsilon}\Big).
		\end{equation*}
		\item (Finite-sum or online) Suppose Assumptions \ref{asp:avgsmooth}, \ref{asp:bv} and \ref{asp:pl} hold. We let batch size  $B=\min\{n,\frac{\sigma^2}{\mu\epsilon}\}$ and $m=b$. Then
		the number of SFO calls can be bounded by
		\begin{equation*}
		\Big(\frac{B}{b}+b\Big)\frac{2L}{\mu}\log\frac{2(\Phi(x_0)-\Phi^*)}{\epsilon}=O\Big(\big(\frac{\min\{n,\frac{\sigma^2}{\mu\epsilon}\} }{b}+b\big)\kappa\log\frac{1}{\epsilon}\Big).
		\end{equation*}
	\end{enumerate}
	In both cases, the number of PO calls equals to the total number of iterations $T=Sm$ which is bounded by $$\frac{2L}{\mu}\log\frac{2(\Phi(x_0)-\Phi^*)}{\epsilon}=O\Big(\kappa\log\frac{1}{\epsilon}\Big),$$
	where $\kappa:= \frac{L}{\mu}$.
\end{theorem}

\topic{Remark}
Similarly, under the PL condition, SSRGD (Algorithm~\ref{alg:ssrgd_hl}) also achieves a faster linear convergence $O(\cdot\log \frac{1}{\epsilon})$ rather than the previous sublinear convergence $O(\cdot\frac{1}{\epsilon^2})$ (see previous Theorem~\ref{thm:ssrgd}). Theorem \ref{thm:pl-ssrgd} also uses the same parameter setting as in Theorem \ref{thm:ssrgd} for the finite-sum case and hence SSRGD can also switch to this faster linear convergence rate when PL condition is satisfied as ProxSVRG+.
Compared with the convergence results of ProxSVRG+ (Theorem \ref{thm:pl1}), SSRGD improves the factor $\sqrt{b}$ to $b$, i.e., $O((\frac{B}{\sqrt{b}}+b)\kappa\log\frac{1}{\epsilon})$ in Theorem \ref{thm:pl1} to $O((\frac{B}{b}+b)\kappa\log\frac{1}{\epsilon})$ in Theorem \ref{thm:pl-ssrgd}.
In particular, the best result for ProxSVRG+ is $O(n^{2/3}\kappa\log\frac{1}{\epsilon})$ where minibatch $b=n^{2/3}$, while the best result for SSRGD is $O(\sqrt{n}\kappa\log\frac{1}{\epsilon})$ where minibatch $b=\sqrt{n}$.

\section{Finding Approximate Local Minima}
\label{sec:localmin}
In this section, we show that our SSRGD algorithm \citep{li2019ssrgd} can further find the approximate local minima.
SSRGD (Algorithm~\ref{alg:ssrgd_hl}) in Section \ref{sec:ssrgd} is just to finding an $\epsilon$-approximate (first-order) solution (see Definition \ref{def:eps}) not the $(\epsilon,\delta)$-local minimum (see Definition \ref{def:localmin}),
we ignored the super epoch part. In this section, we present the details of the algorithm which can be found in Algorithm~\ref{alg:ssrgd}. In particular, our algorithm is either in a normal epoch ($\spe=0$) or in a super epoch ($\spe=1$).
We call each inner loop ($m$ iterations) a normal epoch (Line \ref{line:innerstart}--\ref{line:innerend} of Algorithm~\ref{alg:ssrgd}), i.e., iterations from $t=sm+1$ to $t=sm+m$ consist of the epoch $s$.
A super epoch may contains multiple normal epochs.
We enter a super epoch if we are currently in a normal epoch and $v_{sm}$ has a small norm
(i.e., near a saddle point) (Line 4 of Algorithm~\ref{alg:ssrgd}).
When we enter a super epoch, we add a random perturbation to the current point $\tx$ (Line 7 of Algorithm~\ref{alg:ssrgd}).
We exit a super epoch if the function value decrease significantly $(f(\tx)-f(x_t)\geq \mathf)$
or the number of iterations exceeds a threshold ($ t-t_{\mathrm{init}}\geq \mathT$).
We exit a normal epoch (not in a super epoch) by stopping at a uniformly randomly chosen iteration out of $m$ iterations (Line 17 of Algorithm~\ref{alg:ssrgd}).

\begin{algorithm}[tb]
	\caption{SSRGD (full version for finding approximate local minima)}
	\label{alg:ssrgd}
		\textbf{Input:}
		initial point $x_0$, batch size $B$, minibatch size $b$, epoch length $m$, step size $\eta$, perturbation radius $r$, threshold function value $\mathf$, super epoch length $\mathT$  
		
		$\spe \leftarrow 0$
		
		\For{$s=0,1,2,\ldots$}{
			\If{$\spe = 0  \mathrm{~and~} \n{v_{sm}}\leq \mathG$}{ \label{line:super}
				 {$\spe \leftarrow 1$}
				
				 {$\tx\leftarrow x_{sm}, t_{\mathrm{init}}\leftarrow sm$}
				
				 $x_{sm}\leftarrow \tx +\xi,$ where $\xi$ uniformly $\sim \mathbb{B}_0(r)$ \label{line:init} \\ \nonl
				 {\footnotesize // we use super epoch to avoid adding the perturbation steps too often near a saddle point}
			}
			$v_{sm}\leftarrow \frac{1}{B}\sum_{j\in I_B}\nabla f_j(x_{sm})$ \label{line:up1}
			
			\For{$k=1, 2, \ldots, m$}{  \label{line:innerstart}
				 $t \leftarrow sm+k$
				
				 $x_{t} \leftarrow x_{t-1} - \eta v_{t-1}$ \label{line:update}
				
				 $v_{t}\leftarrow \frac{1}{b}\sum_{i\in I_b}\big(\nabla f_i(x_{t})-\nabla f_i(x_{t-1})\big) + v_{t-1}$   \label{line:up2}  
				
				\uIf{$\spe=1 \mathrm{~and~} (f(\tx)-f(x_t)\geq \mathf \mathrm{~or~}  t-t_{\mathrm{init}}\geq \mathT$)}{
					 $\spe \leftarrow 0;$ break
				}\ElseIf{$\spe=0$}{        	   		
					 	{break with probability $\frac{1}{m-k+1}$} \label{line:randomstop} \\ \nonl
						{\footnotesize // we stop this epoch by randomly choosing a point as the starting point of the next epoch }
					}
			} \label{line:innerend}
			{$x_{(s+1)m}\leftarrow x_t$} \label{line:randompoint}
		}
\end{algorithm}

\subsection{Convergence results of SSRGD for finding approximate local minima}
\label{sec:result-ssrgd-lm}

Now, we present the main theorem for SSRGD (Algorithm~\ref{alg:ssrgd}) for finding approximate local minima which corresponds to the convergence results listed in Table \ref{tab:localmin-finite} and \ref{tab:localmin-online}.
We would like to point out that in this local minima setting, we consider the \emph{smooth} nonconvex case $\Phi(x)=f(x)$ in problem \eqref{eq:problem}, i.e., the nonsmooth term $h(x)\equiv 0$. Otherwise the second-order guarantee in the definition of $(\epsilon,\delta)$-local minimum (Definition \ref{def:localmin}) is not well-defined for the nonsmooth term.
Also note that our SSRGD for finding an $(\epsilon,\delta)$-local minimum is as simple as its counterpart for finding an $\epsilon$-approximate first-order solution ($\|\nabla f(x)\|\leq \epsilon$) just by adding a random perturbation sometimes, without requiring a negative curvature search subroutine (such as Neon/Neon2) which is typically required by other algorithms.
Thus our SSRGD can be simply applied in practice for finding approximate local minimum, and also it leads to simpler convergence analysis.

\begin{theorem}\label{thm:ssrgd-lm}
	 Suppose $f$ satisfies Assumption \ref{asp:smoothgandh}, i.e., $f$ has an $L$-Lipshitz gradient and a $\rho$-Lipschitz Hessian.  Let step size $\eta= \tdo(\frac{1}{L})$, epoch length $m=b=\sqrt{B}$, where $B, b$ denote the batch and minibatch size. Moreover, let perturbation radius $r=\tdo\big(\min(\frac{\delta^3}{\rho^2\epsilon}, \frac{\delta^{3/2}}{\rho\sqrt{L}})\big)$, threshold function value $\mathf=\tdo(\frac{\delta^3}{\rho^2})$ and super epoch length $\mathT=\tdo(\frac{1}{\eta\delta})$. Denote $\Delta_0:= f(x_0)-f^*$, where $x_0$ is the initial point and $f^*$ is the optimal value of $f$.
	 Then Algorithm~\ref{alg:ssrgd} reaches to an ($\epsilon,\delta$)-local minimum at least once with high probability $1-\zeta$.
	We distinguish the following two cases:
	\begin{enumerate}
		\item (Finite-sum)
		Let batch size $B=n$.
		Then the number of stochastic gradient computations is at most
		\begin{equation*}
		\tdo\Big(\frac{L\Delta_0\sqrt{n}}{\epsilon^2}
		+\frac{L\rho^2\Delta_0\sqrt{n}}{\delta^4}
		+ \frac{\rho^2\Delta_0n}{\delta^3}\Big).
		\end{equation*}
		\item (Online)
		We further assume Assumptions \ref{asp:var2} holds. 
		Let batch size $B=\tdo(\frac{\sigma^2}{\epsilon^2})$.
		Then the number of stochastic gradient computations is at most
		\begin{equation*}
		\tdo\Big(\frac{L\Delta_0 \sigma}{\epsilon^3}
		+ \frac{L\rho^2\Delta_0\sigma}{\epsilon\delta^4}
		+\frac{\rho^2\Delta_0\sigma^2}{\epsilon^2\delta^3}\Big).
		\end{equation*}
	\end{enumerate}
\end{theorem}

\topic{Remark}
Note that we can also write Case 2 of Theorem \ref{thm:ssrgd-lm} as 
$$
\tdo(\frac{L\Delta_0 \sqrt{\min\{n,\frac{\sigma^2}{\epsilon^2}\}}}{\epsilon^2}
+ \frac{L\rho^2\Delta_0\sqrt{\min\{n,\frac{\sigma^2}{\epsilon^2}\}}}{\delta^4}
+\frac{\rho^2\Delta_0\min\{n,\frac{\sigma^2}{\epsilon^2}\}}{\delta^3})
$$ by letting $B=\min\{n, \tdo(\frac{\sigma^2}{\epsilon^2})\}$ 
in a similar way to Case 2 of Theorem \ref{thm:1} and Theorem \ref{thm:ssrgd}.
Due to the second-order guarantee, the proofs of the finite-sum case and the online case have more difference than previous first-order guarantee methods, so we split the proof of Theorem \ref{thm:ssrgd-lm} into two parts, one for case $B=n$ and one for $B\neq n$ (see Appendix \ref{app:proof-localmin} for more details).
Also note that if we ignore $\delta$ (second-order guarantee $\lambda_{\min}(\nabla^2 f(\hx))\geq -\delta$), e.g., $\delta=\infty$, then the convergence result provided in Theorem \ref{thm:ssrgd-lm} (i.e., $\frac{\sqrt{n}}{\epsilon^2}$ or $\frac{1}{\epsilon^3}$) matches its corresponding result with first-order guarantee in Theorem \ref{thm:ssrgd} (which is optimal for finding the $\epsilon$-approximate first-order solution $\n{\nabla f(\hx)}\leq \epsilon$).

Finally, we show that better convergence rate can be achieved if we further assume that $f$ has $L_3$-Lipschitz continuous third-order derivative (i.e., Assumption \ref{asp:smooth-3rd}).
This can be achieved by replacing the super epoch part of Algorithm~\ref{alg:ssrgd} by a negative-curvature search step (e.g., Neon2~\citep{allen2018neon2}). 
Our convergence result matches the best known result by \cite{zhou2018finding}, which also uses a negative-curvature search procedure. In particular, we obtain the following theorem.

\begin{theorem}[Online case under third-order Lipschitz]\label{thm:ssrgd-lm-3rd}
	Suppose that Assumptions \ref{asp:smoothgandh}, \ref{asp:var2} and \ref{asp:smooth-3rd} hold.  Let step size $\eta= \tdo(\frac{1}{L})$ and batch size $B=\tdo(\frac{\sigma^2}{\epsilon^2})$, epoch length $m=b=\sqrt{B}$, where $B, b$ denote the batch and minibatch size. 
	Denote $\Delta_0:= f(x_0)-f^*$, where $x_0$ is the initial point and $f^*$ is the optimal value of $f$.
	If we replace the super epoch part of Algorithm~\ref{alg:ssrgd} by a negative-curvature search step (e.g., Neon2~\citep{allen2018neon2}),
	then it reaches to an ($\epsilon,\delta$)-local minimum at least once with high probability $1-\zeta$.
	The number of stochastic gradient computations is at most
	\begin{equation*}
		\tdo\Big(\frac{L\Delta_0\sigma}{\epsilon^3}
		+\frac{L_3\Delta_0\sigma^2}{\epsilon^2\delta^2}
		+ \frac{L_3L^2\Delta_0}{\delta^4}\Big).
	\end{equation*}
\end{theorem}

\vspace{3mm}
\section*{Acknowledgements}
The authors would like to thank Rong Ge, Chi Jin, Cong Fang for useful discussions and clarifications of their results, and anonymous reviewers for many helpful and constructive suggestions.
The research is supported in part by the National Natural Science Foundation of China Grant 62161146004, Turing AI Institute of Nanjing and Xi'an Institute for Interdisciplinary Information Core Technology.

\appendix
\section{Missing Proofs for Section \ref{sec:svrg} ProxSVRG+}
\label{app:proof-svrg}
In this section, we provide the analysis for ProxSVRG+.
Our new proof simplifies our original proof in~\citet{li2018simple}.
Before proving Theorem \ref{thm:1}, we need a useful lemma for the proximal operator.
Here we use the following lemma in \citet{zhize2019unified},
instead of the previous Lemma 1 in \citet{li2018simple}.

\begin{lemma}[\citealp{zhize2019unified}]\label{lem:proximal}
	Let $x^+ := \prox_{\eta h}(x-\eta v)$. We have
	\begin{equation}\label{eq:proximal}
	h(x^+)\leq h(z) + \inner{v}{z-x^+}
	+\frac{1}{2\eta}\ns{z-x}
	-\frac{1}{2\eta}\ns{x^+-x}
	-\frac{1}{2\eta}\ns{z-x^+}, ~~\forall z\in\R^d.
	\end{equation}
\end{lemma}

\begin{proofof}{Theorem \ref{thm:1}}
Let $x_t := \prox_{\eta h}(x_{t-1}-\eta v_{t-1})$ and $\bx_t := \prox_{\eta h}\big(x_{t-1}-\eta \nabla f(x_{t-1})\big)$.
By letting $x^+=x_t, x=x_{t-1}, v=v_{t-1}$ and $z=\bx_t$ in \eqref{eq:proximal}, we have
\begin{align}\label{eq:xt}
	h(x_t)\leq h(\bx_t) + \inner{v_{t-1}}{\bx_t-x_t}
	+\frac{1}{2\eta}\ns{\bx_t-x_{t-1}}
	-\frac{1}{2\eta}\ns{x_t-x_{t-1}}
	-\frac{1}{2\eta}\ns{\bx_t-x_t}.
\end{align}
Besides, by letting $x^+=\bx_t, x=x_{t-1}, v=\nabla f(x_{t-1})$ and $z=x=x_{t-1}$ in \eqref{eq:proximal}, we have
\begin{align}\label{eq:bxt}
	h(\bx_t)\leq h(x_{t-1}) + \inner{\nabla f(x_{t-1})}{x_{t-1}-\bx_t}
	-\frac{1}{2\eta}\ns{\bx_t-x_{t-1}}
	-\frac{1}{2\eta}\ns{x_{t-1}-\bx_t}.
\end{align}
Moreover, in view of $L$-smoothness of $f$, we have
\begin{align}\label{eq:usesmooth}
	f(x_t)\leq f(x_{t-1}) + \inner{\nabla f(x_{t-1})}{x_t-x_{t-1}}
						+\frac{L}{2}\ns{x_t-x_{t-1}}.
\end{align}
We add \eqref{eq:xt}--\eqref{eq:usesmooth} to obtain (recall that $\Phi(x) :=f(x)+ h(x)$)
\begin{align}
\Phi(x_t)&\leq \Phi(x_{t-1})
		-\frac{1}{2\eta}\ns{x_{t-1}-\bx_t}
		-(\frac{1}{2\eta}-\frac{L}{2})\ns{x_t-x_{t-1}} \notag\\
	&\qquad\qquad	+ \inner{v_{t-1}-\nabla f(x_{t-1})}{\bx_t-x_t}
		-\frac{1}{2\eta}\ns{\bx_t-x_t} \notag \\
	&\leq \Phi(x_{t-1}) -\frac{1}{2\eta}\ns{x_{t-1}-\bx_t}
			-(\frac{1}{2\eta}-\frac{L}{2})\ns{x_t-x_{t-1}}
			+\frac{\eta}{2} \ns{v_{t-1}-\nabla f(x_{t-1})} \label{eq:young}\\
	&= \Phi(x_{t-1}) -\frac{\eta}{2}\ns{\calG_\eta(x_{t-1})}
	-(\frac{1}{2\eta}-\frac{L}{2})\ns{x_t-x_{t-1}}
	+\frac{\eta}{2} \ns{v_{t-1}-\nabla f(x_{t-1})}, \label{eq:key}
\end{align}
where \eqref{eq:young} uses Young's inequality,
and \eqref{eq:key} uses the definition of gradient mapping $\calG_\eta(x_{t-1})$ (see \eqref{eq:gradmap}) and recall $\bx_t := \prox_{\eta h}\big(x_{t-1}-\eta \nabla f(x_{t-1})\big)$.

Now, we bound the variance term in \eqref{eq:key} as follows, where the expectations are over $I_b$ and $I_B$:
\begin{align}
&\E\Big[\ns{v_{t-1}-\nabla f(x_{t-1})}\Big] \notag\\
&= \E\Big[\Big\|\frac{1}{b}\sum_{i\in I_b}\Big(\nabla f_i(x_{t-1})-\nabla f_i(\tx^{s})\Big)
-\big(\nabla f(x_{t-1})-g^s\big)\Big\|^2\Big] \notag\\
&=\E\Big[\Big\|\frac{1}{b}\sum_{i\in I_b}\Big(\nabla f_i(x_{t-1})-\nabla f_i(\tx^{s})\Big)
-\Big(\nabla f(x_{t-1})
- \frac{1}{B}\sum_{j\in I_B}\nabla f_j(\tx^{s})\Big)\Big\|^2\Big] \notag\\
&= \E\Big[\Big\|\frac{1}{b}
\sum_{i\in I_b}\Big(\big(\nabla f_i(x_{t-1})-\nabla f_i(\tx^{s})\big)
-\big(\nabla f(x_{t-1})-\nabla f(\tx^{s})\big)\Big)
+\frac{1}{B}\sum_{j\in I_B}\Big(\nabla f_j(\tx^{s})- \nabla f(\tx^{s})\Big)
\Big\|^2\Big] \notag \\
&= \E\Big[\Big\|\frac{1}{b}
\sum_{i\in I_b}\Big(\big(\nabla f_i(x_{t-1})-\nabla f_i(\tx^{s})\big)
-\big(\nabla f(x_{t-1})-\nabla f(\tx^{s})\big)\Big)\Big\|^2\Big] \notag\\
&\qquad \qquad      +\E\Big[\Big\|
\frac{1}{B}\sum_{j\in I_B}\Big(\nabla f_j(\tx^{s})- \nabla f(\tx^{s})\Big)
\Big\|^2\Big] \label{eq:ijind}\\
&= \frac{1}{b^2}\E\Big[\sum_{i\in I_b}\Big\|
\Big(\big(\nabla f_i(x_{t-1})- \nabla f_i(\tx^{s})\big)
-\big(\nabla f(x_{t-1})-\nabla f(\tx^{s})\big)\Big)\Big\|^2\Big]\notag\\
&\qquad \qquad      +\E\Big[\Big\|
\frac{1}{B}\sum_{j\in I_B}\Big(\nabla f_j(\tx^{s})- \nabla f(\tx^{s})\Big)
\Big\|^2\Big] \label{eq:v1}\\
&\leq \frac{1}{b^2}
\E\Big[\sum_{i\in I_b}\big\|\nabla f_i(x_{t-1})- \nabla f_i(\tx^{s}) \big\|^2\Big]
+\E\Big[\Big\|
\frac{1}{B}\sum_{j\in I_B}\Big(\nabla f_j(\tx^{s})- \nabla f(\tx^{s})\Big)
\Big\|^2\Big]  \label{eq:v2}\\
&\leq \frac{L^2}{b}\E[\ns{x_{t-1}-\tx^{s}}]
+\frac{I\{B<n\}\sigma^2}{B} \label{eq:useasp},
\end{align}
where (\ref{eq:ijind}) holds due to the independence of $I_b$ and $I_B$,
(\ref{eq:v1}) holds since $\E[\ns{x_1+x_2+\cdots+x_k}]=\sum_{i=1}^k\E[\ns{x_i}]$ if $x_1, x_2, \ldots, x_k$ are independent and of mean zero,
(\ref{eq:v2}) uses the fact that $\E[\ns{x-\E x}]\leq \E[\ns{x}]$, for any random variable $x$,
and the last inequality (\ref{eq:useasp}) holds due to the average L-smoothness Assumption \ref{asp:avgsmooth} and bounded variance Assumption \ref{asp:bv}.
Note that the second term $\frac{I\{B<n\}\sigma^2}{B}$ in \eqref{eq:useasp} can be deleted (i.e., Assumption \ref{asp:bv} is not needed) if we choose $B=n$.

Now, we plug \eqref{eq:useasp} into \eqref{eq:key} to obtain
\begin{align}
&\E[\Phi(x_t)]  \notag\\
&\leq \E\Big[\Phi(x_{t-1}) -\frac{\eta}{2}\ns{\calG_\eta(x_{t-1})}
-(\frac{1}{2\eta}-\frac{L}{2})\ns{x_t-x_{t-1}}
+\frac{\eta L^2}{2b} \ns{x_{t-1}-\tx^s}
+\frac{I\{B<n\}\eta\sigma^2}{2B}\Big] \notag\\
&\leq \E\Big[\Phi(x_{t-1}) -\frac{\eta}{2}\ns{\calG_\eta(x_{t-1})}
-(\frac{1}{2\eta}-\frac{L}{2})\frac{1}{\alpha_t +1}\ns{x_t-\tx^{s}}
+(\frac{1}{2\eta}-\frac{L}{2})\frac{1}{\alpha_t}\ns{x_{t-1}-\tx^s} \notag\\
	& \qquad \qquad +\frac{\eta L^2}{2b} \ns{x_{t-1}-\tx^s}
		+\frac{I\{B<n\}\eta\sigma^2}{2B}\Big], \label{eq:useyoung}
\end{align}
where \eqref{eq:useyoung} uses Young's inequality $\ns{x_t-\tx^{s}}\leq (1+\alpha_t)\ns{x_t-x_{t-1}}+\big(1+\frac{1}{\alpha_t}\big)\ns{x_{t-1}-\tx^{s}}$, i.e.,
$-\ns{x_t-x_{t-1}}\leq -\frac{1}{\alpha_t +1}\ns{x_t-\tx^{s}} + \frac{1}{\alpha_t}\ns{x_{t-1}-\tx^{s}}$.
Also let step size $\eta\leq 1/L$ (so that $\frac{1}{2\eta}-\frac{L}{2}\geq 0$).

Adding \eqref{eq:useyoung} for all iteration in epoch $s$, i.e., $t=sm+1$ to $t=sm+m$,  we get
\begin{align}
&\E[\Phi(x_{(s+1)m})]  \notag\\
&\leq \E\Big[\Phi(x_{sm}) - \sum_{t=sm+1}^{sm+m}\frac{\eta}{2}\ns{\calG_\eta(x_{t-1})}
-\sum_{t=sm+1}^{sm+m}(\frac{1}{2\eta}-\frac{L}{2})\frac{1}{\alpha_t +1}\ns{x_t-\tx^{s}} \notag\\
& \qquad \qquad
+\sum_{t=sm+1}^{sm+m}\Big((\frac{1}{2\eta}-\frac{L}{2})\frac{1}{\alpha_t}+\frac{\eta L^2}{2b}\Big)\ns{x_{t-1}-\tx^s}
+\sum_{t=sm+1}^{sm+m}\frac{I\{B<n\}\eta\sigma^2}{2B}\Big] \notag\\
&\leq \E\Big[\Phi(x_{sm}) - \sum_{t=sm+1}^{sm+m}\frac{\eta}{2}\ns{\calG_\eta(x_{t-1})}
-\sum_{t=sm+1}^{sm+m-1}(\frac{1}{2\eta}-\frac{L}{2})\frac{1}{\alpha_t +1}\ns{x_t-\tx^{s}} \notag\\
& \qquad \qquad
+\sum_{t=sm+2}^{sm+m}\Big((\frac{1}{2\eta}-\frac{L}{2})\frac{1}{\alpha_t}+\frac{\eta L^2}{2b}\Big)\ns{x_{t-1}-\tx^s}
+\sum_{t=sm+1}^{sm+m}\frac{I\{B<n\}\eta\sigma^2}{2B}\Big] \label{eq:usexsm}\\
&= \E\Big[\Phi(x_{sm}) - \sum_{t=sm+1}^{sm+m}\frac{\eta}{2}\ns{\calG_\eta(x_{t-1})}
+\sum_{t=sm+1}^{sm+m}\frac{I\{B<n\}\eta\sigma^2}{2B}
\notag\\
& \qquad \qquad
-\sum_{t=sm+1}^{sm+m-1}\Big((\frac{1}{2\eta}-\frac{L}{2})\frac{1}{\alpha_t +1}-(\frac{1}{2\eta}-\frac{L}{2})\frac{1}{\alpha_{t+1}}-\frac{\eta L^2}{2b}\Big)\ns{x_t-\tx^{s}}
\Big] \notag\\
&\leq \E\Big[\Phi(x_{sm}) - \sum_{t=sm+1}^{sm+m}\frac{\eta}{2}\ns{\calG_\eta(x_{t-1})}
+\sum_{t=sm+1}^{sm+m}\frac{I\{B<n\}\eta\sigma^2}{2B}
\Big] \label{eq:alphat}
\end{align}
where \eqref{eq:usexsm} holds since $\ns{\cdot}$ is always non-negative and $\tx^s=x_{sm}$, and \eqref{eq:alphat} holds by choosing $\alpha_t$ and $\eta$ such that $(\frac{1}{2\eta}-\frac{L}{2})\frac{1}{\alpha_t +1}-(\frac{1}{2\eta}-\frac{L}{2})\frac{1}{\alpha_{t+1}}-\frac{\eta L^2}{2b}\geq 0$ for $sm+1\leq t \leq sm+m-1$.
Concretely, if we choose $\alpha_t=2(t \% m)-1$ and $\eta \leq \frac{1}{L}$, then 
for any $sm+1\leq t \leq sm+m-1$, we have that
$$
(\frac{1}{2\eta}-\frac{L}{2})\frac{1}{\alpha_t +1}-(\frac{1}{2\eta}-\frac{L}{2})\frac{1}{\alpha_{t+1}}-\frac{\eta L^2}{2b}
\geq 
\frac{1-\eta L}{2\eta}(\frac{1}{2(m-1)}-\frac{1}{2m-1})-\frac{\eta L^2}{2b}\geq 0.
$$
Note that the last inequality is quadratic in $\eta$.
We can verify that choosing $\eta\leq \frac{1}{(1+2m/\sqrt{b})L}$ suffices to make the inequality hold.

Now, we sum up \eqref{eq:alphat} for all epochs $0\leq s\leq S-1$ as follows:
\begin{align}
\E[\Phi(x_{Sm})-\Phi^*] &\leq \E\Big[\Phi(x_0)-\Phi^*
-\sum_{s=0}^{S-1}\sum_{t=sm+1}^{sm+m}\frac{\eta}{2}\ns{\calG_\eta(x_{t-1})}
+\sum_{s=0}^{S-1}\sum_{t=sm+1}^{sm+m}\frac{I\{B<n\}\eta\sigma^2}{2B}\Big] \notag\\
\E[\ns{\calG_\eta(\hx)}] &\leq \frac{2\big(\Phi(x_0)-\Phi^*\big)}{Sm \eta}
+\frac{I\{B<n\}\sigma^2}{B}   \label{eq:randomhx}\\
&\leq  \frac{\epsilon^2}{2}+\frac{\epsilon^2}{2} =\epsilon^2. \label{eq:finaleps}
\end{align}
Note that $\E[\n{\calG_\eta(\hx)}] \leq \sqrt{\E[\ns{\calG_\eta(\hx)}]} \leq \epsilon$.
The first inequality in \eqref{eq:finaleps} holds by randomly choose $\hx$ from $\{x_{t-1}\}_{t\in [Sm]}$, and the second in \eqref{eq:finaleps} holds by choosing $Sm\geq \frac{4(\Phi(x_0)-\Phi^*)}{\epsilon^2 \eta}$ and $B\geq \min\{n, \frac{2\sigma^2}{\epsilon^2}\}$.

Now, we can see that the total number of iterations is
$$
T=Sm= \frac{4(\Phi(x_0)-\Phi^*\big)}{\epsilon^2 \eta}.
$$
Choosing $\eta= \frac{1}{(1+2m/\sqrt{b})L}$, we can see that
the number of PO calls equals to 
$$T=Sm=\frac{4(\Phi(x_0)-\Phi^*\big)}{\epsilon^2 \eta}= 
\frac{4(\Phi(x_0)-\Phi^*\big)(1+2m/\sqrt{b})L}{\epsilon^2}.$$
The number of SFO calls equals to 
$$SB+Smb=
\frac{4L(\Phi(x_0)-\Phi^*)(1+2m/\sqrt{b})}{\epsilon^2}\big(\frac{B}{m}+b\big).
$$

If we choose $m=\sqrt{b}$ (then $\eta\leq \frac{1}{(1+2m/\sqrt{b})L}=\frac{1}{3L}$), 
the total number of PO calls equals to $T=Sm=\frac{12L(\Phi(x_0)-\Phi^*)}{\epsilon^2}$.
The number of SFO calls is $12L(\Phi(x_0)-\Phi^*)\big(\frac{n}{\epsilon^2\sqrt{b}}+\frac{b}{\epsilon^2}\big)$
if $B=n$
(In this case, the second term in (\ref{eq:randomhx}) is 0 and thus Assumption \ref{asp:bv} is not needed), and
$12L(\Phi(x_0)-\Phi^*)\big(\frac{B}{\epsilon^2\sqrt{b}}+\frac{b}{\epsilon^2}\big)$
if $B\geq \min\{n, \frac{2\sigma^2}{\epsilon^2}\}$. 

In case the number of SFO calls is less than $B$ (i.e., if the total number of epochs $S<1$), we may add an explicit term $B$ to the SFO result since the algorithm at least uses $B$ SFO calls in the first epoch $s=0$ at Line \ref{line:firstB} of Algorithm~\ref{alg:1}. In this situation, ProxSVRG+ (Algorithm~\ref{alg:1}) terminates within the first epoch $s=0$.
\end{proofof}

\section{Missing Proofs for Section \ref{sec:ssrgd} SSRGD}
\label{app:proof-ssrgd}
Now, we provide the detailed proofs for Theorem \ref{thm:ssrgd}.

\begin{proofof}{Theorem \ref{thm:ssrgd}}
First, according to the update step $x_t := \prox_{\eta h}(x_{t-1}-\eta v_{t-1})$, we recall the key inequality \eqref{eq:key}:
\begin{align}
\Phi(x_t) \leq \Phi(x_{t-1}) -\frac{\eta}{2}\ns{\calG_\eta(x_{t-1})}
-(\frac{1}{2\eta}-\frac{L}{2})\ns{x_t-x_{t-1}}
+\frac{\eta}{2} \ns{v_{t-1}-\nabla f(x_{t-1})}.\label{eq:key-ss}
\end{align}
Now, we bound the variance term in \eqref{eq:key-ss} as follows:
\begin{align}
&\E[\ns{v_{t-1}-\nabla f(x_{t-1})}] \notag\\
&= \E\Big[\Big\|\frac{1}{b}\sum_{i\in I_b}\big(\nabla f_i(x_{t-1})-\nabla f_i(x_{t-2})\big)
+ v_{t-2}-\nabla f(x_{t-1})\Big\|^2\Big] \notag\\
&= \E\Big[\Big\|\frac{1}{b}
\sum_{i\in I_b}\Big(\big(\nabla f_i(x_{t-1})-\nabla f_i(x_{t-2})\big)
-\big(\nabla f(x_{t-1})-\nabla f(x_{t-2})\big)\Big)
+v_{t-2}-\nabla f(x_{t-2}) \Big\|^2\Big] \notag \\
&= \E\Big[\Big\|\frac{1}{b}
\sum_{i\in I_b}\Big(\big(\nabla f_i(x_{t-1})-\nabla f_i(x_{t-2})\big)
-\big(\nabla f(x_{t-1})-\nabla f(x_{t-2})\big)\Big)\Big\|^2\Big] \notag\\
&\qquad \qquad
+\E[\| v_{t-2}-\nabla f(x_{t-2}) \|^2] \label{eq:ijind-ss}\\
&= \frac{1}{b^2}\E\Big[\sum_{i\in I_b}\Big\|
\big(\nabla f_i(x_{t-1})-\nabla f_i(x_{t-2})\big)
-\big(\nabla f(x_{t-1})-\nabla f(x_{t-2})\big)\Big\|^2\Big] \notag\\
&\qquad \qquad
+\E[\| v_{t-2}-\nabla f(x_{t-2}) \|^2] \label{eq:v1-ss} \\
&\leq \frac{1}{b^2}
\E\Big[\sum_{i\in I_b}\Big\|\nabla f_i(x_{t-1})-\nabla f_i(x_{t-2})\Big\|^2\Big]
+\E[\| v_{t-2}-\nabla f(x_{t-2}) \|^2] \qquad\qquad\qquad\qquad\label{eq:v2-ss}\\
&\leq \frac{L^2}{b}\E[\ns{x_{t-1}-x_{t-2}}]
+\E[\| v_{t-2}-\nabla f(x_{t-2}) \|^2] \label{eq:useasp-ss},
\end{align}
where \eqref{eq:ijind-ss} and \eqref{eq:v1-ss} use the law of total expectation and $\E[\ns{y_1+y_2+\cdots+y_k}]=\sum_{i=1}^k\E[\ns{y_i}]$ if $y_1, y_2, \ldots, y_k$ are independent and of mean zero, \eqref{eq:v2-ss} uses the fact $\E[\ns{x-\E x}]\leq \E[\ns{x}]$, and \eqref{eq:useasp-ss} holds due to the average L-smoothness Assumption \ref{asp:avgsmooth}.

Note that for $\E[\| v_{t-2}-\nabla f(x_{t-2}) \|^2]$ in \eqref{eq:useasp-ss}, we can reuse the same computation above. Thus we can sum up \eqref{eq:useasp-ss} from the beginning of this epoch $sm$ to the point $t-1$,
\begin{align}
\E[\ns{v_{t-1}-\nabla f(x_{t-1})}] &\leq \frac{L^2}{b}\sum_{j=sm+1}^{t-1}\E[\ns{x_{j}-x_{j-1}}]
+\E[\| v_{sm}-\nabla f(x_{sm}) \|^2] \label{eq:var} \\
&\leq \frac{L^2}{b}\sum_{j=sm+1}^{t-1}\E[\ns{x_{j}-x_{j-1}}]
+\frac{I\{B<n\}\sigma^2}{B} \label{eq:varfinite-ss},
\end{align}
Now, we take expectations for \eqref{eq:key-ss} and then sum it up from the beginning of this epoch $s$, i.e., iterations from $sm+1$ to $t$, by plugging the variance \eqref{eq:varfinite-ss} into them to get:
\begin{align}
\E[\Phi(x_{t})] &\leq \E[\Phi(x_{sm})] - \frac{\eta}{2}\sum_{j=sm+1}^{t}\E[\ns{\calG_\eta(x_{j-1})}]
- \big(\frac{1}{2\eta}- \frac{L}{2}\big)\sum_{j=sm+1}^{t}\E[\ns{x_j-x_{j-1}}] \notag\\
&\qquad + \frac{\eta L^2}{2b}\sum_{k=sm+1}^{t-1}\sum_{j=sm+1}^{k}\E[\ns{x_{j}-x_{j-1}}]
+ \frac{\eta}{2}\sum_{j=sm+1}^{t}\frac{I\{B<n\}\sigma^2}{B} \notag\\
&\leq \E[\Phi(x_{sm})] - \frac{\eta}{2}\sum_{j=sm+1}^{t}\E[\ns{\calG_\eta(x_{j-1})}]
- \big(\frac{1}{2\eta}- \frac{L}{2}\big)\sum_{j=sm+1}^{t}\E[\ns{x_j-x_{j-1}}] \notag\\
&\qquad + \frac{\eta L^2(t-1-sm)}{2b}\sum_{j=sm+1}^{t}\E[\ns{x_{j}-x_{j-1}}]
+ \frac{\eta}{2}\sum_{j=sm+1}^{t}\frac{I\{B<n\}\sigma^2}{B}\notag\\
&\leq \E[\Phi(x_{sm})] - \frac{\eta}{2}\sum_{j=sm+1}^{t}\E[\ns{\calG_\eta(x_{j-1})}]
+ \frac{\eta}{2}\sum_{j=sm+1}^{t}\frac{I\{B<n\}\sigma^2}{B} \notag\\
&\qquad
- \Big(\big(\frac{1}{2\eta}- \frac{L}{2}\big)-\frac{\eta L^2(m-1)}{2b}\Big)\sum_{j=sm+1}^{t}\E[\ns{x_j-x_{j-1}}] \label{eq:bm1-ss}\\
&\leq \E[\Phi(x_{sm})] - \frac{\eta}{2}\sum_{j=sm+1}^{t}\E[\ns{\calG_\eta(x_{j-1})}] + \frac{\eta}{2}\sum_{j=sm+1}^{t}\frac{I\{B<n\}\sigma^2}{B} \label{eq:eta1-ss},
\end{align}
where \eqref{eq:bm1-ss} holds due to here $t\leq sm+m$ in epoch $s$,
\eqref{eq:eta1-ss} holds if the step size $\eta\leq \frac{1}{(1+\sqrt{(m-1)/b})L}$.

Now, we sum up \eqref{eq:eta1-ss} for all epochs $0\leq s\leq S-1$ to finish the proof as follows:
\begin{align}
\E[\Phi(x_{Sm})-\Phi^*] &\leq \E\Big[\Phi(x_0)-\Phi^*
-\frac{\eta}{2}\sum_{s=0}^{S-1}\sum_{t=sm+1}^{sm+m}\ns{\calG_\eta(x_{t-1})}
+\frac{\eta}{2}\sum_{s=0}^{S-1}\sum_{t=sm+1}^{sm+m}\frac{I\{B<n\}\sigma^2}{B}\Big] \notag\\
\E[\ns{\calG_\eta(\hx)}] &\leq \frac{2\big(\Phi(x_0)-\Phi^*\big)}{Sm \eta}
+\frac{I\{B<n\}\sigma^2}{B}  \label{eq:randomhx-ss}\\
&\leq \frac{\epsilon^2}{2}+\frac{\epsilon^2}{2} =\epsilon^2. \label{eq:finaleps-ss}
\end{align}
Note that $\E[\n{\calG_\eta(\hx)}] \leq \sqrt{\E[\ns{\calG_\eta(\hx)}]} \leq \epsilon$.
The inequality \eqref{eq:randomhx-ss} holds by randomly choose $\hx$ from $\{x_{t-1}\}_{t\in [Sm]}$, and \eqref{eq:finaleps-ss} holds by choosing $Sm\geq \frac{4(\Phi(x_0)-\Phi^*)}{\epsilon^2 \eta}$ and $B\geq \min\{n, \frac{2\sigma^2}{\epsilon^2}\}$.

By choosing $\eta=\frac{1}{(1+\sqrt{(m-1)/b})L}$,
the total number of iterations is
$$
T=Sm= 
\frac{4(\Phi(x_0)-\Phi^*\big)}{\epsilon^2 \eta}
=\frac{4(\Phi(x_0)-\Phi^*\big)(1+\sqrt{(m-1)/b})L}{\epsilon^2},
$$
which is also the number of PO calls.
The number of SFO calls is 
$$
SB+Smb=4L(\Phi(x_0)-\Phi^*)(1+\sqrt{(m-1)/b}) \left(\frac{B}{\epsilon^2 m}+\frac{b}{\epsilon^2}\right).
$$

If we choose $m=b$ (then $\eta\leq \frac{1}{(1+\sqrt{(m-1)/b})L}=\frac{1}{2L}$),
the total number of PO calls  is
$
T= \frac{8L(\Phi(x_0)-\Phi^*)}{\epsilon^2}.
$
The number of SFO calls equals to $Sn+Smb=8L(\Phi(x_0)-\Phi^*)\big(\frac{n}{\epsilon^2 b}+\frac{b}{\epsilon^2}\big)$
if $B=n$ (i.e., the second term in (\ref{eq:randomhx-ss}) is 0 and thus Assumption \ref{asp:bv} is not needed), or equals to
$SB+Smb=8L(\Phi(x_0)-\Phi^*)\big(\frac{B}{\epsilon^2 b}+\frac{b}{\epsilon^2}\big)$
if $B\geq \min\{n, \frac{2\sigma^2}{\epsilon^2}\}$. 

In case the number of SFO calls is less than $B$ (i.e., if the total number of epochs $S<1$), we may add an explicit term $B$ to the SFO result since the algorithm at least uses $B$ SFO calls in the first epoch $s=0$ at Line \ref{line:full} of Algorithm~\ref{alg:ssrgd_hl}. In this situation, SSRGD (Algorithm~\ref{alg:ssrgd_hl}) terminates within the first epoch $s=0$.
\end{proofof}

\section{Missing Proofs for Section \ref{sec:pl} PL Condition}
\label{app:proof-pl}
Now we provide the proofs for ProxSVRG+ (Theorem \ref{thm:pl1}) and SSRGD (Theorem \ref{thm:pl-ssrgd}) under PL condition.

\subsection{Proof for ProxSVRG+ under PL condition}
\begin{proofof}{Theorem \ref{thm:pl1}}
First, we recall a key inequality \eqref{eq:useyoung} from the proof of Theorem \ref{thm:1}, i.e.,
\begin{align}
\E[\Phi(x_t)]
&\leq \E\Big[\Phi(x_{t-1}) -\frac{\eta}{2}\ns{\calG_\eta(x_{t-1})}
-(\frac{1}{2\eta}-\frac{L}{2})\frac{1}{\alpha_t +1}\ns{x_t-\tx^{s}}
 \notag\\
& \qquad \qquad
+\Big((\frac{1}{2\eta}-\frac{L}{2})\frac{1}{\alpha_t}+\frac{\eta L^2}{2b}\Big)\ns{x_{t-1}-\tx^s}
+\frac{I\{B<n\}\eta\sigma^2}{2B}\Big].\notag
\end{align}
Then we plug the PL inequality \eqref{eq:pl}, i.e.,  $\ns{\calG_\eta(x)} \geq 2\mu (\Phi(x)-\Phi^*)$ into it to obtain
\begin{align}
\E[\Phi(x_t)-\Phi^*]
&\leq \E\Big[(1-\mu \eta)(\Phi(x_{t-1})-\Phi^*)
-(\frac{1}{2\eta}-\frac{L}{2})\frac{1}{\alpha_t +1}\ns{x_t-\tx^{s}}
\notag\\
& \qquad \qquad
+\Big((\frac{1}{2\eta}-\frac{L}{2})\frac{1}{\alpha_t}+\frac{\eta L^2}{2b}\Big)\ns{x_{t-1}-\tx^s}
+\frac{I\{B<n\}\eta\sigma^2}{2B}\Big]. \notag
\end{align}
Now, we reorder it as follows:
\begin{align}
&\E\Big[\Phi(x_t)-\Phi^* +(\frac{1}{2\eta}-\frac{L}{2})\frac{1}{\alpha_t +1}\ns{x_t-\tx^{s}} \Big]  \notag\\
&\leq \E\Big[(1-\mu \eta)(\Phi(x_{t-1})-\Phi^*)
+\Big((\frac{1}{2\eta}-\frac{L}{2})\frac{1}{\alpha_t}+\frac{\eta L^2}{2b}\Big)\ns{x_{t-1}-\tx^s}
+\frac{I\{B<n\}\eta\sigma^2}{2B}\Big] \notag\\
&\leq \E\Big[(1-\mu \eta)\Big((\Phi(x_{t-1})-\Phi^*)
+\frac{(\frac{1}{2\eta}-\frac{L}{2})\frac{1}{\alpha_t}+\frac{\eta L^2}{2b}}{1-\mu\eta}\ns{x_{t-1}-\tx^s}\Big)
+\frac{I\{B<n\}\eta\sigma^2}{2B}\Big] \notag\\
&\leq \E\Big[(1-\mu \eta)\Big((\Phi(x_{t-1})-\Phi^*)
+(\frac{1}{2\eta}-\frac{L}{2})\frac{1}{\alpha_{t-1} +1}\ns{x_{t-1}-\tx^s}\Big)
+\frac{I\{B<n\}\eta\sigma^2}{2B}\Big], \label{eq:keypl}
\end{align}
where \eqref{eq:keypl} holds by choosing $\alpha_t$s and $\eta$ to satisfy $(\frac{1}{2\eta}-\frac{L}{2})\frac{1}{\alpha_t}+\frac{\eta L^2}{2b}\leq (\frac{1}{2\eta}-\frac{L}{2})\frac{1-\mu \eta}{\alpha_{t-1} +1}$.
Similar to the proof of Theorem \ref{thm:1}, we can choose $\alpha_t=2(t\%m)-1$ and $\eta\leq \frac{1}{(1+2m/\sqrt{b})L}$.

Telescoping \eqref{eq:keypl} for all iterations $sm+1\leq t \leq sm+m$ in epoch $s$, we have
\begin{align}
&\E[\Phi(x_{(s+1)m})-\Phi^*]  \notag\\
&\leq \E\Big[\Phi(x_{(s+1)m})-\Phi^* +(\frac{1}{2\eta}-\frac{L}{2})\frac{1}{\alpha_{(s+1)m} +1}\ns{x_{(s+1)m}-\tx^{s}} \Big]  \notag\\
&\leq \E\Big[(1-\mu \eta)^m\Big((\Phi(x_{sm})-\Phi^*)
+(\frac{1}{2\eta}-\frac{L}{2})\frac{1}{\alpha_{sm} +1}\ns{x_{sm}-\tx^s}\Big)
 \notag\\
& \qquad \qquad
+\frac{I\{B<n\}\eta\sigma^2}{2B}\sum_{j=0}^{m-1}(1-\mu\eta)^j\Big] \notag\\
&=\E\Big[(1-\mu \eta)^m(\Phi(x_{sm})-\Phi^*)
+\frac{I\{B<n\}\eta\sigma^2}{2B}\frac{(1-(1-\mu\eta)^m)}{\mu\eta}\Big], \label{eq:keypl2}
\end{align}
where the last equation \eqref{eq:keypl2} holds due to $\tx^s = x_{sm}$ (see Line \ref{line:txs} of Algorithm~\ref{alg:1}).

Similarly, we telescope \eqref{eq:keypl2} for all epochs $0\leq s\leq S-1$ to finish the proof:
\begin{align}
&\E[\Phi(x_{Sm})-\Phi^*]  \notag\\
&\leq\E\Big[(1-\mu \eta)^{Sm}(\Phi(x_{0})-\Phi^*)
+\frac{I\{B<n\}\eta\sigma^2}{2B}\frac{(1-(1-\mu\eta)^m)}{\mu\eta}\frac{(1-(1-\mu\eta)^{Sm})}{1-(1-\mu\eta)^m}\Big] \notag\\
&\leq (1-\mu \eta)^{Sm}(\Phi(x_{0})-\Phi^*) + \frac{I\{B<n\}\sigma^2}{2\mu B} \label{eq:finalpl0}\\
&\leq \frac{\epsilon}{2}+\frac{\epsilon}{2} =\epsilon, \label{eq:finalpl1}
\end{align}
where \eqref{eq:finalpl1} holds by choosing $Sm \geq \frac{1}{\mu\eta}\log\frac{2(\Phi(x_0)-\Phi^*)}{\epsilon}$ and $B\geq \min\{n, \frac{\sigma^2}{\mu \epsilon}\}$.

In the following, for simple presentation, we choose $m=\sqrt{b}$ (then $\eta\leq \frac{1}{(1+2m/\sqrt{b})L}=\frac{1}{3L}$). Note that there is no constraint for $m$ and $b$ in our convergence proof.
The total number of iterations is
$$
T=Sm= \frac{1}{\mu\eta}\log\frac{2(\Phi(x_0)-\Phi^*)}{\epsilon}=\frac{3L}{\mu}\log\frac{2(\Phi(x_0)-\Phi^*)}{\epsilon}.
$$
The number of PO calls equals to $T=Sm=\frac{3L}{\mu}\log\frac{2(\Phi(x_0)-\Phi^*)}{\epsilon}$.
The proof is finished since the number of SFO calls equals to $Sn+Smb=\big(\frac{n}{\sqrt{b}}+b\big)\frac{3L}{\mu}\log\frac{2(\Phi(x_0)-\Phi^*)}{\epsilon}$
if $B=n$ (i.e., the second term in (\ref{eq:finalpl0}) is 0 and thus Assumption \ref{asp:bv} is not needed), or equals to
$SB+Smb=\big(\frac{B}{\sqrt{b}}+b\big)\frac{3L}{\mu}\log\frac{2(\Phi(x_0)-\Phi^*)}{\epsilon}$
if $B\geq \min\{n, \frac{\sigma^2}{\mu\epsilon}\}$.
\end{proofof}

\subsection{Proof for SSRGD under PL condition}
\begin{proofof}{Theorem \ref{thm:pl-ssrgd}}
Similar to the proof of Theorem \ref{thm:pl1}, we first recall a key inequality from the proof of Theorem \ref{thm:ssrgd} which combines \eqref{eq:key-ss} and \eqref{eq:varfinite-ss}, i.e.,
\begin{align}
\E[\Phi(x_t)]
&\leq \E\Big[\Phi(x_{t-1}) -\frac{\eta}{2}\ns{\calG_\eta(x_{t-1})}
-(\frac{1}{2\eta}-\frac{L}{2})\ns{x_t-x_{t-1}}
\notag\\
& \qquad \qquad
+\frac{\eta L^2}{2b}\sum_{j=sm+1}^{t-1}\ns{x_{j}-x_{j-1}}
+\frac{I\{B<n\}\eta\sigma^2}{2B}\Big].\notag
\end{align}
Then we plug the PL inequality \eqref{eq:pl}, i.e.,  $\ns{\calG_\eta(x)} \geq 2\mu (\Phi(x)-\Phi^*)$ into it to obtain
\begin{align}
\E[\Phi(x_t)-\Phi^*]
&\leq \E\Big[(1-\mu \eta)(\Phi(x_{t-1})-\Phi^*)
-(\frac{1}{2\eta}-\frac{L}{2})\ns{x_t-x_{t-1}}
\notag\\
& \qquad \qquad
+\frac{\eta L^2}{2b}\sum_{j=sm+1}^{t-1}\ns{x_{j}-x_{j-1}}
+\frac{I\{B<n\}\eta\sigma^2}{2B}\Big].\label{eq:pl-ss1}
\end{align}
We sum it up (\eqref{eq:pl-ss1} $\times \frac{1}{(1-\mu\eta)^k}$ for iteration $t=sm+k$) for all iterations in epoch $s$, i.e., $t=sm+k$ where $k$ from $1$ to $m$:
\begin{align}
\E\Big[\frac{\Phi(x_{(s+1)m})-\Phi^*}{(1-\mu\eta)^m}\Big]
&\leq \E\Big[\Phi(x_{sm})-\Phi^*
-(\frac{1}{2\eta}-\frac{L}{2})\sum_{k=1}^{m}\frac{1}{(1-\mu\eta)^k}\ns{x_{sm+k}-x_{sm+k-1}}
\notag\\
& \qquad \qquad
+\frac{\eta L^2}{2b}\sum_{k=1}^{m}\Big(\frac{1}{(1-\mu\eta)^k}\sum_{j=1}^{k-1}\ns{x_{sm+j}-x_{sm+j-1}}\Big)
\notag\\
& \qquad \qquad
+\frac{I\{B<n\}\eta\sigma^2}{2B}\sum_{k=1}^{m}\frac{1}{(1-\mu\eta)^k}\Big].
\end{align}
Then we deduce it as follows:
\begin{align}
&\E\Big[\Phi(x_{(s+1)m})-\Phi^*\Big]   \notag\\
&\leq \E\Big[(1-\mu\eta)^m(\Phi(x_{sm})-\Phi^*)
-(\frac{1}{2\eta}-\frac{L}{2})\sum_{k=1}^{m}\frac{(1-\mu\eta)^m}{(1-\mu\eta)^k}\ns{x_{sm+k}-x_{sm+k-1}}
\notag\\
& \qquad \qquad
+\frac{\eta L^2}{2b}\sum_{k=1}^{m}\Big(\frac{(1-\mu\eta)^m}{(1-\mu\eta)^k}\sum_{j=1}^{k-1}\ns{x_{sm+j}-x_{sm+j-1}}\Big) \notag\\
& \qquad \qquad
+\frac{I\{B<n\}\eta\sigma^2}{2B}\sum_{k=1}^{m}\frac{(1-\mu\eta)^m}{(1-\mu\eta)^k}\Big] \notag\\
&=\E\Big[(1-\mu\eta)^m(\Phi(x_{sm})-\Phi^*)
-(\frac{1}{2\eta}-\frac{L}{2})\sum_{k=1}^{m}\frac{(1-\mu\eta)^m}{(1-\mu\eta)^k}\ns{x_{sm+k}-x_{sm+k-1}}
\notag\\
& \qquad \qquad
+\frac{\eta L^2}{2b}\sum_{k=1}^{m}\Big(\sum_{j=k+1}^m\frac{(1-\mu\eta)^m}{(1-\mu\eta)^j}\Big)\ns{x_{sm+k}-x_{sm+k-1}}\notag\\
& \qquad \qquad
+\frac{I\{B<n\}\eta\sigma^2}{2B}\sum_{k=1}^{m}\frac{(1-\mu\eta)^m}{(1-\mu\eta)^k}\Big] \notag\\
&\leq \E\Big[(1-\mu\eta)^m(\Phi(x_{sm})-\Phi^*)
-(\frac{1}{2\eta}-\frac{L}{2})\sum_{k=1}^{m}\frac{(1-\mu\eta)^m}{(1-\mu\eta)^k}\ns{x_{sm+k}-x_{sm+k-1}}
\notag\\
& \qquad \qquad
+\frac{\eta L^2 (m-1)}{2b}\sum_{k=1}^{m} \ns{x_{sm+k}-x_{sm+k-1}}
+\frac{I\{B<n\}\eta\sigma^2}{2B}\sum_{k=1}^{m}\frac{(1-\mu\eta)^m}{(1-\mu\eta)^k}\Big] \label{eq:xxxm}\\
&\leq \E\Big[(1-\mu\eta)^m(\Phi(x_{sm})-\Phi^*) 
+\frac{I\{B<n\}\eta\sigma^2}{2B}\sum_{k=1}^{m}\frac{(1-\mu\eta)^m}{(1-\mu\eta)^k} \Big]\label{eq:pl-ssepoch}\\
&\leq \E\Big[(1-\mu\eta)^m(\Phi(x_{sm})-\Phi^*) + \frac{I\{B<n\}\eta\sigma^2}{2B}\frac{(1-(1-\mu\eta)^m)}{\mu\eta} \Big],\label{eq:pl-xxx}
\end{align}
where \eqref{eq:xxxm} uses the fact $\sum_{i=0}^{m-2} (1-\mu\eta)^i \leq \sum_{i=0}^{m-2}1=m-1$ (here $\mu \eta \leq 1$ due to $\mu\leq L$ and $\eta \leq \frac{1}{L}$), 
and \eqref{eq:pl-ssepoch} holds by choosing appropriate $\eta$ to cancel the point distance terms $\ns{x_{sm+k}-x_{sm+k-1}}$. Similar to the proof of Theorem \ref{thm:ssrgd}, we can choose $\eta\leq \frac{1}{(1+\sqrt{(m-1)/b})L}$.

Now, we telescope \eqref{eq:pl-xxx} for all epochs $0\leq s\leq S-1$ to finish the proof:
\begin{align}
&\E[\Phi(x_{Sm})-\Phi^*]  \notag\\
&\leq\E\Big[(1-\mu \eta)^{Sm}(\Phi(x_{0})-\Phi^*)
+\frac{I\{B<n\}\eta\sigma^2}{2B}\frac{(1-(1-\mu\eta)^m)}{\mu\eta}\frac{(1-(1-\mu\eta)^{Sm})}{1-(1-\mu\eta)^m}\Big] \notag\\
&\leq (1-\mu \eta)^{Sm}(\Phi(x_{0})-\Phi^*) + \frac{I\{B<n\}\sigma^2}{2\mu B} \label{eq:finalpl0-ss}\\
&\leq \frac{\epsilon}{2}+\frac{\epsilon}{2} =\epsilon, \label{eq:finalpl1-ss}
\end{align}
where \eqref{eq:finalpl1-ss} holds by choosing $Sm \geq \frac{1}{\mu\eta}\log\frac{2(\Phi(x_0)-\Phi^*)}{\epsilon}$ and $B\geq \min\{n, \frac{\sigma^2}{\mu \epsilon}\}$.

In the following, for simple presentation, we choose $m=b$ (then $\eta\leq \frac{1}{(1+\sqrt{(m-1)/b})L}=\frac{1}{2L}$). Note that there is no constraint for $m$ and $b$ in our convergence proof.
The total number of iterations is
$$
T=Sm= \frac{1}{\mu\eta}\log\frac{2(\Phi(x_0)-\Phi^*)}{\epsilon}=\frac{2L}{\mu}\log\frac{2(\Phi(x_0)-\Phi^*)}{\epsilon}.
$$
The number of PO calls equals to $T=Sm=\frac{2L}{\mu}\log\frac{2(\Phi(x_0)-\Phi^*)}{\epsilon}$.
The proof is finished since the number of SFO calls equals to $Sn+Smb=\big(\frac{n}{b}+b\big)\frac{2L}{\mu}\log\frac{2(\Phi(x_0)-\Phi^*)}{\epsilon}$
if $B=n$ (i.e., the second term in (\ref{eq:finalpl0-ss}) is 0 and thus Assumption \ref{asp:bv} is not needed), or equals to
$SB+Smb=\big(\frac{B}{b}+b\big)\frac{2L}{\mu}\log\frac{2(\Phi(x_0)-\Phi^*)}{\epsilon}$
if $B\geq \min\{n, \frac{\sigma^2}{\mu\epsilon}\}$.
\end{proofof}

\section{Missing Proofs for Section \ref{sec:localmin} Local Minima}
\label{app:proof-localmin}

Now, we provide the detailed proofs for Theorem \ref{thm:ssrgd-lm}.
Note that due to the second-order guarantee, the proofs of the finite-sum case and the online case 
have more difference than previous first-order guarantee methods (e.g., proof of Theorem \ref{thm:1} and \ref{thm:ssrgd}).
One of the reason is that for the perturbation condition $\n{v_{sm}}\leq \mathG$ in Line \ref{line:super} of Algorithm~\ref{alg:ssrgd}, $v_{sm}=\nabla f(x_{sm})$ for finite-sum case ($B=n$) while $v_{sm} =\frac{1}{B}\sum_{j\in I_B}\nabla f_j(x_{sm})$ for the online case. So we need an extra high probability bound $\n{\nabla f(x_{sm})} \leq \mathG$ in the online case.
In the following, we divide the proof of Theorem \ref{thm:ssrgd-lm} into two parts, i.e., finite-sum (Section \ref{app:ssrgd-lm-finite}) and online (Section \ref{app:ssrgd-lm-online}).
Before the proof, we recall some standard concentration bounds in Section \ref{app:lm-tools}.

\subsection{Tools}
\label{app:lm-tools}
Here, we recall some classical concentration bounds for matrices and vectors.
\begin{proposition} [Bernstein Inequality \citep{tropp2012user}]\label{prop:bernstein_original}
	Consider a finite sequence $\{Z_k\}$ of independent, random matrices with dimension $d_1\times d_2$. Assume that each random matrix satisfies
	\begin{align*}
	\E[Z_k]=0  ~~and~~ \n{Z_k}\leq R ~~almost ~surely.
	\end{align*}
	Define
	$$\sigma^2:= \max\Big\{\big\|\sum_k \E[Z_k Z_k^T]\big\|, \big\| \sum_k \E[Z_k^T Z_k]\big\| \Big\}.$$
	Then, for all $t\geq 0$,
	$$\pr\Big\{\big\|\sum_k Z_k\big\|\geq t \Big\}\leq (d_1+d_2) \exp\Big(\frac{-t^2/2}{\sigma^2+Rt/3}\Big).$$
\end{proposition}
In our proof, we only need its special case vector version as follows, where $z_k=v_k-\E[v_k]$.
\begin{proposition} [Bernstein Inequality \citep{tropp2012user}]\label{prop:bernstein}
	Consider a finite sequence $\{v_k\}$ of independent, random vectors with dimension $d$. Assume that each random matrix satisfies
	\begin{align*}
	\n{v_k-\E[v_k]}\leq R ~~almost ~surely.
	\end{align*}
	Define
	$$\sigma^2 := \sum_{k}\E\ns{v_k-\E[v_k]}.$$
	Then, for all $t\geq 0$,
	$$\pr\Big\{\big\|\sum_k (v_k-\E[v_k])\big\|\geq t \Big\}\leq (d+1) \exp\Big(\frac{-t^2/2}{\sigma^2+Rt/3}\Big).$$
\end{proposition}

Moreover, we also need the following martingale concentration bounds, i.e., Azuma-Hoeffding inequality.
Now, we only state the vector version (the more general matrix version is not needed).
\begin{proposition} [Azuma-Hoeffding Inequality \citep{hoeffding1963probability,tropp2011user}]\label{prop:azuma}
	Consider a martingale vector sequence $\{y_k\}$ with dimension $d$, and let $\{z_k\}$ denote the associated martingale difference sequence with respect to a filtration $\{\mathscr{F}_k\}$, i.e., $z_k:=y_k-\E[y_k|\mathscr{F}_{k-1}]=y_k-y_{k-1}$ and $\E[z_k|\mathscr{F}_{k-1}]=0$.
	Suppose that $\{z_k\}$ satisfies
	\begin{align}
	\n{z_k}=\n{y_k-y_{k-1}} \leq c_k ~~almost ~surely. \label{eq:diffbound}
	\end{align}
	Then, for all $t\geq 0$,
	$$\pr\Big\{\|y_k-y_0\|\geq t \Big\}\leq (d+1) \exp\Big(\frac{-t^2}{8\sum_{i=1}^k c_i^2}\Big).$$
\end{proposition}

However, the assumption that $\n{z_k} \leq c_k$ in \eqref{eq:diffbound} with probability 1 
is too strict and it may fail sometimes.
Fortunately, the Azuma-Hoeffding inequality also holds with a slackness if $\n{z_k} \leq c_k$ with high probability.

\begin{proposition}{\bf(Azuma-Hoeffding Inequality with High Probability \citep{chung2006concentration, tao2015random})}\label{prop:azumahigh}
	Con\-sider a martingale vector sequence $\{y_k\}$ with dimension $d$, and let $\{z_k\}$ denote the associated martingale difference sequence with respect to a filtration $\{\mathscr{F}_k\}$, i.e., $z_k:=y_k-\E[y_k|\mathscr{F}_{k-1}]=y_k-y_{k-1}$ and $\E[z_k|\mathscr{F}_{k-1}]=0$.
	Suppose that $\{z_k\}$ satisfies
	\begin{align*}
	\n{z_k}=\n{y_k-y_{k-1}} \leq c_k ~~with~ high~ probability~ 1-\zeta_k.
	\end{align*}
	Then, for all $t\geq 0$,
	$$\pr\Big\{\|y_k-y_0\|\geq t \Big\}\leq (d+1) \exp\Big(\frac{-t^2}{8\sum_{i=1}^k c_i^2}\Big)+\sum_{i=1}^k\zeta_k.$$
\end{proposition}

\subsection{Proof of Theorem \ref{thm:ssrgd-lm} (finite-sum)}
\label{app:ssrgd-lm-finite}

For proving the second-order guarantee, we divide the proof into two situations.
The first situation (\textbf{large gradients}) is almost the same as the above arguments for first-order guarantee, where the function value decreases significantly since the gradients are large (see \eqref{eq:eta1-ss}).
For the second situation (\textbf{around saddle points}), we show that the function value can also decrease a lot by adding a random perturbation. The reason is that saddle points are usually unstable and the stuck region is relatively small in a random perturbation ball.

\vspace{1mm}
\noindent{{\bf Large Gradients}: }
First, we need a high probability bound for the variance term instead of the expectation one \eqref{eq:varfinite-ss} (note that here $B=n$ in the finite-sum case). Then we use it to get a high probability bound of \eqref{eq:eta1-ss} for the decrease of the function value.
Recall that $v_k=\frac{1}{b}\sum_{i\in I_b}\big(\nabla f_i(x_{k})-\nabla f_i(x_{k-1})\big) + v_{k-1}$ (see Line \ref{line:up2} of Algorithm~\ref{alg:ssrgd}). We let $y_k:=v_k-\nabla f(x_k)$ and $z_k:=y_k-y_{k-1}$.
It is not hard to verify that $\{y_k\}$ is a martingale sequence and $\{z_k\}$ is the associated martingale difference sequence.
In order to apply the Azuma-Hoeffding inequalities to get a high probability bound, we first need to bound the 
martingale difference sequence $\{z_k\}$.
We use the Bernstein inequality to bound the differences as follows.
\begin{align}
z_k=y_k-y_{k-1}&= v_k-\nabla f(x_k) - (v_{k-1}-\nabla f(x_{k-1})) \notag\\
&=\frac{1}{b}\sum_{i\in I_b}\big(\nabla f_i(x_{k})-\nabla f_i(x_{k-1})\big) + v_{k-1}
-\nabla f(x_k) - (v_{k-1}-\nabla f(x_{k-1})) \notag \\
&=\frac{1}{b}\sum_{i\in I_b}\Big(\nabla f_i(x_{k})-\nabla f_i(x_{k-1})
-(\nabla f(x_k) -\nabla f(x_{k-1}))\Big)\notag\\
&=\frac{1}{b}\sum_{i\in I_b}u_i, \label{eq:zk}
\end{align}
where we define $u_i:=\nabla f_i(x_{k})-\nabla f_i(x_{k-1})-(\nabla f(x_k) -\nabla f(x_{k-1}))$ in \eqref{eq:zk}. 
Then we have
\begin{align}
\|u_i\|=\|\nabla f_i(x_{k})-\nabla f_i(x_{k-1})-(\nabla f(x_k) -\nabla f(x_{k-1}))\|\leq 2L\|x_{k}-x_{k-1}\|, \label{eq:b1}
\end{align}
where the last inequality holds due to the gradient Lipschitz Assumption \ref{asp:smoothgandh}.
Then, consider the variance term
\begin{align}
\E\Big[\sum_{i\in I_b}\|u_i\|^2\Big] 
&=b\E_i[\ns{\nabla f_i(x_{k})-\nabla f_i(x_{k-1})-(\nabla f(x_k) -\nabla f(x_{k-1}))}] \notag\\
&\leq b\E_i[\ns{\nabla f_i(x_{k})-\nabla f_i(x_{k-1})}] \notag\\
&\leq bL^2\ns{x_{k}-x_{k-1}}, \label{eq:b2}
\end{align}
where the first inequality uses the fact $\E[\ns{x-\E x}]\leq \E[\ns{x}]$, and the last inequality uses the gradient Lipschitz Assumption \ref{asp:smoothgandh}.
According to \eqref{eq:b1} and \eqref{eq:b2}, we can bound the difference $z_k$ by Bernstein inequality (Proposition \ref{prop:bernstein}) as
\begin{align*}
\pr\Big\{\big\|z_k\big\|\geq \frac{t}{b} \Big\} &\leq (d+1) \exp\Big(\frac{-t^2/2}{\E[\sum_{i\in I_b}\|u_i\|^2] +Rt/3}\Big) \notag \\
& = (d+1) \exp\Big(\frac{-t^2/2}{bL^2\ns{x_{k}-x_{k-1}}+ 2L\|x_{k}-x_{k-1}\|t/3}\Big)
\notag \\
& = \zeta_k,
\end{align*}
where the last equality holds by letting $t=CL\sqrt{b}\n{x_{k}-x_{k-1}}$, where $C=O(\log\frac{d}{\zeta_k})$.
Now, we have a high probability bound for the difference sequence $\{z_k\}$, i.e.,
\begin{align}
\label{eq:zkbound}
\|z_k\| \leq  \frac{CL}{\sqrt{b}}\n{x_{k}-x_{k-1}} \quad \mathrm{~with~probability~} 1-\zeta_k.
\end{align}

Now, we are ready to get a high probability bound for our original variance term \eqref{eq:varfinite-ss} by using the martingale Azuma-Hoeffding inequality.
Consider in a specific epoch $s$, i.e, iterations $t$ from $sm+1$ to current $sm+k$, where $k$ is less than $m$ (note that we only need to consider the current epoch since each epoch we start with $y=0$). We use a union bound for the difference sequence $\{z_t\}$ by letting $\zeta_k = \zeta'/m$ such that
\begin{align}
\|z_t\|\leq c_t=\frac{CL}{\sqrt{b}}\n{x_{t}-x_{t-1}} \mathrm{~~for~ all~} t\in[sm+1,sm+k] \mathrm{~~with~ probability~~} 1-\zeta'.
\end{align}
Define $\beta:=\sqrt{8\sum_{t=sm+1}^{sm+k} c_t^2\log\frac{d}{\zeta'}}
=\frac{C'L}{\sqrt{b}}\sqrt{\sum_{t=sm+1}^{sm+k}\ns{x_{t}-x_{t-1}}}$, where $C'=O(C\sqrt{\log\frac{d}{\zeta'}})=O(\log\frac{d}{\zeta_k}\sqrt{\log\frac{d}{\zeta'}})=O(\log\frac{dm}{\zeta'}\sqrt{\log\frac{d}{\zeta'}})=\tdo(1)$.
According to Azuma-Hoeffding inequality (Proposition \ref{prop:azumahigh}) and noting that $\zeta_k = \zeta'/m$, we have that
\begin{align*}
\pr\Big\{\big\|y_{sm+k}-y_{sm}\big\|\geq \beta \Big\} &\leq (d+1) \exp\Big(\frac{-\beta^2}{8\sum_{t=sm+1}^{sm+k} c_t^2}\Big)+\zeta' 
 = 2\zeta'.
\end{align*}
Recall that $y_k:=v_k-\nabla f(x_k)$ and at the beginning point of this epoch $y_{sm}=0$ due to $v_{sm}=\nabla f(x_{sm})$ since $B=n$ in this finite-sum case (see Line \ref{line:up1} of Algorithm~\ref{alg:ssrgd}). Thus, for any $t\in [sm+1,sm+m]$, we have that
\begin{align}\label{eq:highvar}
\n{v_{t-1}-\nabla f(x_{t-1})}=\n{y_{t-1}} \leq 
\beta := \frac{C'L}{\sqrt{b}}\sqrt{\sum_{j=sm+1}^{t-1}\ns{x_{j}-x_{j-1}}},
\end{align}
holds with probability $1-2\zeta'$, where $C'=O(\log\frac{dm}{\zeta'}\sqrt{\log\frac{d}{\zeta'}})=\tdo(1)$.

Now, we use this high probability version \eqref{eq:highvar} instead of the expectation one \eqref{eq:varfinite-ss} to obtain the high probability bound for the decrease of the function value (see \eqref{eq:eta1-ss}).
We sum up \eqref{eq:key-ss} from the beginning of this epoch $s$, i.e., iterations from $sm+1$ to $t$, by plugging \eqref{eq:highvar} into them to get:
\begin{align}
f(x_{t}) &\leq f(x_{sm}) - \frac{\eta}{2}\sum_{j=sm+1}^{t}\ns{\nabla f(x_{j-1})}
- \big(\frac{1}{2\eta}- \frac{L}{2}\big)\sum_{j=sm+1}^{t}\ns{x_j-x_{j-1}} \notag\\
&\qquad + \frac{\eta}{2}\sum_{k=sm+1}^{t-1}\frac{C'^2L^2\sum_{j=sm+1}^{k}\ns{x_{j}-x_{j-1}}}{b} \label{eq:plughighvar}\\
&\leq f(x_{sm}) - \frac{\eta}{2}\sum_{j=sm+1}^{t}\ns{\nabla f(x_{j-1})}
- \big(\frac{1}{2\eta}- \frac{L}{2}\big)\sum_{j=sm+1}^{t}\ns{x_j-x_{j-1}} \notag\\
&\qquad + \frac{\eta C'^2L^2}{2b}\sum_{k=sm+1}^{t-1}\sum_{j=sm+1}^{k}\ns{x_{j}-x_{j-1}} \notag\\
&\leq f(x_{sm}) - \frac{\eta}{2}\sum_{j=sm+1}^{t}\ns{\nabla f(x_{j-1})}
- \big(\frac{1}{2\eta}- \frac{L}{2}\big)\sum_{j=sm+1}^{t}\ns{x_j-x_{j-1}} \notag\\
&\qquad + \frac{\eta C'^2L^2(t-1-sm)}{2b}\sum_{j=sm+1}^{t}\ns{x_{j}-x_{j-1}} \notag\\
&\leq f(x_{sm}) - \frac{\eta}{2}\sum_{j=sm+1}^{t}\ns{\nabla f(x_{j-1})}
- \big(\frac{1}{2\eta}- \frac{L}{2}-\frac{\eta C'^2L^2}{2}\big)\sum_{j=sm+1}^{t}\ns{x_j-x_{j-1}}  \label{eq:bm2}\\
&\leq f(x_{sm}) - \frac{\eta}{2}\sum_{j=sm+1}^{t}\ns{\nabla f(x_{j-1})} \label{eq:eta2},
\end{align}
where \eqref{eq:bm2} holds if the minibatch size $b\geq m$ (note that here $t\leq (s+1)m$), and
\eqref{eq:eta2} holds if the step size $\eta\leq \frac{1}{(1+C')L}$, where $C'=O(\log\frac{dm}{\zeta'}\sqrt{\log\frac{d}{\zeta'}})$.
Note that \eqref{eq:plughighvar} uses \eqref{eq:highvar} which holds with probability $1-2\zeta'$. Thus by a union bound, we know that \eqref{eq:eta2} holds with probability at least $1-2m\zeta'$.

Note that \eqref{eq:eta2} only guarantees that the function value decreases significantly only
when the summation of gradients in this epoch is large. However, in order to connect the guarantees between first situation (large gradients) and second situation (around saddle points), we need to show guarantees that are related to the \emph{gradient of the starting point} of each epoch (see Line \ref{line:super} of Algorithm~\ref{alg:ssrgd}).
Similar to \citet{ge2019stable}, we achieve this by stopping the epoch at a uniformly random point (see Line \ref{line:randomstop} of Algorithm~\ref{alg:ssrgd}).

Now we prove Lemma \ref{lem:first} to distinguish these two situations (large gradients and around saddle points):

\begin{lemma}[Two Situations]
	\label{lem:first}
	For any epoch $s$, let $x_t$ be a point uniformly sampled from this epoch $\{x_{j} \}_{j=sm+1}^{(s+1)m}$.
	We choose the step size $\eta \leq \frac{1}{(1+C')L}$ (where $C'=O(\log\frac{dm}{\zeta'}\sqrt{\log\frac{d}{\zeta'}})=\tdo(1)$) and the minibatch size $b\geq m$.
	Then for any $\mathG>0$, either of the following two cases happens:
	\begin{enumerate}
		\item (Small gradient, possibly around a saddle point) 
		If at least half of points in this epoch have gradient norm no larger than $\mathG$, then $\|\nabla f(x_t) \| \le \mathG$ holds with probability at least $1/2$;
		\item (Large gradient) Otherwise, we know $f(x_{sm}) - f(x_t) \ge \frac{\eta m\mathG^2}{8}$ holds with probability at least $1/5.$
	\end{enumerate}
	Moreover, $f(x_t) \le f(x_{sm})$ holds with high probability $1-2m\zeta'$ no matter which case happens.
\end{lemma}

\begin{proofof}{Lemma~\ref{lem:first}}
	There are two cases in this epoch:
	\begin{enumerate}
		\item If at least half of points in this epoch $\{x_{j} \}_{j=sm+1}^{(s+1)m}$ have gradient norm no larger than $\mathG$, then it is easy to see that a uniformly sampled point $x_t$ has gradient norm $\n{\nabla f(x_t)}\leq \mathG$ with probability at least $1/2.$
		\item Otherwise, at least half of points have gradient norm larger than $\mathG$. Then, as long as the sampled point $x_t$ falls into the last quarter of $\{x_{j} \}_{j=sm+1}^{(s+1)m}$, we know $\sum_{j=sm+1}^{t}\ns{\nabla f(x_{j-1})}\geq \frac{m\mathG^2}{4}$. This holds with probability at least $1/4$ since $x_t$ is uniformly sampled. Then combining with \eqref{eq:eta2}, i.e.,  $f(x_{sm}) - f(x_t) \geq \frac{\eta}{2}\sum_{j=sm+1}^{t}\ns{\nabla f(x_{j-1})}$, we can see that the function value decreases $f(x_{sm}) - f(x_t) \ge \frac{\eta m\mathG^2}{8}$. Note that
		\eqref{eq:eta2} holds with high probability $1-2m\zeta'$ if we choose the minibatch size $b\geq m$ and the step size $\eta\leq \frac{1}{(1+C')L}$.
		By a union bound, the function value decrease $f(x_{sm}) - f(x_t) \ge \frac{\eta m\mathG^2}{8}$ with probability at least $1/5$ (e.g., choose $\zeta'\leq 1/40m$).
	\end{enumerate}
	Again according to \eqref{eq:eta2}, $f(x_t)\leq f(x_{sm})$ holds with high probability $1-2m\zeta'$.
\end{proofof}

Note that if Case 2 happens, the function value would decrease significantly in this epoch $s$ (corresponding to the first situation large gradients). Otherwise if Case 1 happens, we know the starting point of the next epoch $x_{(s+1)m}=x_t$ (i.e., Line \ref{line:randompoint} of Algorithm~\ref{alg:ssrgd}), and we know $\n{\nabla f(x_{(s+1)m})}=\n{\nabla f(x_{t})} \leq \mathG$. In this case, we start a super epoch (corresponding to the second situation around saddle points). Note that if $\lambda_{\min}(\nabla^2 f(x_{(s+1)m}))> -\delta$, the point $x_{(s+1)m}$ is already an $(\epsilon,\delta)$-local minimum.

\vspace{3mm}
\noindent{{\bf Around Saddle Points} $\n{\nabla f(\tx)} \leq \mathG$ and $\lambda_{\min}(\nabla^2 f(\tx))\leq -\delta$: }
In this situation, we show that the function value decreases significantly in a \emph{super epoch} with high probability.
Recall that we add a random perturbation at the initial point $\tx$. To simplify the presentation, we use $x_0:=\tx+\xi$ to denote the starting point of the super epoch after the perturbation, where $\xi$ uniformly $\sim \mathbb{B}_0(r)$ and the perturbation radius is $r$ (see Line \ref{line:init} of Algorithm~\ref{alg:ssrgd}).
We follow the \emph{two-point analysis} developed in \citet{jin2017escape}.
The high level idea is as follows:
one can divide the
perturbation ball $\mathbb{B}_0(r)$ into two disjoint regions: (1) an escaping region which consists of all the points
whose function value decreases by at least $\mathf$ after $\mathT$ steps; (2) the rest which we call the stuck region.
The key insight in \citet{jin2017escape} is that the stuck region occupies a very small proportion of the volume of perturbation ball. In particular, they show that the stuck region looks like ``thin pancake" (see Figure 1 and 2 in \citet{jin2017escape}).
Let $e_1$ be the smallest eigenvector direction of Hessian $\hess := \nabla^2 f(\tx)$
For any two points along $e_1$ direction that are not very close, one can show 
at least one of them must not be in the stuck regoin. This implies that the intersection of the line along $e_1$ direction and the stuck region can be at most an interval of a small length, which indicates that the pancake is thin in the 
$e_1$ direction, which can be turned into an upper bound on the volume of the stuck region by standard calculus.
Since we use a more involved update rule, our analysis is somewhat more technical. 

In particular, we consider two coupled points $x_0$ and $x_0'$ with $w_0:=x_0-x_0'=r_0e_1$, where $r_0$ is a scalar and $e_1$ denotes the smallest eigenvector direction of Hessian $\hess := \nabla^2 f(\tx)$. 
Then we get two coupled sequences $\{x_t\}$ and $\{x_t'\}$ by running SSRGD update steps (Line \ref{line:up1}--\ref{line:up2} of Algorithm~\ref{alg:ssrgd}) with the same choice of minibatches (i.e., $I_b$'s in Line \ref{line:up2} of Algorithm~\ref{alg:ssrgd}) for a super epoch.
We show that the function value decreases significantly for at least one of these two coupled sequences (escape the saddle point), i.e.,
\begin{align}\label{eq:funcd}
\exists t\leq\mathT, \mathrm{~~such ~that~~} \max\{f(x_0)-f(x_t), f(x_0')-f(x_t')\} \geq 2\mathf.
\end{align}

Now, we prove \eqref{eq:funcd} by contradiction. Assume the contrary, $f(x_0)-f(x_t)<2\mathf$ and $f(x_0')-f(x_t')<2\mathf$.
First, we show that if function value does not decrease a lot, then all iteration points are not far from the starting point with high probability.
Then we show at least one of $x_t$ and $x_t'$ should go far away from their starting point $x_0$ and $x_0'$ with high probability, rendering a contradiction.
We need the following two technical lemmas and their proofs are deferred to the end of this section.

\begin{lemma}[Localization]
	\label{lem:local}
	Let $\{x_t\}$ denote the sequence by running SSRGD update steps (Line \ref{line:up1}--\ref{line:up2} of Algorithm~\ref{alg:ssrgd}) from $x_0$.
	Moreover, let the step size $\eta\leq \frac{1}{(1+2C')L}$ and minibatch size $b\geq m$. With probability $1-2t\zeta'$, we have
	\begin{align}\label{eq:local}
	\forall t\geq 0,~~ \n{x_t-x_0}\leq \sqrt{\frac{4t(f(x_0)-f(x_t))}{C'L}},
	\end{align}
	where $C'=O(\log\frac{dm}{\zeta'}\sqrt{\log\frac{d}{\zeta'}})=\tdo(1)$.
\end{lemma}

\begin{lemma}[Small Stuck Region]
	\label{lem:smallstuck}
	If the initial point $\tx$ satisfies $-\gamma:=\lambda_{\min}(\nabla^2 f(\tx))\leq -\delta$,
	then let $\{x_t\}$ and $\{x_t'\}$ be two coupled sequences by running SSRGD update steps (Line \ref{line:up1}--\ref{line:up2} of Algorithm~\ref{alg:ssrgd}) with the same choice of minibatches (i.e., $I_b$'s in Line \ref{line:up2} of Algorithm~\ref{alg:ssrgd}) from $x_0$ and $x_0'$ with $w_0:=x_0-x_0'=r_0e_1$, where $x_0\in\mathbb{B}_{\tx}(r)$, $x_0'\in\mathbb{B}_{\tx}(r)$ , $r_0=\frac{\zeta' r}{\sqrt{d}}$ and $e_1$ denotes the smallest eigenvector direction of Hessian $\nabla^2 f(\tx)$.
	Moreover, let the super epoch length $\mathT=\frac{\log(\frac{8\delta\sqrt{d}}{\rho r\zeta'})}{\eta\delta}=\tdo(\frac{1}{\eta\delta})$, the step size $\eta\leq  \frac{1}{15(1+\log \mathT) C'L}=\tdo(\frac{1}{L})$, minibatch size $b\geq m$ and
	the perturbation radius $r\leq \frac{\delta}{C_1 \rho}$.
	With probability $1-2T\zeta'$, we have
	\begin{align}\label{eq:stuck}
	\exists T\leq \mathT,~~ \max\{\n{x_T-x_0}, \n{x_T'-x_0'}\}\geq \frac{\delta}{C_1\rho},
	\end{align}
	where $C'=O(\log\frac{dm}{\zeta'}\sqrt{\log\frac{d}{\zeta'}})=\tdo(1)$
	and $C_1\geq 1+48C' \log(\frac{8\delta\sqrt{d}}{\rho r \zeta'})=\tdo(1)$.
\end{lemma}

Based on these two lemmas, we are ready to show that \eqref{eq:funcd} holds with high probability. Without loss of generality, we assume $\n{x_T-x_0} \geq \frac{\delta}{C_1\rho}$ in \eqref{eq:stuck} (note that \eqref{eq:local} holds for both $\{x_t\}$ and $\{x_t'\}$). Then plugging it into \eqref{eq:local}, we obtain
\begin{align}
\sqrt{\frac{4T(f(x_0)-f(x_T))}{C'L}}&\geq \frac{\delta}{C_1\rho} \notag\\
f(x_0)-f(x_T) &\geq \frac{C'L\delta^2}{4C_1^2\rho^2T} \notag\\
&\geq \frac{ C'L\eta\delta^3}{4C_1^2\rho^2\log(\frac{8\delta\sqrt{d}}{\rho r\zeta'})} \notag\\
&=\frac{\delta^3}{C_1'\rho^2} \label{eq:largeeta}\\
&\overset{\mathrm{def}}{=}2\mathf,\notag
\end{align}
where the last inequality is due to $T\leq \mathT:=\frac{\log(\frac{8\delta\sqrt{d}}{\rho r\zeta'})}{\eta\delta}$, \eqref{eq:largeeta} holds by letting
$C_1'=\frac{4C_1^2\log(\frac{8\delta\sqrt{d}}{\rho r\zeta'})}{C'L\eta} =\tdo(1)$, and the last equality is due to the definition of $\mathf:=\frac{\delta^3}{2C_1'\rho^2}=\tdo(\frac{\delta^3}{\rho^2})$.
Thus, we have already proved that at least one of sequences $\{x_t\}$ and $\{x_t'\}$
escapes the saddle point with probability 
$1-4T\zeta'$ (by union bound of \eqref{eq:local} and \eqref{eq:stuck}), i.e.,
\begin{align}
\exists T\leq \mathT,~~ \max\{f(x_0)-f(x_T), f(x_0')-f(x_T')\} \geq 2\mathf,
\end{align}
if their starting points $x_0$ and $x_0'$ satisfying $w_0:=x_0-x_0'=r_0e_1$.
Now, using the same argument as in \citet{jin2017escape}, we know that in the random perturbation ball, the stuck points can only be a short interval in each line along the $e_1$ direction, i.e.,
at least one of two points in the $e_1$ direction would escape the saddle point if their distance is larger than $r_0=\frac{\zeta' r}{\sqrt{d}}$.
Thus, we know that the probability of the starting point $x_0=\tx+\xi$ (where $\xi$ uniformly $\sim \mathbb{B}_0(r)$) located in the stuck region is less than
\begin{align}\label{eq:goodx0}
\frac{r_0V_{d-1}(r)}{V_d(r)}=
\frac{r_0\Gamma(\frac{d}{2}+1)}{\sqrt{\pi}r\Gamma(\frac{d}{2}+\frac{1}{2})}
\leq \frac{r_0}{\sqrt{\pi}r}\big(\frac{d}{2}+1\big)^{1/2}
\leq \frac{r_0\sqrt{d}}{r}=\zeta',
\end{align}
where $V_d(r)$ denotes the volume of a Euclidean ball with radius $r$ in $d$ dimension,
and the first inequality holds due to Gautschi's inequality.
By a union bound for \eqref{eq:largeeta}  and \eqref{eq:goodx0} ($x_0$ is not in a stuck region), we know that
\begin{align}\label{eq:escape0}
f(x_0)-f(x_T) \geq 2\mathf=\frac{\delta^3}{C_1'\rho^2}
\end{align}
holds with probability $1-(4T+1)\zeta'$.
Note that the initial point of this super epoch is $\tx$ before the perturbation (see Line \ref{line:init} of Algorithm~\ref{alg:ssrgd}), thus we need to show that the perturbation step $x_0=\tx+\xi$ (where $\xi$ uniformly $\sim \mathbb{B}_0(r)$) does not increase the function value a lot, i.e.,
\begin{align}
f(x_0)&\leq f(\tx) +\inner{\nabla f(\tx)}{x_0-\tx}
+ \frac{L}{2}\ns{x_0-\tx}  \notag \\
&\leq f(\tx) +\n{\nabla f(\tx)}\n{x_0-\tx}
+ \frac{L}{2}\ns{x_0-\tx}  \notag\\
&\leq f(\tx) +\mathG \cdot r +\frac{L}{2}r^2 \notag\\
&\leq f(\tx) + \frac{\delta^3}{2C_1'\rho^2} \notag\\
&= f(\tx) +\mathf, \label{eq:perturbless}
\end{align}
where the last inequality holds by letting the perturbation radius $r\leq \min\{\frac{\delta^3}{4C_1'\rho^2\mathG}, \sqrt{\frac{\delta^3}{2C_1'\rho^2L}}\}$.

Now we combine with \eqref{eq:escape0} and \eqref{eq:perturbless} to obtain that
\begin{align}\label{eq:escapehigh}
f(\tx)-f(x_T)=f(\tx)-f(x_0)+f(x_0)-f(x_T) \geq -\mathf+2\mathf=\frac{\delta^3}{2C_1'\rho^2}
\end{align}
holds with probability at least $1-(4T+1)\zeta' \geq 1-5\mathT\zeta'$,
where $C_1'=\tdo(1)$.

Thus we have finished the proof for the second situation (around saddle points), i.e., we show that the function value decreases a lot ($\mathf=\frac{\delta^3}{2C_1'\rho^2}$) in a \emph{super epoch} (recall that $T\leq \mathT=\frac{\log(\frac{8\delta\sqrt{d}}{\rho r\zeta'})}{\eta\delta}$).

\vspace{3mm}
\noindent{{\bf Combing these two situations (large gradients and around saddle points) to prove Theorem \ref{thm:ssrgd-lm}:}}
Now, we prove the theorem by distinguishing the epochs into three types as follows:
	\begin{enumerate}
		\item \emph{Type-1 useful epoch}: If at least half of points in this epoch have gradient norm larger than $\mathG$ (Case 2 of Lemma \ref{lem:first});
		\item \emph{Wasted epoch}: If at least half of points in this epoch have gradient norm no larger than $\mathG$ and the starting point of the next epoch has gradient norm larger than $\mathG$ (it means that in this epoch one can
		not guarantee a significant decrease of the function value as in the large gradients situation, and 
		it does not lead to a super epoch (the second situation) since the starting point of the next epoch has gradient norm larger than $\mathG$);
		\item \emph{Type-2 useful super epoch}: If at least half of points in this epoch have gradient norm no larger than $\mathG$ and the starting point of the next epoch (here we denote this point as $\tx$) has gradient norm no larger than $\mathG$ (i.e., $\n{\nabla f(\tx)}\leq \mathG$) (Case 1 of Lemma \ref{lem:first}), according to Line \ref{line:super} of Algorithm~\ref{alg:ssrgd}, we start a super epoch. So here we denote this epoch along with its following super epoch as a type-2 useful super epoch.
	\end{enumerate}
	First, it is easy to see that the probability of a wasted epoch happened is less than $1/2$ due to the random stop (see Case 1 of Lemma \ref{lem:first}). Note for different wasted epochs, returned points are independently sampled.
	Thus, with high probability $1-\zeta'$, there are at most $\log\frac{1}{\zeta'}=\tdo(1)$ wasted epochs happened before a type-1 useful epoch or type-2 useful super epoch.
	Now, we use $N_1$ and $N_2$ to denote the number of type-1 useful epochs and type-2 useful super epochs that the algorithm is needed. 
	Also recall that the function value always does not increase with high probability $1-2m\zeta'$ for any epoch (see Lemma \ref{lem:first}).
	
	For type-1 useful epoch, according to Case 2 of Lemma \ref{lem:first}, we know that the function value decreases at least $\frac{\eta m\mathG^2}{8}$ with probability at least $1/5$.
	Using a union bound, we know that with probability $1-4N_1/5$, $N_1$ type-1 useful epochs will decrease the function value at least $\frac{\eta m\mathG^2N_1}{40}$.
	 Note that the function value can decrease at most $\Delta_0:= f(x_0)-f^*$ and also recall that the function value always does not increase with high probability $1-2m\zeta'$ for any epoch (see Lemma \ref{lem:first}).
	So let $\frac{\eta m\mathG^2N_1}{40}\leq \Delta_0$, we get $N_1\leq \frac{40\Delta_0}{\eta m\mathG^2}$ with probability at least $1-\tdo(N_1m\zeta')$ by a union bound. We can let $\zeta'\leq \tdo(1/N_1m)$.
	
	For type-2 useful super epoch, first we know that the starting point of the super epoch $\tx$ has gradient norm $\n{\nabla f(\tx)}\leq \mathG$. Now if $\lambda_{\min}(\nabla^2 f(\tx))\geq -\delta$, then $\tx$ is already a $(\epsilon,\delta)$-local minimum. Otherwise,
	$\n{\nabla f(\tx)}\leq \mathG$ and $\lambda_{\min}(\nabla^2 f(\tx)) \leq -\delta$, this is exactly our second situation (around saddle points).
	According to \eqref{eq:escapehigh}, we know that the the function value decrease ($f(\tx)-f(x_T)$) is at least $\mathf=\frac{\delta^3}{2C_1'\rho^2}$ with probability at least $1-5\mathT\zeta'\geq 1/2$ (let $\zeta'\leq 1/10\mathT$), where $C_1' =\tdo(1)$.
	Similar to type-1 useful epoch, we know $N_2\leq \frac{4C_1'\rho^2\Delta_0}{\delta^3}$ with probability at least $1-\tdo(N_2\mathT\zeta')$ by a union bound. We can let $\zeta'\leq \tdo(1/N_2\mathT)$.
	
	Now, we are ready to bound the number of SFO calls in Theorem \ref{thm:ssrgd-lm} (finite-sum) as follows:
	\begin{align}
	&N_1(\tdo(1)n+n+mb) +N_2\Bigl(\tdo(1)n+\left\lceil\frac{\mathT}{m}\right\rceil n+\mathT b\Bigr) \notag\\
	&\leq \tdo\Big(\frac{\Delta_0n}{\eta m\mathG^2}+\frac{\rho^2\Delta_0}{\delta^3}(n+\frac{\sqrt{n}}{\eta \delta})\Big) \notag\\
	& \leq \tdo\Big(\frac{L\Delta_0\sqrt{n}}{\epsilon^2}
	+\frac{L\rho^2\Delta_0\sqrt{n}}{\delta^4}
	+ \frac{\rho^2\Delta_0n}{\delta^3}\Big). \label{eq:sfo-finite}
	\end{align}
By a union bound of these types and set $\zeta=\tdo(N_1m+N_2\mathT)\zeta'$ (note that $\zeta'$ only appears in the log term $\log(\frac{1}{\zeta'})$, so it can be chosen as small as we want), we know that the SFO calls of SSRGD can be bounded by \eqref{eq:sfo-finite} with probability $1-\zeta$.
This finishes the proof of Theorem~\ref{thm:ssrgd-lm}.
Now, it remains to prove Lemma \ref{lem:local} and \ref{lem:smallstuck}.

\begin{proofof}{Lemma~\ref{lem:local}}
	First, we know the variance bound \eqref{eq:highvar} holds with probability $1-2\zeta'$. Then by a union bound, it holds with probability $1-2t\zeta'$ for all $0\leq j\leq t-1$.
	Then, according to \eqref{eq:bm2}, we know for any $\tau \leq t$ in some epoch $s$
	\begin{align}
	f(x_{\tau})
	&\leq f(x_{sm}) - \frac{\eta}{2}\sum_{j=sm+1}^{\tau}\ns{\nabla f(x_{j-1})}
	- \big(\frac{1}{2\eta}- \frac{L}{2}-\frac{\eta C'^2L^2}{2}\big)\sum_{j=sm+1}^{\tau}\ns{x_j-x_{j-1}}  \notag\\
	&\leq f(x_{sm}) - \big(\frac{1}{2\eta}- \frac{L}{2}-\frac{\eta C'^2L^2}{2}\big)\sum_{j=sm+1}^{\tau}\ns{x_j-x_{j-1}} \notag\\
	&\leq f(x_{sm}) - \frac{C'L}{4}\sum_{j=sm+1}^{\tau}\ns{x_j-x_{j-1}}, \label{eq:dist}
	\end{align}
	where \eqref{eq:dist} holds by setting the step size $\eta\leq \frac{1}{(1+2C')L}$. Recall that $C' = O(\log\frac{dm}{\zeta'}\sqrt{\log\frac{d}{\zeta'}}) = \tdo(1)$.
	Now, we sum up \eqref{eq:dist} for all epochs before iteration $t$,
	\begin{align*}
	f(x_{t})
	&\leq f(x_{0}) - \frac{C'L}{4}\sum_{j=1}^{t}\ns{x_j-x_{j-1}}.
	\end{align*}
	Then, the proof is finished as
	\begin{align*}
	\n{x_t-x_0}\leq \sum_{j=1}^t\n{x_j-x_{j-1}}\leq \sqrt{t\sum_{j=1}^{t}\ns{x_j-x_{j-1}}}
	\leq \sqrt{\frac{4t(f(x_0)-f(x_t))}{C'L}}.
	\end{align*}
\end{proofof}

\begin{proofof}{Lemma~\ref{lem:smallstuck}}
	We prove this lemma by contradiction. Assume the contrary,
	\begin{align}\label{eq:distbound}
	\forall t\leq T,~~ \n{x_t-x_0} \leq \frac{\delta}{C_1\rho} \mathrm{~~and~~} \n{x_t'-x_0'} \leq \frac{\delta}{C_1\rho},
	\end{align}
	where $T:=\frac{\log(\frac{8\delta\sqrt{d}}{\rho r\zeta'})}{\eta\gamma} \leq \mathT :=\frac{\log(\frac{8\delta\sqrt{d}}{\rho r\zeta'})}{\eta\delta}$ (note that $\gamma\geq \delta$ due to $-\gamma:=\lambda_{\min}(\nabla^2 f(\tx))\leq -\delta$).
	We show that the distance between these two coupled sequences $w_t:=x_t-x_t'$ grows exponentially if they are not very close in the $e_1$ direction at the beginning, i.e., $w_0:=x_0-x_0'=r_0e_1$. Recall that $r_0=\frac{\zeta' r}{\sqrt{d}}$ and $e_1$ denotes the smallest eigenvector direction of Hessian $\hess := \nabla^2 f(\tx)$.
	However, $\n{w_t}=\n{x_t-x_t'}\leq \n{x_t-x_0}+\n{x_0-\tx}+\n{x_t'-x_0'}+\n{x_0'-\tx}\leq 2r+2\frac{\delta}{C_1\rho}$ according to \eqref{eq:distbound} and the perturbation radius $r$.
	It is not hard to see that if $\n{w_t}$ increases exponentially, this inequality cannot be true for reasonably large $t$, rendering a contradiction.
	
	In the following, we prove that $\n{w_t}$ increases exponentially by induction on $t$.
	First, we need the expression of $w_t$. 
	Define $\Delta_{\tau} :=\int_0^1(\nabla^2 f(x_\tau'+\theta(x_\tau-x_\tau'))-\hess)d\theta$
	and $y_\tau :=v_{\tau}-\nabla f(x_{\tau})-v_{\tau}'+\nabla f(x_{\tau}')$.
	Recall that $x_t=x_{t-1}-\eta v_{t-1}$ (see Line \ref{line:update} of Algorithm~\ref{alg:ssrgd}). Hence one can easily see that
	\begin{align}
	w_t&=w_{t-1}-\eta(v_{t-1}-v_{t-1}') \notag\\
	&=w_{t-1}-\eta\big(\nabla f(x_{t-1})- \nabla f(x_{t-1}')
	+v_{t-1}-\nabla f(x_{t-1})-v_{t-1}'+\nabla f(x_{t-1}')\big) \notag\\
	&=w_{t-1}-\eta\Big(\int_0^1\nabla^2 f(x_{t-1}' + \theta(x_{t-1}-x_{t-1}'))d\theta(x_{t-1}-x_{t-1}')+y_{t-1}\Big)
	\notag\\
	&=w_{t-1}-\eta\big((\hess +  \Delta_{t-1})w_{t-1}+y_{t-1}\big) \notag\\
	&=(I-\eta \hess)w_{t-1}-\eta(\Delta_{t-1}w_{t-1}+y_{t-1}) \label{eq:expwtandwt-1}\\
	&=(I-\eta \hess)^{t}w_0-\eta\sum_{\tau=0}^{t-1}(I-\eta \hess)^{t-1-\tau}(\Delta_\tau w_\tau+y_\tau). \label{eq:expw}
	\end{align}
	First, one can see that the first term of \eqref{eq:expw} is in the $e_1$ direction and it increases exponentially with respect to $t$, i.e., $(1+\eta\gamma)^t r_0 e_1$, where $-\gamma:=\lambda_{\min}(\hess)=\lambda_{\min}(\nabla^2 f(\tx))\leq -\delta$.
	Hence, to prove that $\n{w_t}$ increases exponentially, it suffices to show that the norm of the first term of \eqref{eq:expw} dominate that of the second term.
	For this purpose, we need the following bounds for $\n{w_t}$ and $\n{y_t}$, stated in the following lemma.
	
	\begin{lemma}\label{lem:boundwtyt}
	Suppose	$w_0:=x_0-x_0'=r_0e_1$ where $r_0=\frac{\zeta' r}{\sqrt{d}}$ and $e_1$ is the eigenvector corresponding to the smallest eigenvalue of Hessian $\hess := \nabla^2 f(\tx)$.
	If \eqref{eq:distbound} holds, then with probability $1-2T\zeta'$, the following bounds hold for all $t \leq T$:
	\begin{enumerate}
		\item $\frac{1}{2}(\base)^t r_0\leq\n{w_t}\leq\frac{3}{2}(\base)^t r_0$;
		\item $\n{y_t}\leq \frac{\gamma}{4C_2}(\base)^t r_0$.
	\end{enumerate}
	where $C_2 :=\log(\frac{8\delta\sqrt{d}}{\rho r\zeta'})$.
	\end{lemma}

\begin{proofof}{Lemma~\ref{lem:boundwtyt}}
	We prove this lemma inductively.
	First, check the base case $t=0$, $\n{w_0}=\n{r_0 e_1}=r_0$ and
	$\n{y_0}= \n{v_{0}-\nabla f(x_0)-v_{0}'+\nabla f(x_{0}')}=\n{\nabla f(x_0)-\nabla f(x_0)-\nabla f(x_{0}')+\nabla f(x_{0}')}=0$. 
	Now, assuming they hold for all $\tau\leq t-1$, we now prove they hold for $t$.
	For the bounds of $\n{w_t}$, 
	it suffices to show that the second term of \eqref{eq:expw} is dominated by half of the first term.
	Now, we first consider the first part of the second term:
	\begin{align}
	\n{\eta\sum_{\tau=0}^{t-1}(I-\eta \hess)^{t-1-\tau}(\Delta_\tau w_\tau)}
	&\leq \eta\sum_{\tau=0}^{t-1}(\base)^{t-1-\tau}\n{\Delta_\tau}\n{w_\tau} \notag\\
	&\leq \frac{3}{2}\eta(\base)^{t-1}r_0\sum_{\tau=0}^{t-1}\n{\Delta_\tau} \label{eq:0}\\
	&\leq \frac{3}{2}\eta(\base)^{t-1}r_0\sum_{\tau=0}^{t-1}\rho D_\tau^x \label{eq:1}\\
	&\leq \frac{3}{2}\eta(\base)^{t-1}r_0t\rho \big(\Dtop\big)\label{eq:2}\\
	&\leq \frac{3}{C_1}\eta\delta t(\base)^{t-1}r_0\label{eq:3}\\
	&\leq \frac{3\log(\frac{8\delta\sqrt{d}}{\rho r\zeta'})}{C_1}(\base)^{t-1}r_0 \label{eq:4}\\
	&\leq \frac{1}{4}(\base)^{t}r_0, \label{eq:5}
	\end{align}
	where \eqref{eq:0} uses the induction hypothesis for $w_\tau$ with $\tau\leq t-1$,
	\eqref{eq:1} uses Assumption~\ref{asp:smoothgandh} and the definition $D_\tau^x:=\max\{\n{x_\tau-\tx},\n{x_\tau'-\tx}\}$, \eqref{eq:2} follows from $\n{x_t-\tx}\leq\n{x_t-x_0}+\n{x_0-\tx}=\Dtop$ due to  \eqref{eq:distbound} and the perturbation radius $r$,
	\eqref{eq:3} holds by letting the perturbation radius $r\leq \frac{\delta}{C_1\rho}$,
	\eqref{eq:4} holds since $t\leq T\leq \mathT:=\frac{1}{\eta\delta}\log(\frac{8\delta\sqrt{d}}{\rho r\zeta'})$,
	and \eqref{eq:5} holds due to the definition of $C_1\geq 12\log(\frac{8\delta\sqrt{d}}{\rho r\zeta'})$.
	
	Now, the second part can be bounded as follows:
	\begin{align}
	\n{\eta\sum_{\tau=0}^{t-1}(I-\eta \hess)^{t-1-\tau}y_\tau}
	&\leq \eta\sum_{\tau=0}^{t-1}(\base)^{t-1-\tau}\n{y_\tau} \notag\\
	&\leq \eta\sum_{\tau=0}^{t-1}(\base)^{t-1-\tau} \frac{\gamma}{4C_2}(\base)^\tau r_0 \label{eq:10}\\
	&=\frac{\eta \gamma}{4C_2} t(\base)^{t-1} r_0 \notag\\
	&\leq \frac{\log(\frac{8\delta\sqrt{d}}{\rho r\zeta'})}{4C_2} (\base)^{t-1} r_0 \label{eq:11}\\
	&= \frac{1}{4}(\base)^{t}r_0, \label{eq:13}
	\end{align}
	where \eqref{eq:10} uses the induction for $y_\tau$ with $\tau\leq t-1$,
	\eqref{eq:11} holds since $t\leq T:=\frac{2\log(\frac{8\delta\sqrt{d}}{\rho r\zeta'})}{\eta\gamma}$,
	and \eqref{eq:13} holds due to the definition of $C_2=\log(\frac{8\delta\sqrt{d}}{\rho r\zeta'})$.
	
	Combining \eqref{eq:5} and \eqref{eq:13}, we can see that the norm of the second term of \eqref{eq:expw} is 
	at most one half of that of the first term.
	Note that the norm of the first term of \eqref{eq:expw} is $\n{(I-\eta \hess)^{t}w_0}=(1+\eta\gamma)^t r_0$. Thus, we have
	\begin{align}\label{eq:wt}
	\frac{1}{2}(\base)^t r_0\leq\n{w_t}\leq\frac{3}{2}(\base)^t r_0.
	\end{align}
	
	Now, the remaining thing is to prove the second bound $\n{y_t}\leq \frac{\gamma}{4C_2}(\base)^t r_0$,
	which is somewhat technical. 
	First, we write the concrete expression of $y_t$:
	\begin{align}
	y_t &= v_{t}-\nabla f(x_{t})-v_{t}'+\nabla f(x_{t}') \notag \\
	&= \frac{1}{b}\sum_{i\in I_b}\big(\nabla f_i(x_{t})-\nabla f_i(x_{t-1})\big)
	+ v_{t-1}-\nabla f(x_{t})\notag\\
	&\qquad
	-\frac{1}{b}\sum_{i\in I_b}\big(\nabla f_i(x_{t}')-\nabla f_i(x_{t-1}')\big)
	- v_{t-1}'+\nabla f(x_{t}')  \label{eq:30} \\
	&= \frac{1}{b}\sum_{i\in I_b}\big(\nabla f_i(x_{t})-\nabla f_i(x_{t-1})\big)
	+ \nabla f(x_{t-1})-\nabla f(x_{t})  \notag\\
	&\qquad -\frac{1}{b}\sum_{i\in I_b}\big(\nabla f_i(x_{t}')-\nabla f_i(x_{t-1}')\big)
	- \nabla f(x_{t-1}') +\nabla f(x_{t}') +y_{t-1} \notag\\
	&= \frac{1}{b}\sum_{i\in I_b}\big(\nabla f_i(x_{t})-\nabla f_i(x_{t}')
	-\nabla f_i(x_{t-1})+\nabla f_i(x_{t-1}')\big)\notag\\
	&\qquad
	-\big(\nabla f(x_{t})-\nabla f(x_{t}') - \nabla f(x_{t-1}) + \nabla f(x_{t-1}')\big)
	+ y_{t-1}, \notag
	\end{align}
	where \eqref{eq:30} is due to the definition of the estimator $v_t$ (see Line \ref{line:up2} of Algorithm~\ref{alg:ssrgd}).
	We further define the difference $z_t:=y_t-y_{t-1}$.
	It is not hard to verify that $\{y_t\}$ is a martingale sequence and $\{z_t\}$ is the associated martingale difference sequence.
	We can apply the Azuma-Hoeffding inequality to get an upper bound for $\n{y_t}$
	and then we prove $\n{y_t}\leq \frac{\gamma}{4C_2}(\base)^t r_0$ based on that upper bound.
	In order to apply the Azuma-Hoeffding inequality for martingale sequence $\n{y_t}$, we first need to bound the difference sequence $\{z_t\}$. 
	\begin{align}
	z_t=y_t-y_{t-1}& = \frac{1}{b}\sum_{i\in I_b}\big(\nabla f_i(x_{t})-\nabla f_i(x_{t}')
	-\nabla f_i(x_{t-1})+\nabla f_i(x_{t-1}')\big)\notag\\
	&\qquad
	-\big(\nabla f(x_{t})-\nabla f(x_{t}')
	- \nabla f(x_{t-1}) + \nabla f(x_{t-1}')\big)\notag\\
	&=\frac{1}{b}\sum_{i\in I_b} u_i, \label{eq:zk2}
	\end{align}
	where we define $u_i:=\big(\nabla f_i(x_{t})-\nabla f_i(x_{t}')\big)
	-\big(\nabla f_i(x_{t-1})-\nabla f_i(x_{t-1}')\big)
	-\big(\nabla f(x_{t})-\nabla f(x_{t}') \big)
	+ \big(\nabla f(x_{t-1}) - \nabla f(x_{t-1}')\big)$ in the last equality \eqref{eq:zk2}. 
	Then we have
	\begin{align}
	\|u_i\|&\leq \Big\|\int_0^1\nabla^2 f_i(x_{t}'
	+ \theta(x_{t}-x_{t}'))d\theta(x_{t}-x_{t}')
	-\int_0^1\nabla^2 f_i(x_{t-1}'
	+ \theta(x_{t-1}-x_{t-1}'))d\theta(x_{t-1}-x_{t-1}')\notag\\
	&\qquad
	-\int_0^1\nabla^2 f(x_{t}'
	+\theta(x_{t}-x_{t}'))d\theta(x_{t}-x_{t}')
	+\int_0^1\nabla^2 f(x_{t-1}'
	+\theta(x_{t-1}-x_{t-1}'))d\theta(x_{t-1}-x_{t-1}')\Big\| \notag\\
	&=\n{\hess_iw_t+\Delta_{t}^i w_t - (\hess_iw_{t-1}+\Delta_{t-1}^i w_{t-1})
		-(\hess w_t+\Delta_{t} w_t)+(\hess w_{t-1}+\Delta_{t-1} w_{t-1})} \notag  \\
	&\leq \n{(\hess_i-\hess)(w_t-w_{t-1})}
	+\n{(\Delta_{t}^i -\Delta_{t}) w_t-(\Delta_{t-1}^i-\Delta_{t-1}) w_{t-1}} \notag\\
	&\leq 2L\|w_t-w_{t-1}\|+2\rho D_t^x\n{w_t}+2\rho D_{t-1}^x\n{w_{t-1}}, \label{eq:b12}
	\end{align}
	where the equality holds since we define $\Delta_{t}:=\int_0^1(\nabla^2 f(x_t'+\theta(x_t-x_t'))-\hess)d\theta$ and $\Delta_{t}^i:=\int_0^1(\nabla^2 f_i(x_t'+\theta(x_t-x_t'))-\hess_i)d\theta$,
	and the last inequality holds due to the gradient and Hessian Lipschitz Assumption \ref{asp:smoothgandh} (recall $D_t^x:=\max\{\n{x_t-\tx},\n{x_t'-\tx}\}$).
	Then, consider the variance term:
	\begin{align}
	\E\Big[\sum_{i\in I_b}\|u_i\|^2\Big] &\leq b\E_i[\ns{\big(\nabla f_i(x_{t})-\nabla f_i(x_{t}')\big)
		-\big(\nabla f_i(x_{t-1})-\nabla f_i(x_{t-1}')\big)}] \notag\\
	&=b\E_i[\ns{\hess_iw_t+\Delta_{t}^i w_t
		- (\hess_iw_{t-1}+\Delta_{t-1}^i w_{t-1})}] \notag\\
	&\leq b(L\|w_t-w_{t-1}\|+\rho D_t^x\n{w_t}+\rho D_{t-1}^x\n{w_{t-1}})^2, \label{eq:b22}
	\end{align}
	where the first inequality uses the fact $\E[\ns{x-\E x}]\leq \E[\ns{x}]$, and the last inequality uses the gradient and Hessian Lipschitz Assumption \ref{asp:smoothgandh}.
	According to \eqref{eq:b12} and \eqref{eq:b22}, we can bound the difference $z_k$ by Bernstein inequality (Proposition \ref{prop:bernstein}) as (where $R=2L\|w_t-w_{t-1}\|+2\rho D_t^x\n{w_t}+2\rho D_{t-1}^x\n{w_{t-1}}$ and $\sigma^2=b(L\|w_t-w_{t-1}\|+\rho D_t^x\n{w_t}+\rho D_{t-1}^x\n{w_{t-1}})^2$)
	\begin{align*}
	\pr\Big\{\big\|z_t\big\|\geq \frac{\alpha}{b} \Big\} &\leq (d+1) \exp\Big(\frac{-\alpha^2/2}{\sigma^2+R\alpha/3}\Big)= \zeta_k,\\
	\end{align*}
	where the last equality holds by letting $\alpha=C\sqrt{b}(L\|w_t-w_{t-1}\|+\rho D_t^x\n{w_t}+\rho D_{t-1}^x\n{w_{t-1}})$, where $C=O(\log\frac{d}{\zeta_k})$.

	Now, we have a high probability bound for the difference sequence $\{z_k\}$, i.e.,
	\begin{align}
	\|z_k\| \leq c_k= \frac{C}{\sqrt{b}}(L\|w_t-w_{t-1}\|+\rho D_t^x\n{w_t}+\rho D_{t-1}^x\n{w_{t-1}})  \mathrm{~with~probability~} 1-\zeta_k. \notag
	\end{align}
	
	Next, we provide an upper bound for $\n{y_t}$ by using the martingale Azuma-Hoeffding inequality.
	Note that we only need to consider the current epoch that contains the iteration $t$ since each epoch we start with $y=0$.
	Let $s$ denote the current epoch, i.e, iterations from $sm+1$ to current $t$, where $t$ is no larger than $(s+1)m$.
	Define
	$$\beta:=\sqrt{8\sum_{k=sm+1}^{t} c_k^2\log\frac{d}{\zeta'}} = 
	\frac{C'}{\sqrt{b}}\sqrt{\sum_{k=sm+1}^{t}(L\|w_t-w_{t-1}\|+\rho D_t^x\n{w_t}+\rho D_{t-1}^x\n{w_{t-1}})^2},$$ 
	where $C'=O(C\sqrt{\log\frac{d}{\zeta'}})=O(\log\frac{d}{\zeta_k}\sqrt{\log\frac{d}{\zeta'}})=\tdo(1)$.
	According to Azuma-Hoeffding inequality (Proposition \ref{prop:azumahigh}) and letting $\zeta_k = \zeta'/m$, we have
	\begin{align*}
	\pr\Big\{\big\|y_{t}-y_{sm}\big\|\geq \beta \Big\} &\leq (d+1) \exp\Big(\frac{-\beta^2}{8\sum_{k=sm+1}^{t} c_k^2}\Big)+\zeta' = 2\zeta'.
	\end{align*}
	Recall that $y_k:=v_{k}-\nabla f(x_{k})-v_{k}'+\nabla f(x_{k}')$ and at the beginning point of this epoch $y_{sm}=0$ due to $v_{sm}=\nabla f(x_{sm})$ and $v_{sm}'=\nabla f(x_{sm}')$ (note that batch size $B=n$ in this finite-sum case). Thus. for any $t\in [sm+1,(s+1)m]$, we have 
	\begin{align}\label{eq:highvar2}
	\n{y_{t}}=\n{y_t-y_{sm}} \leq \beta:= \frac{C'}{\sqrt{b}}\sqrt{\sum_{k=sm+1}^{t}(L\|w_t-w_{t-1}\|+\rho D_t^x\n{w_t}+\rho D_{t-1}^x\n{w_{t-1}})^2}
	\end{align}
	holds with high probability $1-2\zeta'$. Furthermore, by a union bound, we know that \eqref{eq:highvar2} holds with probability at least $1-2T\zeta'$ for all $t\leq T$.
	
	Now, we show how to bound the right-hand-side of \eqref{eq:highvar2}.
	First, we show that the last two terms in the right-hand-side of \eqref{eq:highvar2} can be bounded as
	\begin{align}
	\rho D_t^x\n{w_t}+\rho D_{t-1}^x\n{w_{t-1}} &\leq
	\rho\big(\Dtop\big)\frac{3}{2}(\base)^t r_0 + \rho\big(\Dtop\big)\frac{3}{2}(\base)^{t-1} r_0 \notag\\
	& \leq 3\rho\big(\Dtop\big)(\base)^t r_0 \notag\\
	& \leq \frac{6\delta}{C_1}(\base)^t r_0, \label{eq:50}
	\end{align}
	where the first inequality follows from the induction hypothesis of $\n{w_{t-1}}\leq\frac{3}{2}(\base)^{t-1} r_0$ and the bound $\n{w_t}\leq\frac{3}{2}(\base)^t r_0$ in \eqref{eq:wt} which we have already proved,
	and the last inequality holds by letting the perturbation radius $r\leq \frac{\delta}{C_1 \rho}$.
	
	Now, we bound the first term of right-hand-side of \eqref{eq:highvar2}.
	According to \eqref{eq:expwtandwt-1}, we have 
	\begin{align}
	L\|w_t-w_{t-1}\| 
	&=L\n{\eta \hess w_{t-1}-\eta(\Delta_{t-1}w_{t-1}+y_{t-1})} \notag\\
	&\leq L\eta\n{\hess w_{t-1}} + L\eta\n{\Delta_{t-1}w_{t-1}+y_{t-1}} \notag\\
	&\leq  L\eta\n{\proj_{S_-}\hess w_{t-1}}
				+ L\eta\n{\proj_{S_+}\hess w_{t-1}} 
				+ L\eta\n{\Delta_{t-1}w_{t-1}+y_{t-1}} \label{eq:splitspace}\\
	&\leq L\eta\gamma\n{w_{t-1}}
	+ L\eta\n{\proj_{S_+}\hess w_{t-1}} 
	+ L\eta\n{\Delta_{t-1}}\n{w_{t-1}}+L\eta\n{y_{t-1}} \notag\\
	&\leq  (1+\frac{2}{C_1\rho}) L\eta\gamma\n{w_{t-1}}
	+ L\eta\n{\proj_{S_+}\hess w_{t-1}} 
	+L\eta\n{y_{t-1}} \label{eq:useDtx}\\
	&\leq (\frac{3}{2}+\frac{3}{C_1}+\frac{1}{4C_2}) L\eta\gamma(\base)^{t-1} r_0
	+ L\eta\n{\proj_{S_+}\hess w_{t-1}}, \label{eq:plugwandy}
	\end{align}
	where \eqref{eq:splitspace} holds by splitting the space into two subspace: 1) subspace $S_-$ spanned by the eigenvectors of $\hess$ with eigenvalues within $[-\gamma,0]$; 2) subspace $S_+$ spanned by the eigenvectors of $\hess$ with eigenvalues within $(0,L]$,
	\eqref{eq:useDtx} holds from the following \eqref{eq:Dtx}, and
	the last inequality \eqref{eq:plugwandy}  follows from the induction hypothesis of $\n{w_{t-1}}\leq\frac{3}{2}(\base)^{t-1} r_0$ and  $\n{y_{t-1}}\leq\frac{\gamma}{4C_2}(\base)^{t-1} r_0$.
	\begin{align}\label{eq:Dtx}
	\forall t\leq T,\quad  \n{\Delta_{t}}\leq \rho D_t^x \leq \rho\big(\Dtop\big) \leq \frac{2\delta}{C_1}\leq \frac{2\gamma}{C_1},
	\end{align}
	which holds by letting the perturbation radius $r\leq \frac{\delta}{C_1 \rho}$, and noting that $\gamma\geq \delta$ (recall $-\gamma:=\lambda_{\min}(\hess)=\lambda_{\min}(\nabla^2 f(\tx))\leq -\delta$).
	
	Now, we bound the second term of \eqref{eq:plugwandy} as follows:
	\begin{align}
	&L\eta\n{\proj_{S_+}\hess w_{t-1}}	\notag\\
	&=
	L\eta\big\|-\proj_{S_+}\hess(I-\eta \hess)^{t-1}w_0
	-\sum_{\tau=0}^{t-2}\proj_{S_+} \eta \hess(I-\eta \hess)^{t-2-\tau}(\Delta_\tau w_\tau+y_\tau)\big\|
	\label{eq:usewt-1}\\
	&=
	L\eta\big\|-\sum_{\tau=0}^{t-2}\proj_{S_+} \eta \hess(I-\eta \hess)^{t-2-\tau}(\Delta_\tau w_\tau+y_\tau)\big\| \label{eq:e1-neg}\\
	& \leq 
	L\eta\sum_{\tau=0}^{t-2}\big\|
	\proj_{S_+} \eta \hess(I-\eta \hess)^{t-2-\tau}\big\|
	\n{\Delta_\tau w_\tau+y_\tau} \notag\\
	&\leq L\eta\sum_{\tau=0}^{t-2}\frac{1}{t-1-\tau}
	\n{\Delta_\tau w_\tau+y_\tau}  \label{eq:span+}\\
	& \leq L\eta\log t	\max_{0\leq k\leq t-2}\n{\Delta_k w_k+y_k} \notag\\
	&\leq L\eta \log t \Big(\frac{2\gamma}{C_1}\frac{3}{2}(\base)^{t-2} r_0 + \frac{\gamma}{4C_2}(\base)^{t-2} r_0\Big) \label{eq:plugDtx-wandy}\\
	& =  \Big(\frac{3}{C_1}\log t +\frac{1}{4C_2}\log t\Big)L\eta\gamma(\base)^{t-2} r_0, \label{eq:65}
	\end{align}
	where the first equality \eqref{eq:usewt-1} follows from \eqref{eq:expw},
	\eqref{eq:e1-neg} holds since $w_0=r_0e_1$ is in the $e_1$ direction,
	\eqref{eq:plugDtx-wandy} uses \eqref{eq:Dtx} and
	the induction hypothesis of $\n{w_{k}}\leq\frac{3}{2}(\base)^{k} r_0$ and  $\n{y_{k}}\leq\frac{\gamma}{4C_2}(\base)^{k} r_0$, for all $k\leq t-1$.
	The inequality \eqref{eq:span+} follows from the fact $\max_{x\in[0,1]} x(1-x)^t \leq \frac{1}{t+1}$. Note that $S_+$ denotes the subspace spanned by the eigenvectors of $\hess$ with eigenvalues within $(0,L]$, thus $\big\| \proj_{S_+} \eta \hess(I-\eta \hess)^{t-2-\tau}\big\| \leq \max_{\lambda \in (0,L]} \eta\lambda(1-\eta\lambda)^{t-2-\tau} \leq \frac{1}{t-1-\tau}$. Also note that $\eta\lambda\leq 1$ due to $\eta\leq \frac{1}{L}$ and $\lambda \in (0,L]$.

	By plugging \eqref{eq:65} into \eqref{eq:plugwandy}, we have
	\begin{align}
	L\|w_t-w_{t-1}\| 
	&\leq \Big(\frac{3}{2}+\frac{3 (1+\log t)}{C_1}+\frac{1+\log t}{4C_2}\Big) L\eta\gamma(\base)^{t-1} r_0. \label{eq:firstterm}
	\end{align}
	
	Now we can bound $\n{y_t}$ by plugging \eqref{eq:50} and \eqref{eq:firstterm} into \eqref{eq:highvar2} and noting that $t-sm\leq m\leq b$:
	\begin{align}
	\n{y_{t}}
	&\leq C'\left(\frac{6\delta}{C_1}(\base)^t r_0+\Big(\frac{3}{2}+\frac{3 (1+\log t)}{C_1}+\frac{1+\log t}{4C_2}\Big) L\eta\gamma(\base)^{t-1} r_0\right) \notag\\
	&\leq \Big(\frac{6C'}{C_1}+\big(\frac{3}{2}+\frac{3 (1+\log t)}{C_1}+\frac{1+\log t}{4C_2}\big)C' L\eta\Big)\gamma(\base)^t r_0 \notag\\
	&\leq \Big(\frac{1}{8C_2}+\frac{1}{8C_2} \Big)\gamma(\base)^t r_0 \notag\\
	&= \frac{1}{4C_2}\gamma(\base)^t r_0, 
	\end{align}
	where the second inequality holds due to $\delta\leq \gamma$ (recall $-\gamma:=\lambda_{\min}(\hess)=\lambda_{\min}(\nabla^2 f(\tx))\leq -\delta$),
	and the last inequality holds by letting
	$C_1\geq 48C'C_2$ (recall that $C_2:=\log(\frac{8\delta\sqrt{d}}{\rho r\zeta'})$ defined in Lemma~\ref{lem:boundwtyt}), 
	and $\eta \leq \frac{1}{15(1+\log t)C' L}$.
	Recall that \eqref{eq:highvar2} holds with probability at least $1-2T\zeta'$ for all $t\leq T$.
	This finishes the proof of Lemma~\ref{lem:boundwtyt}.
	\end{proofof}

	From the Lemma \ref{lem:boundwtyt}, one can see that $\n{w_t} \geq \frac{1}{2}(\base)^t r_0=\frac{1}{2}(\base)^t\frac{\zeta' r}{\sqrt{d}}$. On the other hand, $\n{w_t}:=\n{x_t-x_t'}\leq \n{x_t-x_0}+\n{x_0-\tx}+\n{x_t'-x_0'}+\n{x_0'-\tx}\leq 2r+2\frac{\delta}{C_1\rho}\leq \frac{4\delta}{C_1\rho}$ according to \eqref{eq:distbound} and the perturbation radius $r\leq \frac{\delta}{C_1 \rho}$.
	Hence, for any $t\geq T= \frac{1}{\eta\gamma}\log(\frac{8\delta\sqrt{d}}{\rho r\zeta'})$, we get a contradiction to \eqref{eq:distbound}, i.e., $\n{w_t} \geq \frac{1}{2}(\base)^T\frac{\zeta' r}{\sqrt{d}}\geq \frac{4\delta}{\rho} \geq \frac{4\delta}{C_1\rho}$, where
	$C_1\geq 1+48C'\log(\frac{8\delta\sqrt{d}}{\rho r\zeta'}) \geq 1$ defined in Lemma~\ref{lem:smallstuck}.
	Also note that $T= \frac{1}{\eta\gamma}\log(\frac{8\delta\sqrt{d}}{\rho r\zeta'})\leq \mathT:=\frac{1}{\eta\delta}\log(\frac{8\delta\sqrt{d}}{\rho r\zeta'})$ due to $\delta\leq \gamma$. 
	This contradiction finishes the proof of Lemma~\ref{lem:smallstuck}.
	
\end{proofof}

\subsection{Proof of Theorem \ref{thm:ssrgd-lm} (online)}
\label{app:ssrgd-lm-online}

The proof for the online case follows almost the same framework as in the finite-sum case in Section \ref{app:ssrgd-lm-finite}.
Although the only difference in the algorithm is that here we compute a large batch of stochastic gradient ($v_{sm}\neq \nabla f(x_{sm})$ in Line \ref{line:super} and \ref{line:up1} of Algorithm~\ref{alg:ssrgd}), instead of a 
full gradient, it leads to many changes in the analysis. 
Hence, we present the full proof for the online case as well. 
Again, we distinguish two situations, the {\em large gradients} case, in which the function value decreases 
significantly, and the {\em around saddle points} case, in which we add a random perturbation.

\vspace{1mm}
\noindent{{\bf Large Gradients}: }
First, we provide a high probability bound for the variance term, and then use it to get a high probability bound
for the decrease of the function.
Note that in this online case, $v_{sm}= \frac{1}{B}\sum_{j\in I_B}\nabla f_j(x_{sm})$ at the beginning of each epoch instead of $v_{sm}= \nabla f(x_{sm})$ (where $B=n$) in the previous finite-sum case.
Thus we first need a high probability bound for $\n{v_{sm}-\nabla f(x_{sm})}$.
According to Assumption \ref{asp:var2}, we have
\begin{align*}
\n{\nabla f_j(x)-\nabla f(x)} &\leq \sigma, \notag\\
\sum_{j\in I_B}\ns{\nabla f_j(x)-\nabla f(x)} &\leq B\sigma^2.
\end{align*}
By applying Bernstein inequality (Proposition \ref{prop:bernstein}), we get the high probability bound for $\n{v_{sm}-\nabla f(x_{sm})}$ as follows:
\begin{align*}
\pr\Big\{\big\|v_{sm}-\nabla f(x_{sm})\big\|\geq \frac{t}{B} \Big\} &\leq  (d+1) \exp\Big(\frac{-t^2/2}{B\sigma^2+ \sigma t/3}\Big)
\notag  = \zeta',
\end{align*}
where the last equality holds by letting $t=C_3\sqrt{B}\sigma$, where $C_3=O(\log\frac{d}{\zeta'})=\tdo(1)$.
Now, we have a high probability bound for $\n{v_{sm}-\nabla f(x_{sm})}$, i.e.,
\begin{align}\label{eq:varhigh_start}
\big\|v_{sm}-\nabla f(x_{sm})\big\| \leq  \frac{C_3\sigma}{\sqrt{B}} \quad \mathrm{~with~probability~} 1-\zeta'.
\end{align}

Now we obtain a high probability bound for the variance term of other points beyond the starting points. Recall that $v_k=\frac{1}{b}\sum_{i\in I_b}\big(\nabla f_i(x_{k})-\nabla f_i(x_{k-1})\big) + v_{k-1}$ (see Line \ref{line:up2} of Algorithm~\ref{alg:ssrgd}), and the martingale sequence $y_k:=v_k-\nabla f(x_k)$, $z_k:=y_k-y_{k-1}$, which is the associated martingale difference sequence, and 
$u_i:=\nabla f_i(x_{k})-\nabla f_i(x_{k-1})-(\nabla f(x_k) -\nabla f(x_{k-1}))$.
By \eqref{eq:zk}, we know that
\begin{align}
	z_k=y_k-y_{k-1}= \frac{1}{b}\sum_{i\in I_b}\Big(\nabla f_i(x_{k})-\nabla f_i(x_{k-1})
	-(\nabla f(x_k) -\nabla f(x_{k-1}))\Big)=\frac{1}{b}\sum_{i\in I_b} u_i. 
\end{align}
Using the same argument as in Section~\ref{app:ssrgd-lm-finite} (See \eqref{eq:b1},\eqref{eq:b2},\eqref{eq:zkbound}), one can see that 
$\|u_i\|\leq 2L\|x_{k}-x_{k-1}\|$ and
$
\E[\sum_{i\in I_b}\|u_i\|^2] \leq bL^2\ns{x_{k}-x_{k-1}},
$
and then one can apply Bernstein inequality (Proposition \ref{prop:bernstein}) to see that
\begin{align}
\|z_k\| \leq c_k= \frac{CL}{\sqrt{b}}\n{x_{k}-x_{k-1}} \quad \mathrm{~with~probability~} 1-\zeta_k,
\end{align}
where $C=O(\log\frac{d}{\zeta_k})=\tdo(1)$. 

Now, we are ready to get a high probability bound for the variance term using the martingale Azuma-Hoeffding inequality.
Consider in a specific epoch $s$, i.e, iterations $t$ from $sm+1$ to current $sm+k$, where $k$ is less than $m$.
Let $\beta:=\sqrt{8\sum_{t=sm+1}^{sm+k} c_t^2\log\frac{d}{\zeta'}}
=\frac{C'L}{\sqrt{b}}\sqrt{\sum_{t=sm+1}^{sm+k}\ns{x_{t}-x_{t-1}}}$, where $C'=O(C\sqrt{\log\frac{d}{\zeta'}})=O(\log\frac{d}{\zeta_k}\sqrt{\log\frac{d}{\zeta'}})=\tdo(1)$.
According to Azuma-Hoeffding inequality (Proposition \ref{prop:azumahigh}) and letting $\zeta_k = \zeta'/m$, we have
\begin{align*}
\pr\Big\{\big\|y_{sm+k}-y_{sm}\big\|\geq \beta \Big\} &\leq (d+1) \exp\Big(\frac{-\beta^2}{8\sum_{t=sm+1}^{sm+k} c_t^2}\Big)+\zeta' = 2\zeta'.
\end{align*}
Recall that $y_k:=v_k-\nabla f(x_k)$ and at the beginning point of this epoch $\n{y_{sm}}=\n{v_{sm}-\nabla f(x_{sm})}\leq C_3\sigma/\sqrt{B}$ with probability $1-\zeta'$, where $C=O(\log\frac{d}{\zeta'})=\tdo(1)$ (see \eqref{eq:varhigh_start}).
Combining with \eqref{eq:varhigh_start} and using a union bound, for any  $t\in [sm+1,(s+1)m]$, we have that
\begin{align}\label{eq:highvar1}
\n{v_{t-1}-\nabla f(x_{t-1})}=\n{y_{t-1}} \leq \beta +\n{y_{sm}} \leq \frac{C'L\sqrt{\sum_{j=sm+1}^{t-1}\ns{x_{j}-x_{j-1}}}}{\sqrt{b}} +\frac{C_3\sigma}{\sqrt{B}}
\end{align}
holds with probability $1-3\zeta'$.

Now, we use it to obtain a high probability bound for the decrease of the function value.
We sum up \eqref{eq:key-ss} from the beginning of this epoch $s$, i.e., iterations from $sm+1$ to $t$, by plugging \eqref{eq:highvar1} into them to get:
\begin{align}
f(x_{t}) &\leq f(x_{sm}) - \frac{\eta}{2}\sum_{j=sm+1}^{t}\ns{\nabla f(x_{j-1})}
- \big(\frac{1}{2\eta}- \frac{L}{2}\big)\sum_{j=sm+1}^{t}\ns{x_j-x_{j-1}} \notag\\
&\qquad + \frac{\eta}{2}\sum_{k=sm+1}^{t-1}\frac{2C'^2L^2\sum_{j=sm+1}^{k}\ns{x_{j}-x_{j-1}}}{b}
+\frac{\eta}{2}\sum_{j=sm+1}^{t}\frac{2C_3^2\sigma^2}{B} \label{eq:plughighvar1}\\
&\leq f(x_{sm}) - \frac{\eta}{2}\sum_{j=sm+1}^{t}\ns{\nabla f(x_{j-1})}
- \big(\frac{1}{2\eta}- \frac{L}{2}\big)\sum_{j=sm+1}^{t}\ns{x_j-x_{j-1}} \notag\\
&\qquad + \frac{\eta C'^2L^2}{b}\sum_{k=sm+1}^{t-1}\sum_{j=sm+1}^{k}\ns{x_{j}-x_{j-1}}
+ \frac{(t-sm)\eta C_3^2\sigma^2}{B} \notag\\
&\leq f(x_{sm}) - \frac{\eta}{2}\sum_{j=sm+1}^{t}\ns{\nabla f(x_{j-1})}
- \big(\frac{1}{2\eta}- \frac{L}{2}\big)\sum_{j=sm+1}^{t}\ns{x_j-x_{j-1}} \notag\\
&\qquad + \frac{\eta C'^2L^2(t-1-sm)}{b}\sum_{j=sm+1}^{t}\ns{x_{j}-x_{j-1}}
+ \frac{(t-sm)\eta C_3^2\sigma^2}{B} \notag\\
&\leq f(x_{sm}) - \frac{\eta}{2}\sum_{j=sm+1}^{t}\ns{\nabla f(x_{j-1})}
- \big(\frac{1}{2\eta}- \frac{L}{2}-\eta C'^2L^2\big)\sum_{j=sm+1}^{t}\ns{x_j-x_{j-1}} \notag\\
&\qquad + \frac{(t-sm)\eta C^2\sigma^2}{B} \label{eq:bm21}\\
&\leq f(x_{sm}) - \frac{\eta}{2}\sum_{j=sm+1}^{t}\ns{\nabla f(x_{j-1})} + \frac{(t-sm)\eta C_3^2\sigma^2}{B} \label{eq:eta21},
\end{align}
where \eqref{eq:bm21} holds if the minibatch size $b\geq m$ (note that here $t\leq (s+1)m$), and
\eqref{eq:eta21} holds if the step size $\eta\leq \frac{1}{(1+2C')L}$, where $C'=O(\log\frac{dm}{\zeta'}\sqrt{\log\frac{d}{\zeta'}})$.
Note that \eqref{eq:plughighvar1} uses \eqref{eq:highvar1} which holds with probability $1-3\zeta'$. Thus by a union bound, we know that \eqref{eq:eta21} holds with probability at least $1-3m\zeta'$.

Next, we show an analogue of Lemma~\ref{lem:first}
which connects the guarantees between first situation (large gradients) and second situation (around saddle points) by  relating to the \emph{gradient of the starting point} of each epoch (see Line \ref{line:super}
of Algorithm~\ref{alg:ssrgd}). This proof requires several modifications since we use stochastic gradients for $v_{sm}$.

\begin{lemma}[Two Situations]
	\label{lem:firstonline}
	For any epoch $s$, let $x_t$ be a point uniformly sampled from this epoch $\{x_{j} \}_{j=sm+1}^{(s+1)m}$ and choose the step size $\eta \leq \frac{1}{(1+2C')L}$ (where $C'=O(\log\frac{dm}{\zeta'}\sqrt{\log\frac{d}{\zeta'}})=\tdo(1)$) and the minibatch size $b\geq m$.
	Then for any $\mathG>0$, by letting batch size $B\geq \frac{256C_3^2\sigma^2}{\mathG^2}$ (where $C_3=O(\log\frac{d}{\zeta'})=\tdo(1)$), we have two cases:
	\begin{enumerate}
		\item If at least half of points in this epoch have gradient norm no larger than $\frac{\mathG}{2}$, then $\n{\nabla f(x_{(s+1)m})}\leq \frac{\mathG}{2}$ and $\|v_{(s+1)m} \| \le \mathG$ hold with probability at least $1/3$;
		\item Otherwise, we know $f(x_{sm}) - f(x_t) \ge \frac{7\eta m\mathG^2}{256}$ holds with probability at least $1/5$.
	\end{enumerate}
	Moreover, $f(x_t) \le f(x_{sm}) + \frac{(t-sm)\eta C_3^2\sigma^2}{B}$ holds with high probability $1-3m\zeta'$ no matter which case happens.
\end{lemma}

\begin{proofof}{Lemma~\ref{lem:firstonline}}
	There are two cases in this epoch:
	\begin{enumerate}
		\item If at least half of points in this epoch $\{x_{j} \}_{j=sm+1}^{(s+1)m}$ have gradient norm no larger than $\frac{\mathG}{2}$, then it is easy to see that a uniformly sampled point $x_t$ has gradient norm $\n{\nabla f(x_t)}\leq \frac{\mathG}{2}$ with probability at least $1/2.$ Moreover, note that the starting point of the next epoch $x_{(s+1)m}=x_t$ (i.e., Line \ref{line:randompoint} of Algorithm~\ref{alg:ssrgd}), thus we have $\n{\nabla f(x_{(s+1)m})}\leq \frac{\mathG}{2}$ with probability $1/2$. According to \eqref{eq:varhigh_start}, we have
		$\|v_{(s+1)m}-\nabla f(x_{(s+1)m})\| \leq  \frac{C_3\sigma}{\sqrt{B}}$ with probability $1-\zeta'$, where $C=O(\log\frac{d}{\zeta'})=\tdo(1)$.
		By a union bound, with probability at least $1/3$ (e.g., choose $\zeta'\leq 1/6$), we have
		$$\|v_{(s+1)m}\|\leq \frac{C_3\sigma}{\sqrt{B}} + \frac{\mathG}{2}\leq \frac{\mathG}{16}+\frac{\mathG}{2} \leq \mathG.$$
		\item Otherwise, at least half of points have gradient norm larger than $\frac{\mathG}{2}$. Then, as long as the sampled point $x_t$ falls into the last quarter of $\{x_{j} \}_{j=sm+1}^{(s+1)m}$, we know $\sum_{j=sm+1}^{t}\ns{\nabla f(x_{j-1})}\geq \frac{m\mathG^2}{16}$. This holds with probability at least $1/4$ since $x_t$ is uniformly sampled. Then by combining with \eqref{eq:eta21}, we obtain that the function value decreases
		$$f(x_{sm}) - f(x_t) \geq \frac{\eta}{2}\sum_{j=sm+1}^{t}\ns{\nabla f(x_{j-1})}-\frac{(t-sm)\eta C_3^2\sigma^2}{B} \ge \frac{\eta m\mathG^2}{32}-\frac{\eta m\mathG^2}{256}=\frac{7\eta m\mathG^2}{256},$$ where the last inequality is due to $B\geq \frac{256C_3^2\sigma^2}{\mathG^2}$.
		Note that
		\eqref{eq:eta21} holds with high probability $1-3m\zeta'$ if we choose the minibatch size $b\geq m$ and the step size $\eta\leq \frac{1}{(1+2C')L}$.
		By a union bound, the function value decrease $f(x_{sm}) - f(x_t) \ge \frac{\eta m\mathG^2}{64}$ with probability at least $1/5$ (e.g., choose $\zeta'\leq 1/60m$).
	\end{enumerate}
	Again according to \eqref{eq:eta21}, $f(x_t)\leq f(x_{sm})+\frac{(t-sm)\eta C_3^2\sigma^2}{B}$holds with high probability $1-3m\zeta'$.
\end{proofof}

Note that if Case 2 happens, the function value would decrease significantly in this epoch $s$ (corresponding to the first situation large gradients). Otherwise if Case 1 happens, we know the starting point of the next epoch $x_{(s+1)m}=x_t$ (i.e., Line \ref{line:randompoint} of Algorithm~\ref{alg:ssrgd}), then we know $\n{\nabla f(x_{(s+1)m})}\leq \frac{\mathG}{2}$ and $\|v_{(s+1)m} \| \le \mathG$. In this case, we start a super epoch (corresponding to the second situation around saddle points). Note that if $\lambda_{\min}(\nabla^2 f(x_{(s+1)m}))> -\delta$, the point $x_{(s+1)m}$ is already an $(\epsilon,\delta)$-local minimum.

\vspace{3mm}
\noindent{{\bf Around Saddle Points} $\|v_{(s+1)m} \| \le \mathG$ and $\lambda_{\min}(\nabla^2 f(x_{(s+1)m})) \leq -\delta$: }
In this situation, we show that the function value decreases significantly in a \emph{super epoch} with high probability by adding a random perturbation at the initial point $\tx=x_{(s+1)m}$. We denote $x_0:=\tx+\xi$ to denote the starting point of the super epoch after the perturbation, where $\xi$ uniformly $\sim \mathbb{B}_0(r)$ and the perturbation radius is $r$ (see Line \ref{line:init} of Algorithm~\ref{alg:ssrgd}).
Again, we follow the \emph{two-point analysis} developed in \citet{jin2017escape}.
In particular, consider two coupled points $x_0$ and $x_0'$ with $w_0:=x_0-x_0'=r_0e_1$, where $r_0$ is a scalar and $e_1$ denotes the smallest eigenvector direction of Hessian $\hess := \nabla^2 f(\tx)$. 
We show that at least for one of these two coupled sequences $\{x_t\}$ and $\{x_t'\}$,
the function value decrease a lot (escape the saddle point), i.e.,
\begin{align}\label{eq:funcd1}
\exists t\leq\mathT, \mathrm{~~such ~that~~} \max\{f(x_0)-f(x_t), f(x_0')-f(x_t')\} \geq 2\mathf.
\end{align}
The proof outline of \eqref{eq:funcd1} is the same as that in Section~\ref{app:ssrgd-lm-finite}. 
We assume by contradiction that $f(x_0)-f(x_t)<2\mathf$ and $f(x_0')-f(x_t')<2\mathf$.
Similar to Lemma \ref{lem:local} and Lemma \ref{lem:smallstuck}, we
need the following two technical lemmas in the online setting.  
Their proofs are deferred to the end of this section.

\begin{lemma}[Localization]
	\label{lem:localonline}
	Let $\{x_t\}$ denote the sequence by running SSRGD update steps (Line \ref{line:up1}--\ref{line:up2} of Algorithm~\ref{alg:ssrgd}) from $x_0$.
	Moreover, let the step size $\eta\leq \frac{1}{(1+2C')L}$ and minibatch size $b\geq m$.
	With probability $1-3t\zeta'$, we have
	\begin{align}\label{eq:localonline}
	\forall t\geq 0,~~ \n{x_t-x_0}\leq \sqrt{\frac{4t(f(x_0)-f(x_t))}{C'L} +\frac{4t^2\eta C_3^2\sigma^2}{C'LB}},
	\end{align}
	where $C'=O(\log\frac{dm}{\zeta'}\sqrt{\log\frac{d}{\zeta'}})=\tdo(1)$ and $C_3=O(\log\frac{d}{\zeta'})=\tdo(1)$.
\end{lemma}

\begin{lemma}[Small Stuck Region]
	\label{lem:smallstuckonline}
	If the initial point $\tx$ satisfies $-\gamma:=\lambda_{\min}(\nabla^2 f(\tx))\leq -\delta$,
	then let $\{x_t\}$ and $\{x_t'\}$ be two coupled sequences by running SSRGD update steps (Line \ref{line:up1}--\ref{line:up2} of Algorithm~\ref{alg:ssrgd}) with the same choice of batches and minibatches (i.e., $I_B$'s in Line \ref{line:up1} of Algorithm~\ref{alg:ssrgd} and $I_b$'s in Line \ref{line:up2} of Algorithm~\ref{alg:ssrgd}) from $x_0$ and $x_0'$ with $w_0:=x_0-x_0'=r_0e_1$, where $x_0\in\mathbb{B}_{\tx}(r)$, $x_0'\in\mathbb{B}_{\tx}(r)$ , $r_0=\frac{\zeta' r}{\sqrt{d}}$ and $e_1$ denotes the smallest eigenvector direction of Hessian $\nabla^2 f(\tx)$.
	Moreover, let the super epoch length $\mathT=\frac{\log(\frac{8\delta\sqrt{d}}{\rho r\zeta'})}{\eta\delta}=\tdo(\frac{1}{\eta\delta})$, the step size $\eta\leq  \frac{1}{30(1+\log \mathT)C'L}=\tdo(\frac{1}{L})$, minibatch size $b\geq m$, batch size $B=\tdo(\frac{\sigma^2}{\mathG^2})$ and
	the perturbation radius $r\leq \frac{\delta}{C_1 \rho}$. Then with probability $1-3T\zeta'$, we have
	\begin{align}\label{eq:stuckonline}
	\exists T\leq \mathT,~~ \max\{\n{x_T-x_0}, \n{x_T'-x_0'}\}\geq \frac{\delta}{C_1\rho},
	\end{align}
	where where $C'=O(\log\frac{dm}{\zeta'}\sqrt{\log\frac{d}{\zeta'}})=\tdo(1)$
	and $C_1\geq 1+96C' \log(\frac{8\delta\sqrt{d}}{\rho r \zeta'})=\tdo(1)$.
\end{lemma}

Based on these two lemmas, we are ready to show that \eqref{eq:funcd1} holds with high probability. Without loss of generality, we assume $\n{x_T-x_0} \geq \frac{\delta}{C_1\rho}$ in \eqref{eq:stuckonline} (note that \eqref{eq:localonline} holds for both $\{x_t\}$ and $\{x_t'\}$). Then plugging it into \eqref{eq:localonline}, we obtain
\begin{align}
\sqrt{\frac{4T(f(x_0)-f(x_T))}{C'L} +\frac{4T^2\eta C_3^2\sigma^2}{C'LB}} &\geq \frac{\delta}{C_1\rho} 
\end{align}
Hence, we can see that
\begin{align}
f(x_0)-f(x_T) &\geq \frac{C'L\delta^2}{4C_1^2\rho^2T}
-\frac{T\eta C_3^2\sigma^2}{B}\notag\\
&\geq \frac{ C'L\eta\delta^3}{4C_1^2\rho^2\log(\frac{8\delta\sqrt{d}}{\rho r\zeta'})}
- \frac{C_3^2\sigma^2\log(\frac{8\delta\sqrt{d}}{\rho r\zeta'})}{B\delta}
\label{eq:plugTthres}\\
&\geq\frac{\delta^3}{C_1'\rho^2} \label{eq:largeeta1}\\
&\overset{\mathrm{def}}{=}2\mathf,\notag
\end{align}
where the last equality is due to the definition of $\mathf:=\frac{\delta^3}{2C_1'\rho^2}=\tdo(\frac{\delta^3}{\rho^2})$,
\eqref{eq:plugTthres} is due to $T\leq \mathT:=\frac{\log(\frac{8\delta\sqrt{d}}{\rho r\zeta'})}{\eta\delta}$, and \eqref{eq:largeeta1} holds by letting
$C_1'=\frac{8C_1^2\log(\frac{8\delta\sqrt{d}}{\rho r\zeta'})}{C'L\eta} =\tdo(1)$.
Recall that $B=\tdo(\frac{\sigma^2}{\mathG^2})$ and $\mathG\leq \delta^2/\rho$.
Thus, we have already proved that at least one of sequences $\{x_t\}$ and $\{x_t'\}$
escapes the saddle point with probability 
$1-6T\zeta'$ (by union bound of \eqref{eq:localonline} and \eqref{eq:stuckonline}), i.e.,
\begin{align}
\exists T\leq \mathT,~~ \max\{f(x_0)-f(x_T), f(x_0')-f(x_T')\} \geq 2\mathf,
\end{align}
if their starting points $x_0$ and $x_0'$ satisfying $w_0:=x_0-x_0'=r_0e_1$.

Next, using exactly the same volume argument as in Section~\ref{app:ssrgd-lm-finite}, we obtain that 
\begin{align}\label{eq:escapehigh1}
	f(\tx)-f(x_T)=f(\tx)-f(x_0)+f(x_0)-f(x_T) \geq -\mathf+2\mathf=\frac{\delta^3}{2C_1'\rho^2}
\end{align}
holds with probability $1-(6T+1)\zeta'\geq 1-7\mathT\zeta'$, where $C_1'=\tdo(1)$. Here we use the fact $f(x_0)\leq f(\tx) +\mathf$ which follows from \eqref{eq:perturbless}.
Hence, we have finished the proof for the second situation (around saddle points). 

\vspace{3mm}
\noindent{{\bf Combing these two situations (large gradients and around saddle points) to prove Theorem \ref{thm:ssrgd-lm}:}}
We distinguishing the epochs into three types, \emph{Type-1 useful epoch}, \emph{Wasted epoch} and \emph{Type-2 useful super epoch} in exactly the same way as in Section~\ref{app:ssrgd-lm-finite}.
Recall in a Type-1 useful epoch, at least half of points in this epoch have gradient norm larger than $\mathG/2$ (Case 2 of Lemma \ref{lem:firstonline}).
If at least half of points in this epoch have gradient norm no larger than $\mathG/2$ and the starting point of the next epoch has estimated gradient norm larger than $\mathG$, we say it is a  wasted epoch.
In a Type-2 useful super epoch, at least half of points in this epoch have gradient norm no larger than $\mathG$ and the starting point of the next epoch has estimated gradient norm no larger than $\mathG$
(Case 1 of Lemma \ref{lem:firstonline}).
The argument is very similar to the one in Section~\ref{app:ssrgd-lm-finite} as well, except some quantitative details.

	First,  we can see that the probability of a wasted epoch happened is less than $2/3$ due to the random stop (see Case 1 of Lemma \ref{lem:firstonline}). Note for different wasted epochs, returned points are independently sampled.
	Thus, with high probability $1-\zeta'$, at most $O(\log\frac{1}{\zeta'})=\tdo(1)$ wasted epochs would happen before a type-1 useful epoch or type-2 useful super epoch.
	We use $N_1$ and $N_2$ to denote the number of type-1 useful epochs and type-2 useful super epochs.
	
	For type-1 useful epoch, according to Case 2 of Lemma \ref{lem:firstonline}, we know that the function value decreases at least $\frac{7\eta m\mathG^2}{256}$ with probability at least $1/5$.
	Using a union bound, we know that with probability $1-4N_1/5$, $N_1$ type-1 useful epochs will decrease the function value at least $\frac{7\eta m\mathG^2N_1}{1280}$. 
	Note that the function value can decrease at most $\Delta_0:= f(x_0)-f^*$ and also recall that the function value can only increase at most $\frac{\eta mC_3^2\sigma^2}{B}$ with high probability $1-3m\zeta'$ for any (wasted) epoch, where $C_3=O(\log\frac{d}{\zeta'})=\tdo(1)$ (see Lemma \ref{lem:firstonline}).
	By choosing $B=\tdo(\frac{\sigma^2}{\epsilon^2})$ and small enough $\zeta'$, $N_1$ type-1 useful epochs will decrease the function value at least $\frac{\eta m\mathG^2N_1}{200}$ with probability at least $1-\tdo(N_1m\zeta')$ by a union bound. We can let $\zeta'\leq \tdo(1/N_1m)$.
	So let $\frac{\eta m\mathG^2N_1}{200}\leq \Delta_0$, we get $N_1\leq \frac{200\Delta_0}{\eta m\mathG^2}$.
	
	For type-2 useful super epoch, first we know that the starting point of the super epoch $\tx:=x_{(s+1)m}$ has gradient norm $\n{\nabla f(\tx)}\leq \mathG/2$ and estimated gradient norm $\n{v_{(s+1)m}}\leq \mathG$. Now if $\lambda_{\min}(\nabla^2 f(\tx))\geq -\delta$, then $\tx$ is already a $(\epsilon,\delta)$-local minimum. Otherwise,
	$\n{v_{(s+1)m}}\leq \mathG$ and $\lambda_{\min}(\nabla^2 f(\tx)) \leq -\delta$, this is exactly our second situation (around saddle points).
	According to \eqref{eq:escapehigh1}, we know that the the function value decrease ($f(\tx)-f(x_T)$) is at least $\mathf=\frac{\delta^3}{2C_1'\rho^2}$ with  probability at least $1-7\mathT\zeta'\geq 1/2$ (let $\zeta'\leq 1/14\mathT$), where $C_1' =\tdo(1)$.
	Similar to type-1 useful epoch, we know $N_2\leq \frac{4C_1'\rho^2\Delta_0}{\delta^3}$ with probability at least $1-\tdo(N_2\mathT\zeta')$ by a union bound. We can let $\zeta'\leq \tdo(1/N_2\mathT)$.
	
	Now, we are ready to bound the number of SFO calls in Theorem \ref{thm:ssrgd-lm} (online) as follows:
	\begin{align}
	&N_1(\tdo(1)B+B+mb) +N_2(\tdo(1)B+\big\lceil\frac{\mathT}{m}\big\rceil B+\mathT b) \notag \\
	&\leq \tdo\Big(\frac{\Delta_0\sigma}{\eta \mathG^2\epsilon}+\frac{\rho^2\Delta_0}{\delta^3}(\frac{\sigma^2}{\epsilon^2}+\frac{\sigma}{\eta \delta\epsilon})\Big) \notag\\
	& \leq \tdo\Big(\frac{L\Delta_0\sigma}{\epsilon^3}
	+\frac{\rho^2\Delta_0\sigma^2}{\epsilon^2\delta^3}
	+ \frac{L\rho^2\Delta_0\sigma}{\epsilon\delta^4}\Big).  \label{eq:sfo-online}
	\end{align}
By a union bound of these types and set $\zeta=\tdo(N_1m+N_2\mathT)\zeta'$ (note that $\zeta'$ only appears in the log term $\log(\frac{1}{\zeta'})$, so it can be chosen as small as we want), we know that the SFO calls of SSRGD can be bounded by \eqref{eq:sfo-online} with probability $1-\zeta$.
This finishes the proof of Theorem~\ref{thm:ssrgd-lm} (the online case).
Now, the only remaining thing is to prove Lemma \ref{lem:localonline} and \ref{lem:smallstuckonline}.

\begin{proofof}{Lemma~\ref{lem:localonline}}
	First, we know that the variance bound \eqref{eq:highvar1} holds with probability $1-3\zeta'$. Then by a union bound, it holds with probability $1-3t\zeta'$ for all $0\leq j\leq t-1$.
	Then, according to \eqref{eq:bm21}, we know for any $\tau \leq t$ in some epoch $s$
	\begin{align}
	f(x_{\tau})&\leq f(x_{sm}) - \frac{\eta}{2}\sum_{j=sm+1}^{\tau}\ns{\nabla f(x_{j-1})}
	- \big(\frac{1}{2\eta}- \frac{L}{2}-\eta C'^2L^2\big)\sum_{j=sm+1}^{\tau}\ns{x_j-x_{j-1}} \notag\\
	&\qquad + \frac{(\tau-sm)\eta C_3^2\sigma^2}{B} \notag\\
	&\leq f(x_{sm}) - \big(\frac{1}{2\eta}- \frac{L}{2}-\eta C'^2L^2\big)\sum_{j=sm+1}^{\tau}\ns{x_j-x_{j-1}}+ \frac{(\tau-sm)\eta C^2\sigma^2}{B} \notag\\
	&\leq f(x_{sm}) - \frac{C'L}{4}\sum_{j=sm+1}^{\tau}\ns{x_j-x_{j-1}}+ \frac{(\tau-sm)\eta C_3^2\sigma^2}{B}, \label{eq:dist1}
	\end{align}
	where the last inequality holds since the step size $\eta\leq \frac{1}{(1+2C')L}$. Recall that $C'=O(\log\frac{dm}{\zeta'}\sqrt{\log\frac{d}{\zeta'}})=\tdo(1)$ and $C_3=O(\log\frac{d}{\zeta'})=\tdo(1)$.
	Now, we sum up \eqref{eq:dist1} for all epochs before iteration $t$,
	\begin{align*}
	f(x_{t})
	&\leq f(x_{0}) - \frac{C'L}{4}\sum_{j=1}^{t}\ns{x_j-x_{j-1}}+ \frac{t\eta C^2\sigma^2}{B}.
	\end{align*}
	Then, the proof is finished as
	\begin{align*}
	\n{x_t-x_0}\leq \sum_{j=1}^t\n{x_j-x_{j-1}}\leq \sqrt{t\sum_{j=1}^{t}\ns{x_j-x_{j-1}}}
	\leq \sqrt{\frac{4t(f(x_0)-f(x_t))}{C'L} +\frac{4t^2\eta C_3^2\sigma^2}{C'LB}}.
	\end{align*}
\end{proofof}

\begin{proofof}{Lemma~\ref{lem:smallstuckonline}}
	We prove this lemma by contradiction. Assume the contrary,
	\begin{align}\label{eq:distbound1}
	\forall t\leq T,~~ \n{x_t-x_0} \leq \frac{\delta}{C_1\rho} \mathrm{~~and~~} \n{x_t'-x_0'} \leq \frac{\delta}{C_1\rho},
	\end{align}
	where $T:=\frac{\log(\frac{8\delta\sqrt{d}}{\rho r\zeta'})}{\eta\gamma} \leq \mathT :=\frac{\log(\frac{8\delta\sqrt{d}}{\rho r\zeta'})}{\eta\delta}$ (note that $\gamma\geq \delta$ due to $-\gamma:=\lambda_{\min}(\nabla^2 f(\tx))\leq -\delta$).
	We will show that the distance between these two coupled sequences $w_t:=x_t-x_t'$ grows exponentially since they have a gap in the $e_1$ direction at the beginning, i.e., $w_0:=x_0-x_0'=r_0e_1$, where $r_0=\frac{\zeta' r}{\sqrt{d}}$ and $e_1$ denotes the smallest eigenvector direction of Hessian $\hess := \nabla^2 f(\tx)$.
	However, $\n{w_t}=\n{x_t-x_t'}\leq \n{x_t-x_0}+\n{x_0-\tx}+\n{x_t'-x_0'}+\n{x_0'-\tx}\leq 2r+2\frac{\delta}{C_1\rho}$ according to \eqref{eq:distbound1} and the perturbation radius $r$.
	It is not hard to see that if $\n{w_{t}}$ increases exponentially, this inequality cannot be true for reasonably large $t$, rendering a contradiction.
	
	In the following, we prove the exponential increase of $\n{w_t}$ by induction.
	First, recall the expression of $w_t$ in \eqref{eq:expwtandwt-1} and \eqref{eq:expw}:
	\begin{align}
		w_t&=(I-\eta \hess)w_{t-1}-\eta(\Delta_{t-1}w_{t-1}+y_{t-1}) \label{eq:expwtandwt-11} \\
		&=(I-\eta \hess)^{t}w_0-\eta\sum_{\tau=0}^{t-1}(I-\eta \hess)^{t-1-\tau}(\Delta_\tau w_\tau+y_\tau),
		\label{eq:expw1} 
	\end{align}
	where $\Delta_{\tau} :=\int_0^1(\nabla^2 f(x_\tau'+\theta(x_\tau-x_\tau'))-\hess)d\theta$
	and $y_\tau :=v_{\tau}-\nabla f(x_{\tau})-v_{\tau}'+\nabla f(x_{\tau}')$.	
	
	Again, to show the exponential increase of $\n{w_t}$, it is sufficient to show that the first term of \eqref{eq:expw1} dominates the second term. To this end, we show the following bound, which is almost the same as Lemma~\ref{lem:boundwtyt}, except that the succeed probability changes to $1-3T\zeta'$.
	
	\begin{lemma}\label{lem:boundwtyt2}
		Suppose	$w_0:=x_0-x_0'=r_0e_1$ where $r_0=\frac{\zeta' r}{\sqrt{d}}$ and $e_1$ is the eigenvector corresponding to the smallest eigenvalue of Hessian $\hess := \nabla^2 f(\tx)$.
		If \eqref{eq:distbound1} holds, then with probability $1-3T\zeta'$, the following bounds hold for all $t \leq T$:
		\begin{enumerate}
			\item $\frac{1}{2}(\base)^t r_0\leq\n{w_t}\leq\frac{3}{2}(\base)^t r_0$;
			\item $\n{y_t}\leq \frac{\gamma}{4C_2}(\base)^t r_0$.
		\end{enumerate}
		where $C_2 :=\log(\frac{8\delta\sqrt{d}}{\rho r\zeta'})$.
	\end{lemma}

	\begin{proofof}{Lemma~\ref{lem:boundwtyt2}}
	First, check the base case $t=0$, $\n{w_0}=\n{r_0 e_1}=r_0$ holds for Bound 1. However, the base case of Bound 2
	requires more work.
	Here, we use Bernstein inequality (Proposition \ref{prop:bernstein}) to show that
	$\n{y_0}= \n{v_{0}-\nabla f(x_0)-v_{0}'+\nabla f(x_{0}')}\leq \eta\gamma L r_0$.
	According to Line \ref{line:up1} of Algorithm~\ref{alg:ssrgd}, we know that $v_{0}= \frac{1}{B}\sum_{j\in I_B}\nabla f_j(x_{0})$ and $v_{0}'= \frac{1}{B}\sum_{j\in I_B}\nabla f_j(x_{0}')$ (recall that these two coupled sequence $\{x_t\}$ and $\{x_t'\}$ use the same choice of batches and minibatches (i.e., same $I_B$'s and $I_b$'s).
	Now, we have
	\begin{align}
	y_0&=v_{0}-\nabla f(x_0)-v_{0}'+\nabla f(x_{0}') \notag\\
	&=\frac{1}{B}\sum_{j\in I_B} \nabla f_j(x_{0})-\nabla f(x_0)
	-\frac{1}{B}\sum_{j\in I_B}\nabla f_j(x_{0}')+\nabla f(x_{0}')\notag\\
	&=\frac{1}{B}\sum_{j\in I_B}\Big(\nabla f_j(x_{0})-\nabla f_j(x_{0}')
	-(\nabla f(x_0) -\nabla f(x_{0}'))\Big). \label{eq:base}
	\end{align}
	We first bound the norm of each individual term of \eqref{eq:base}:
	\begin{align}
	\|\nabla f_j(x_{0})-\nabla f_j(x_{0}')
	-(\nabla f(x_0) -\nabla f(x_{0}'))\|\leq 2L\|x_{0}-x_{0}'\|=2L\n{w_0}=2Lr_0,\label{eq:b1x}
	\end{align}
	where the inequality holds due to the gradient Lipschitz Assumption \ref{asp:smoothgandh}.
	Then, consider the corresponding variance:
	\begin{align}
	&\E\Big[\sum_{j\in I_B}\ns{\nabla f_j(x_{0})-\nabla f_j(x_{0}')
		-(\nabla f(x_0) -\nabla f(x_{0}'))}\Big] \notag\\
	&\leq B\E_j[\ns{\nabla f_j(x_{0})-\nabla f_j(x_{0}')}] \leq BL^2\ns{x_{0}-x_{0}'} =BL^2\ns{w_0}=BL^2r_0^2, \label{eq:b2x}
	\end{align}
	where the first inequality uses the fact $\E[\ns{x-\E x}]\leq \E[\ns{x}]$, and the last inequality uses the gradient Lipschitz Assumption \ref{asp:smoothgandh}.
	According to \eqref{eq:b1x} and \eqref{eq:b2x}, we can bound $\n{y_0}$ by Bernstein inequality (Proposition \ref{prop:bernstein}) as
	\begin{align*}
	\pr\Big\{\big\|y_0\big\|\geq \frac{\alpha}{B} \Big\} &\leq (d+1) \exp\Big(\frac{-\alpha^2/2}{\sigma^2+R\alpha/3}\Big) \notag \\
	& = (d+1) \exp\Big(\frac{-\alpha^2/2}{BL^2r_0^2+ 2Lr_0\alpha/3}\Big)
	\notag \\
	& = \zeta',
	\end{align*}
	where the last equality holds by letting $\alpha=C_3L\sqrt{B}r_0$, where $C_3=O(\log\frac{d}{\zeta'})$. 
	By further choosing $B=\tdo(\frac{\sigma^2}{\mathG^2})$, we can see that the base case 
	\begin{align}
	\|y_0\| \leq  \frac{C_3Lr_0}{\sqrt{B}} \leq \frac{\gamma}{8C_2} r_0,
	\label{eq:basey}
	\end{align}
	holds with probability $1-\zeta'$.
	
	Now, we proceed to the induction step: 	Assuming Bound 1 and Bound 2 hold for all $\tau\leq t-1$, 
	we now prove they hold for $t$.
	For Bound 1, same arguments as in Lemma~\ref{lem:boundwtyt} can show that the second term of \eqref{eq:expw1} is dominated by half of the first term. We do not repeat the proof which are exactly the same.
	Note that the first term of \eqref{eq:expw1} is $\n{(I-\eta \hess)^{t}w_0}=(1+\eta\gamma)^t r_0$. Thus, we have the first bound:
	\begin{align}\label{eq:wtx}
	\frac{1}{2}(\base)^t r_0\leq\n{w_t}\leq\frac{3}{2}(\base)^t r_0
	\end{align}

	Now, we proceed to the second bound $\n{y_t}\leq \frac{\gamma}{4C_2}(\base)^t r_0$. 
	Define
	$$\beta:=\sqrt{8\sum_{k=sm+1}^{t} c_k^2\log\frac{d}{\zeta'}} = 
	\frac{C'}{\sqrt{b}}\sqrt{\sum_{k=sm+1}^{t}(L\|w_t-w_{t-1}\|+\rho D_t^x\n{w_t}+\rho D_{t-1}^x\n{w_{t-1}})^2},$$ 
	where $C'=O(C\sqrt{\log\frac{d}{\zeta'}})=O(\log\frac{d}{\zeta_k}\sqrt{\log\frac{d}{\zeta'}})=\tdo(1)$ and  $\zeta_k = \zeta'/m$.
	The same proof as in Lemma~\ref{lem:boundwtyt} show that
	\begin{align}
		\pr\Big\{\big\|y_{t}-y_{sm}\big\|\geq \beta \Big\} &\leq (d+1) \exp\Big(\frac{-\beta^2}{8\sum_{k=sm+1}^{t} c_k^2}\Big)+\zeta' = 2\zeta'.
		\label{eq:boundyt1}
	\end{align}

	Recall that $y_k:=v_{k}-\nabla f(x_{k})-v_{k}'+\nabla f(x_{k}')$ and at the beginning point of this epoch $y_{sm}=\n{v_{sm}-\nabla f(x_{sm})-v_{sm}'+\nabla f(x_{sm}')} \leq \frac{\gamma}{8C_2} r_0$ with probability $1-\zeta'$ (see \eqref{eq:basey}).
	Combining with \eqref{eq:boundyt1} and using a union bound, for any $t\in [sm+1,(s+1)m]$, we have that
	\begin{align}\label{eq:highvar2on}
	\n{y_{t}}\leq \beta +\n{y_{sm}}  
	\leq \frac{C'}{\sqrt{b}} \sqrt{\sum_{k=sm+1}^{t}(L\|w_t-w_{t-1}\|+\rho D_t^x\n{w_t}+\rho D_{t-1}^x\n{w_{t-1}})^2}+\frac{\gamma}{8C_2}r_0
	\end{align}
	holds with probability $1-3\zeta'$. Furthermore, by a union bound, we know that \eqref{eq:highvar2on} holds with probability at least $1-3T\zeta'$ for all $t\leq T$.
	
	Now, we bound the right-hand-side of \eqref{eq:highvar2on} to finish the proof. The proof will be the same as in Lemma \ref{lem:boundwtyt}.
	The last two terms inside the square root can be bounded as in \eqref{eq:50}:
	\begin{align}
	\rho D_t^x\n{w_t}+\rho D_{t-1}^x\n{w_{t-1}} \leq \frac{6\delta}{C_1}(\base)^t r_0, \label{eq:50x}
	\end{align}
	
	The first term in the square root can also be bounded in the same way as in \eqref{eq:splitspace}--\eqref{eq:firstterm}:
	\begin{align}
	L\|w_t-w_{t-1}\| 
	&\leq \Big(\frac{3}{2}+\frac{3 (1+\log t)}{C_1}+\frac{1+\log t}{4C_2}\Big) L\eta\gamma(\base)^{t-1} r_0. \label{eq:firsttermx}
	\end{align}

	By plugging \eqref{eq:50x} and \eqref{eq:firsttermx} into \eqref{eq:highvar2on}, we have
	
	\begin{align}
	\n{y_{t}}
	&\leq C'\left(\frac{6\delta}{C_1}(\base)^t r_0+\Big(\frac{3}{2}+\frac{3 (1+\log t)}{C_1}+\frac{1+\log t}{4C_2}\Big) L\eta\gamma(\base)^{t-1} r_0\right) + \frac{\gamma}{8C_2}r_0 \notag\\
	&\leq \Big(\frac{6C'}{C_1}+\big(\frac{3}{2}+\frac{3 (1+\log t)}{C_1}+\frac{1+\log t}{4C_2}\big)C' L\eta\Big)\gamma(\base)^t r_0 
		+ \frac{\gamma}{8C_2}r_0\notag\\
	&\leq \Big(\frac{1}{16C_2}+\frac{1}{16C_2} \Big)\gamma(\base)^t r_0 
	+ \frac{\gamma}{8C_2}r_0\notag\\
	&= \frac{1}{4C_2}\gamma(\base)^t r_0, 
	\end{align}
	where the second inequality holds due to $\delta\leq \gamma$ (recall $-\gamma:=\lambda_{\min}(\hess)=\lambda_{\min}(\nabla^2 f(\tx))\leq -\delta$),
	and the last inequality holds by letting
	$C_1 \geq 1 + 96C'C_2$ (recall that $C_2:=\log(\frac{8\delta\sqrt{d}}{\rho r\zeta'})$ defined in Lemma~\ref{lem:boundwtyt2}), 
	and $\eta \leq \frac{1}{30(1+\log t)C' L}$.
	Recall that \eqref{eq:highvar2on} holds with probability at least $1-3T\zeta'$ for all $t\leq T$.
	This finishes the proof of Lemma~\ref{lem:boundwtyt2}.
	\end{proofof}

	From the Lemma \ref{lem:boundwtyt2}, one can see that $\n{w_t} \geq \frac{1}{2}(\base)^t r_0=\frac{1}{2}(\base)^t\frac{\zeta' r}{\sqrt{d}}$. On the other hand, $\n{w_t}:=\n{x_t-x_t'}\leq \n{x_t-x_0}+\n{x_0-\tx}+\n{x_t'-x_0'}+\n{x_0'-\tx}\leq 2r+2\frac{\delta}{C_1\rho}\leq \frac{4\delta}{C_1\rho}$ according to \eqref{eq:distbound1} and the perturbation radius $r\leq \frac{\delta}{C_1 \rho}$.
	Hence, for any $t\geq T= \frac{1}{\eta\gamma}\log(\frac{8\delta\sqrt{d}}{\rho r\zeta'})$, we get a contradiction to \eqref{eq:distbound1}, i.e., $\n{w_t} \geq \frac{1}{2}(\base)^T\frac{\zeta' r}{\sqrt{d}}\geq \frac{4\delta}{\rho} \geq \frac{4\delta}{C_1\rho}$, where
	$C_1\geq 1+96C'\log(\frac{8\delta\sqrt{d}}{\rho r\zeta'}) \geq 1$ defined in Lemma~\ref{lem:smallstuckonline}.
	Also note that $T= \frac{1}{\eta\gamma}\log(\frac{8\delta\sqrt{d}}{\rho r\zeta'})\leq \mathT:=\frac{1}{\eta\delta}\log(\frac{8\delta\sqrt{d}}{\rho r\zeta'})$ due to $\delta\leq \gamma$. 
	This contradiction finishes the proof of Lemma~\ref{lem:smallstuckonline}.
\end{proofof}

\subsection{Proof of Theorem \ref{thm:ssrgd-lm-3rd} (Under third-order Lipschitzness assumption)}
\begin{proofof}{Theorem \ref{thm:ssrgd-lm-3rd}}
	The proof is similar to the proof for the online case of Theorem \ref{thm:ssrgd-lm} provided in Section \ref{app:ssrgd-lm-online}.
	Again, we distinguish two situations, the {\em large gradients} case, in which the function value decreases 
	significantly, and the {\em around saddle points} case.
	The proof for the first case (large gradients) is exactly same as in the first case in Section \ref{app:ssrgd-lm-online} (i.e., Lemma \ref{lem:firstonline}).
	
	The difference is in the second case (around saddle points).
	In previous Section \ref{app:ssrgd-lm-online}, we add a random perturbation at the starting point of the super epoch. 
	Concretely, we show that the function value decreases a lot in this super epoch with high probability (see \eqref{eq:escapehigh1}), i.e.,
	\begin{align}\label{eq:func-superepoch}
		\exists T\leq \mathT,~~ f(\tx)-f(x_T) \geq \mathf=\frac{\delta^3}{2C_1'\rho^2}
	\end{align}
	holds with probability at least $1-7\mathT\zeta'$, where $C_1'=\tdo(1)$. Recall that the super epoch length $\mathT:=\frac{1}{\eta\delta}\log(\frac{8\delta\sqrt{d}}{\rho r\zeta'})=\tdo(\frac{1}{\eta\delta})$ (see Lemma \ref{lem:smallstuckonline}) and $\tx$ is the starting point of this super epoch.
	However, for Theorem \ref{thm:ssrgd-lm-3rd} which further assumes the $L_3$-Lipschitz of third-order derivative (i.e., Assumption \ref{asp:smooth-3rd}), one can show that the function value decreases by a larger amount (\emph{improving a factor of $\delta$}), i.e.,  $\frac{3\blue{\delta^2}}{8L_3}$ in \eqref{eq:func-superepoch-3rd} instead of $\frac{\blue{\delta^3}}{2C_1'\rho^2}$ in \eqref{eq:func-superepoch}. 
	Finally we can see that the result of Theorem \ref{thm:ssrgd-lm-3rd} indeed improves the previous online case of Theorem \ref{thm:ssrgd-lm} by a factor of $\delta$.
	Now we formalize the proof of Theorem \ref{thm:ssrgd-lm-3rd} in this second case (around saddle points).
	Here we directly reuse the function value decrease lemma provided in \citet{yu2017third}.
	Note that here we can remove the expectation in Lemma 4.6 of \citet{yu2017third}
	by choosing $y=\arg\min_{y\in\{y_{-},y_{+}\}}f(y)$.

	\begin{lemma}[Lemma 4.6 in \citep{yu2017third}]\label{lem:3rd}
		Suppose that Assumptions \ref{asp:smoothgandh}, \ref{asp:var2} and \ref{asp:smooth-3rd} hold.  If the start point $\tx$ satisfies $\lambda_{\min}(\nabla^2 f(\tx)) \leq -\delta$. Then one can apply a negative curvature search step for finding a direction to decrease the function value. In particular, Neon2$^\text{online}$ \citep{allen2018neon2} can return a point $y$ such that 
		\begin{align}\label{eq:func-superepoch-3rd}
			f(\tx)-f(y) \geq \frac{3\delta^2}{8L_3}
		\end{align}
		holds with probability $1-\zeta'$ and the total number of stochastic gradient computations is at most $T=O(\frac{L^2}{\delta^2}\log^2\frac{d}{\zeta'})=\tdo(\frac{L^2}{\delta^2})$.
	\end{lemma}
	
	Now, we are ready to combine these two situations (large gradients and around saddle points) to prove Theorem \ref{thm:ssrgd-lm-3rd}.
	The arguments are similar to that in Section \ref{app:ssrgd-lm-online}. The only difference is that here we replace super epoch step by the negative curvature search step (i.e., replace \eqref{eq:func-superepoch} by \eqref{eq:func-superepoch-3rd}) in the around saddle points situation.
	Concretely, i) for large gradients situation, $N_1$ type-1 useful epochs will decrease the function value at least $\frac{\eta m\mathG^2N_1}{200}$ with probability at least $1-\tdo(N_1m\zeta')$ by a union bound. We can let $\zeta'\leq \tdo(1/N_1m)$. So let $\frac{\eta m\mathG^2N_1}{200}\leq \Delta_0$, we get $N_1\leq \frac{200\Delta_0}{\eta m\mathG^2}$.
	ii) for around saddle points situation,
	according to \eqref{eq:func-superepoch-3rd}, we know that the the function value decrease ($f(\tx)-f(y)$) is  $\frac{3\delta^2}{8L_3}$ with probability at least $1-\zeta'$.
	Similar to the large gradients situation, we know $N_2\leq \frac{16L_3\Delta_0}{3\delta^2}$ with probability at least $1-\tdo(N_2\zeta')$ by a union bound. We can let $\zeta'\leq \tdo(1/N_2)$.
	Now, we bound the number of SFO calls in Theorem \ref{thm:ssrgd-lm-3rd} (online case under third-order Lipschitz) as follows:
	\begin{align}
		N_1(\tdo(1)B+B+mb) +N_2(\tdo(1)B+T) 
		 \leq \tdo\Big(\frac{L\Delta_0\sigma}{\epsilon^3}
		+\frac{L_3\Delta_0\sigma^2}{\epsilon^2\delta^2}
		+ \frac{L_3L^2\Delta_0}{\delta^4}\Big).  \label{eq:sfo-online-3rd}
	\end{align}
	By a union bound of these types and set $\zeta=\tdo(N_1m+N_2)\zeta'$ (note that $\zeta'$ only appears in the log term $\log(\frac{1}{\zeta'})$, so it can be chosen as small as we want), we know that the SFO calls can be bounded by \eqref{eq:sfo-online-3rd} with probability $1-\zeta$.
\end{proofof}

\vskip 0.2in

\bibliography{nc}

\begin{thebibliography}{71}
\providecommand{\natexlab}[1]{#1}
\providecommand{\url}[1]{\texttt{#1}}
\expandafter\ifx\csname urlstyle\endcsname\relax
  \providecommand{\doi}[1]{doi: #1}\else
  \providecommand{\doi}{doi: \begingroup \urlstyle{rm}\Url}\fi

\bibitem[Agarwal et~al.(2016)Agarwal, Allen-Zhu, Bullins, Hazan, and
  Ma]{agarwal2016finding}
N.~Agarwal, Z.~Allen-Zhu, B.~Bullins, E.~Hazan, and T.~Ma.
\newblock Finding approximate local minima for nonconvex optimization in linear
  time.
\newblock \emph{arXiv preprint arXiv:1611.01146}, 2016.

\bibitem[Allen-Zhu(2017)]{allen2017katyusha}
Z.~Allen-Zhu.
\newblock Katyusha: the first direct acceleration of stochastic gradient
  methods.
\newblock In \emph{Symposium on Theory of Computing}, pages 1200--1205. ACM,
  2017.

\bibitem[Allen-Zhu(2018)]{allen2018natasha}
Z.~Allen-Zhu.
\newblock Natasha 2: Faster non-convex optimization than {SGD}.
\newblock In \emph{Advances in {N}eural {I}nformation {P}rocessing {S}ystems},
  pages 2680--2691, 2018.

\bibitem[Allen-Zhu and Li(2018)]{allen2018neon2}
Z.~Allen-Zhu and Y.~Li.
\newblock Neon2: Finding local minima via first-order oracles.
\newblock In \emph{Advances in Neural Information Processing Systems}, pages
  3720--3730, 2018.

\bibitem[Anandkumar and Ge(2016)]{anandkumar2016efficient}
A.~Anandkumar and R.~Ge.
\newblock Efficient approaches for escaping higher order saddle points in
  non-convex optimization.
\newblock In \emph{Conference on learning theory}, pages 81--102, 2016.

\bibitem[Anitescu(2000)]{anitescu2000degenerate}
M.~Anitescu.
\newblock Degenerate nonlinear programming with a quadratic growth condition.
\newblock \emph{SIAM Journal on Optimization}, 10\penalty0 (4):\penalty0
  1116--1135, 2000.

\bibitem[Bhojanapalli et~al.(2016)Bhojanapalli, Neyshabur, and
  Srebro]{bhojanapalli2016global}
S.~Bhojanapalli, B.~Neyshabur, and N.~Srebro.
\newblock Global optimality of local search for low rank matrix recovery.
\newblock In \emph{Advances in Neural Information Processing Systems}, pages
  3873--3881, 2016.

\bibitem[Carmon et~al.(2016)Carmon, Duchi, Hinder, and
  Sidford]{carmon2016accelerated}
Y.~Carmon, J.~C. Duchi, O.~Hinder, and A.~Sidford.
\newblock Accelerated methods for non-convex optimization.
\newblock \emph{arXiv preprint arXiv:1611.00756}, 2016.

\bibitem[Carmon et~al.(2017)Carmon, Duchi, Hinder, and
  Sidford]{carmon2017convex}
Y.~Carmon, J.~C. Duchi, O.~Hinder, and A.~Sidford.
\newblock “convex until proven guilty”: Dimension-free acceleration of
  gradient descent on non-convex functions.
\newblock In \emph{International Conference on Machine Learning}, pages
  654--663. PMLR, 2017.

\bibitem[Chung and Lu(2006)]{chung2006concentration}
F.~Chung and L.~Lu.
\newblock Concentration inequalities and martingale inequalities: a survey.
\newblock \emph{Internet Mathematics}, 3\penalty0 (1):\penalty0 79--127, 2006.

\bibitem[Daneshmand et~al.(2018)Daneshmand, Kohler, Lucchi, and
  Hofmann]{daneshmand2018escaping}
H.~Daneshmand, J.~Kohler, A.~Lucchi, and T.~Hofmann.
\newblock Escaping saddles with stochastic gradients.
\newblock In \emph{International Conference on Machine Learning}, pages
  1155--1164, 2018.

\bibitem[Defazio et~al.(2014)Defazio, Bach, and
  Lacoste-Julien]{defazio2014saga}
A.~Defazio, F.~Bach, and S.~Lacoste-Julien.
\newblock {SAGA}: A fast incremental gradient method with support for
  non-strongly convex composite objectives.
\newblock In \emph{Advances in Neural Information Processing Systems}, pages
  1646--1654, 2014.

\bibitem[Du et~al.(2017)Du, Jin, Lee, Jordan, Singh, and
  Poczos]{du2017gradient}
S.~S. Du, C.~Jin, J.~D. Lee, M.~I. Jordan, A.~Singh, and B.~Poczos.
\newblock Gradient descent can take exponential time to escape saddle points.
\newblock In \emph{Advances in Neural Information Processing Systems}, pages
  1067--1077, 2017.

\bibitem[Fang et~al.(2018)Fang, Li, Lin, and Zhang]{fang2018spider}
C.~Fang, C.~J. Li, Z.~Lin, and T.~Zhang.
\newblock {SPIDER}: Near-optimal non-convex optimization via stochastic
  path-integrated differential estimator.
\newblock In \emph{Advances in Neural Information Processing Systems}, pages
  687--697, 2018.

\bibitem[Fang et~al.(2019)Fang, Lin, and Zhang]{fang2019sharp}
C.~Fang, Z.~Lin, and T.~Zhang.
\newblock Sharp analysis for nonconvex sgd escaping from saddle points.
\newblock In \emph{Conference on Learning Theory}, pages 1192--1234, 2019.

\bibitem[Fatkhullin et~al.(2021)Fatkhullin, Sokolov, Gorbunov, Li, and
  Richt{\'a}rik]{fatkhullin2021ef21}
I.~Fatkhullin, I.~Sokolov, E.~Gorbunov, Z.~Li, and P.~Richt{\'a}rik.
\newblock {EF21} with bells \& whistles: Practical algorithmic extensions of
  modern error feedback.
\newblock \emph{arXiv preprint arXiv:2110.03294}, 2021.

\bibitem[Ge et~al.(2015)Ge, Huang, Jin, and Yuan]{ge2015escaping}
R.~Ge, F.~Huang, C.~Jin, and Y.~Yuan.
\newblock Escaping from saddle points --- online stochastic gradient for tensor
  decomposition.
\newblock In \emph{Conference on Learning Theory}, pages 797--842, 2015.

\bibitem[Ge et~al.(2016)Ge, Lee, and Ma]{ge2016matrix}
R.~Ge, J.~D. Lee, and T.~Ma.
\newblock Matrix completion has no spurious local minimum.
\newblock In \emph{Advances in Neural Information Processing Systems}, pages
  2973--2981, 2016.

\bibitem[Ge et~al.(2017)Ge, Lee, and Ma]{ge2017learning}
R.~Ge, J.~D. Lee, and T.~Ma.
\newblock Learning one-hidden-layer neural networks with landscape design.
\newblock \emph{arXiv preprint arXiv:1711.00501}, 2017.

\bibitem[Ge et~al.(2019)Ge, Li, Wang, and Wang]{ge2019stable}
R.~Ge, Z.~Li, W.~Wang, and X.~Wang.
\newblock Stabilized {SVRG}: Simple variance reduction for nonconvex
  optimization.
\newblock In \emph{Conference on learning theory}, pages 1394--1448. PMLR,
  2019.

\bibitem[Ghadimi and Lan(2013)]{ghadimi2013stochastic}
S.~Ghadimi and G.~Lan.
\newblock Stochastic first-and zeroth-order methods for nonconvex stochastic
  programming.
\newblock \emph{SIAM Journal on Optimization}, 23\penalty0 (4):\penalty0
  2341--2368, 2013.

\bibitem[Ghadimi et~al.(2016)Ghadimi, Lan, and Zhang]{ghadimi2016mini}
S.~Ghadimi, G.~Lan, and H.~Zhang.
\newblock Mini-batch stochastic approximation methods for nonconvex stochastic
  composite optimization.
\newblock \emph{Mathematical Programming}, 155\penalty0 (1-2):\penalty0
  267--305, 2016.

\bibitem[Gorbunov et~al.(2021)Gorbunov, Burlachenko, Li, and
  Richt{\'a}rik]{gorbunov2021marina}
E.~Gorbunov, K.~P. Burlachenko, Z.~Li, and P.~Richt{\'a}rik.
\newblock {MARINA}: Faster non-convex distributed learning with compression.
\newblock In \emph{International Conference on Machine Learning}, pages
  3788--3798. PMLR, 2021.

\bibitem[Hoeffding(1963)]{hoeffding1963probability}
W.~Hoeffding.
\newblock Probability inequalities for sums of bounded random variables.
\newblock \emph{Journal of the American Statistical Association}, 58\penalty0
  (301):\penalty0 13--30, 1963.

\bibitem[Jin et~al.(2017)Jin, Ge, Netrapalli, Kakade, and
  Jordan]{jin2017escape}
C.~Jin, R.~Ge, P.~Netrapalli, S.~M. Kakade, and M.~I. Jordan.
\newblock How to escape saddle points efficiently.
\newblock In \emph{International Conference on Machine Learning}, pages
  1724--1732, 2017.

\bibitem[Jin et~al.(2018)Jin, Netrapalli, and Jordan]{jin2018accelerated}
C.~Jin, P.~Netrapalli, and M.~I. Jordan.
\newblock Accelerated gradient descent escapes saddle points faster than
  gradient descent.
\newblock In \emph{Conference On Learning Theory}, pages 1042--1085. PMLR,
  2018.

\bibitem[Jin et~al.(2019)Jin, Netrapalli, Ge, Kakade, and
  Jordan]{jin2019stochastic}
C.~Jin, P.~Netrapalli, R.~Ge, S.~M. Kakade, and M.~I. Jordan.
\newblock Stochastic gradient descent escapes saddle points efficiently.
\newblock \emph{arXiv preprint arXiv:1902.04811}, 2019.

\bibitem[Johnson and Zhang(2013)]{johnson2013accelerating}
R.~Johnson and T.~Zhang.
\newblock Accelerating stochastic gradient descent using predictive variance
  reduction.
\newblock In \emph{Advances in Neural Information Processing Systems}, pages
  315--323, 2013.

\bibitem[Karimi et~al.(2016)Karimi, Nutini, and Schmidt]{karimi2016linear}
H.~Karimi, J.~Nutini, and M.~Schmidt.
\newblock Linear convergence of gradient and proximal-gradient methods under
  the polyak-{\l}ojasiewicz condition.
\newblock In \emph{Joint European Conference on Machine Learning and Knowledge
  Discovery in Databases}, pages 795--811. Springer, 2016.

\bibitem[Karimireddy et~al.(2020)Karimireddy, Kale, Mohri, Reddi, Stich, and
  Suresh]{karimireddy2020scaffold}
S.~P. Karimireddy, S.~Kale, M.~Mohri, S.~Reddi, S.~Stich, and A.~T. Suresh.
\newblock {SCAFFOLD}: Stochastic controlled averaging for federated learning.
\newblock In \emph{International Conference on Machine Learning}, pages
  5132--5143. PMLR, 2020.

\bibitem[Kovalev et~al.(2019)Kovalev, Horv{\'a}th, and
  Richt{\'a}rik]{kovalev2020don}
D.~Kovalev, S.~Horv{\'a}th, and P.~Richt{\'a}rik.
\newblock Don’t jump through hoops and remove those loops: {SVRG} and
  {Katyusha} are better without the outer loop.
\newblock \emph{arXiv preprint arXiv:1901.08689}, 2019.

\bibitem[Lan and Zhou(2015)]{lan2015optimal}
G.~Lan and Y.~Zhou.
\newblock An optimal randomized incremental gradient method.
\newblock \emph{arXiv preprint arXiv:1507.02000}, 2015.

\bibitem[Lan and Zhou(2018)]{lan2018random}
G.~Lan and Y.~Zhou.
\newblock Random gradient extrapolation for distributed and stochastic
  optimization.
\newblock \emph{SIAM Journal on Optimization}, 28\penalty0 (4):\penalty0
  2753--2782, 2018.

\bibitem[Lan et~al.(2019)Lan, Li, and Zhou]{zhize2019unified}
G.~Lan, Z.~Li, and Y.~Zhou.
\newblock A unified variance-reduced accelerated gradient method for convex
  optimization.
\newblock In \emph{Advances in Neural Information Processing Systems}, pages
  10462--10472, 2019.

\bibitem[Lei and Jordan(2017)]{lei2017less}
L.~Lei and M.~Jordan.
\newblock Less than a single pass: Stochastically controlled stochastic
  gradient.
\newblock In \emph{Artificial Intelligence and Statistics}, pages 148--156,
  2017.

\bibitem[Lei et~al.(2017)Lei, Ju, Chen, and Jordan]{lei2017non}
L.~Lei, C.~Ju, J.~Chen, and M.~I. Jordan.
\newblock Non-convex finite-sum optimization via {SCSG} methods.
\newblock In \emph{Advances in Neural Information Processing Systems}, pages
  2345--2355, 2017.

\bibitem[Li et~al.(2022{\natexlab{a}})Li, Li, and Chi]{li2022destress}
B.~Li, Z.~Li, and Y.~Chi.
\newblock {DESTRESS}: Computation-optimal and communication-efficient
  decentralized nonconvex finite-sum optimization.
\newblock \emph{SIAM Journal on Mathematics of Data Science}, 4\penalty0
  (3):\penalty0 1031--1051, 2022{\natexlab{a}}.

\bibitem[Li(2019)]{li2019ssrgd}
Z.~Li.
\newblock {SSRGD}: Simple stochastic recursive gradient descent for escaping
  saddle points.
\newblock In \emph{Advances in Neural Information Processing Systems}, pages
  1523--1533, 2019.

\bibitem[Li(2021)]{li2021anita}
Z.~Li.
\newblock {ANITA}: An optimal loopless accelerated variance-reduced gradient
  method.
\newblock \emph{arXiv preprint arXiv:2103.11333}, 2021.

\bibitem[Li and Li(2018)]{li2018simple}
Z.~Li and J.~Li.
\newblock A simple proximal stochastic gradient method for nonsmooth nonconvex
  optimization.
\newblock In \emph{Advances in Neural Information Processing Systems}, pages
  5569--5579, 2018.

\bibitem[Li and Richt{\'a}rik(2020)]{li2020unified}
Z.~Li and P.~Richt{\'a}rik.
\newblock A unified analysis of stochastic gradient methods for nonconvex
  federated optimization.
\newblock \emph{arXiv preprint arXiv:2006.07013}, 2020.

\bibitem[Li and Richt{\'a}rik(2021{\natexlab{a}})]{li2021canita}
Z.~Li and P.~Richt{\'a}rik.
\newblock {CANITA}: Faster rates for distributed convex optimization with
  communication compression.
\newblock In \emph{Advances in Neural Information Processing Systems}, pages
  13770--13781, 2021{\natexlab{a}}.

\bibitem[Li and Richt{\'a}rik(2021{\natexlab{b}})]{li2021zerosarah}
Z.~Li and P.~Richt{\'a}rik.
\newblock {ZeroSARAH}: Efficient nonconvex finite-sum optimization with zero
  full gradient computation.
\newblock \emph{arXiv preprint arXiv:2103.01447}, 2021{\natexlab{b}}.

\bibitem[Li et~al.(2020)Li, Kovalev, Qian, and
  Richt{\'a}rik]{li2020acceleration}
Z.~Li, D.~Kovalev, X.~Qian, and P.~Richt{\'a}rik.
\newblock Acceleration for compressed gradient descent in distributed and
  federated optimization.
\newblock In \emph{International Conference on Machine Learning}, pages
  5895--5904. PMLR, 2020.

\bibitem[Li et~al.(2021)Li, Bao, Zhang, and Richt{\'a}rik]{li2021page}
Z.~Li, H.~Bao, X.~Zhang, and P.~Richt{\'a}rik.
\newblock {PAGE}: A simple and optimal probabilistic gradient estimator for
  nonconvex optimization.
\newblock In \emph{International Conference on Machine Learning}, pages
  6286--6295. PMLR, 2021.

\bibitem[Li et~al.(2022{\natexlab{b}})Li, Zhao, Li, and Chi]{li2022soteriafl}
Z.~Li, H.~Zhao, B.~Li, and Y.~Chi.
\newblock {SoteriaFL}: A unified framework for private federated learning with
  communication compression.
\newblock \emph{arXiv preprint arXiv:2206.09888}, 2022{\natexlab{b}}.

\bibitem[Lin et~al.(2015)Lin, Mairal, and Harchaoui]{lin2015universal}
H.~Lin, J.~Mairal, and Z.~Harchaoui.
\newblock A universal catalyst for first-order optimization.
\newblock In \emph{Advances in Neural Information Processing Systems}, pages
  3384--3392, 2015.

\bibitem[Luo and Tseng(1993)]{luo1993error}
Z.-Q. Luo and P.~Tseng.
\newblock Error bounds and convergence analysis of feasible descent methods: a
  general approach.
\newblock \emph{Annals of Operations Research}, 46\penalty0 (1):\penalty0
  157--178, 1993.

\bibitem[Necoara et~al.(2015)Necoara, Nesterov, and Glineur]{necoara2015linear}
I.~Necoara, Y.~Nesterov, and F.~Glineur.
\newblock Linear convergence of first order methods for non-strongly convex
  optimization.
\newblock \emph{arXiv preprint arXiv:1504.06298}, 2015.

\bibitem[Nesterov(2004)]{nesterov2014introductory}
Y.~Nesterov.
\newblock \emph{Introductory Lectures on Convex Optimization: A Basic Course}.
\newblock Kluwer, 2004.

\bibitem[Nguyen et~al.(2017)Nguyen, Liu, Scheinberg, and
  Tak{\'a}{\v{c}}]{nguyen2017sarah}
L.~M. Nguyen, J.~Liu, K.~Scheinberg, and M.~Tak{\'a}{\v{c}}.
\newblock {SARAH}: A novel method for machine learning problems using
  stochastic recursive gradient.
\newblock In \emph{International Conference on Machine Learning}, pages
  2613--2621, 2017.

\bibitem[Pham et~al.(2019)Pham, Nguyen, Phan, and Tran-Dinh]{pham2019proxsarah}
N.~H. Pham, L.~M. Nguyen, D.~T. Phan, and Q.~Tran-Dinh.
\newblock {ProxSARAH}: An efficient algorithmic framework for stochastic
  composite nonconvex optimization.
\newblock \emph{arXiv preprint arXiv:1902.05679}, 2019.

\bibitem[Polyak(1963)]{polyak1963gradient}
B.~T. Polyak.
\newblock Gradient methods for minimizing functionals.
\newblock \emph{Zhurnal Vychislitel'noi Matematiki i Matematicheskoi Fiziki},
  3\penalty0 (4):\penalty0 643--653, 1963.

\bibitem[Reddi et~al.(2016{\natexlab{a}})Reddi, Hefny, Sra, P{\'o}czos, and
  Smola]{reddi2016stochastic}
S.~J. Reddi, A.~Hefny, S.~Sra, B.~P{\'o}czos, and A.~Smola.
\newblock Stochastic variance reduction for nonconvex optimization.
\newblock In \emph{International Conference on Machine Learning}, pages
  314--323, 2016{\natexlab{a}}.

\bibitem[Reddi et~al.(2016{\natexlab{b}})Reddi, Sra, P{\'o}czos, and
  Smola]{reddi2016proximal}
S.~J. Reddi, S.~Sra, B.~P{\'o}czos, and A.~J. Smola.
\newblock Proximal stochastic methods for nonsmooth nonconvex finite-sum
  optimization.
\newblock In \emph{Advances in Neural Information Processing Systems}, pages
  1145--1153, 2016{\natexlab{b}}.

\bibitem[Richt{\'a}rik et~al.(2021)Richt{\'a}rik, Sokolov, and
  Fatkhullin]{richtarik2021ef21}
P.~Richt{\'a}rik, I.~Sokolov, and I.~Fatkhullin.
\newblock {EF21}: A new, simpler, theoretically better, and practically faster
  error feedback.
\newblock In \emph{Advances in Neural Information Processing Systems}, pages
  4384--4396, 2021.

\bibitem[Richt{\'a}rik et~al.(2022)Richt{\'a}rik, Sokolov, Gasanov, Fatkhullin,
  Li, and Gorbunov]{richtarik20223pc}
P.~Richt{\'a}rik, I.~Sokolov, E.~Gasanov, I.~Fatkhullin, Z.~Li, and
  E.~Gorbunov.
\newblock {3PC}: Three point compressors for communication-efficient
  distributed training and a better theory for lazy aggregation.
\newblock In \emph{International Conference on Machine Learning}, pages
  18596--18648. PMLR, 2022.

\bibitem[Schmidt et~al.(2013)Schmidt, Roux, and Bach]{schmidt2013minimizing}
M.~Schmidt, N.~L. Roux, and F.~Bach.
\newblock Minimizing finite sums with the stochastic average gradient.
\newblock \emph{arXiv preprint arXiv:1309.2388}, 2013.

\bibitem[Tao and Vu(2015)]{tao2015random}
T.~Tao and V.~Vu.
\newblock Random matrices: Universality of local spectral statistics of
  non-hermitian matrices.
\newblock \emph{The Annals of Probability}, 43\penalty0 (2):\penalty0 782--874,
  2015.

\bibitem[Tropp(2011)]{tropp2011user}
J.~A. Tropp.
\newblock User-friendly tail bounds for matrix martingales.
\newblock Technical report, CALIFORNIA INST OF TECH PASADENA, 2011.

\bibitem[Tropp(2012)]{tropp2012user}
J.~A. Tropp.
\newblock User-friendly tail bounds for sums of random matrices.
\newblock \emph{Foundations of computational mathematics}, 12\penalty0
  (4):\penalty0 389--434, 2012.

\bibitem[Wang et~al.(2019)Wang, Ji, Zhou, Liang, and
  Tarokh]{wang2019spiderboost}
Z.~Wang, K.~Ji, Y.~Zhou, Y.~Liang, and V.~Tarokh.
\newblock {SpiderBoost} and momentum: Faster variance reduction algorithms.
\newblock In \emph{Advances in Neural Information Processing Systems}, pages
  2406--2416, 2019.

\bibitem[Woodworth and Srebro(2016)]{woodworth2016tight}
B.~E. Woodworth and N.~Srebro.
\newblock Tight complexity bounds for optimizing composite objectives.
\newblock In \emph{Advances in Neural Information Processing Systems}, pages
  3639--3647, 2016.

\bibitem[Xiao and Zhang(2014)]{xiao2014proximal}
L.~Xiao and T.~Zhang.
\newblock A proximal stochastic gradient method with progressive variance
  reduction.
\newblock \emph{SIAM Journal on Optimization}, 24\penalty0 (4):\penalty0
  2057--2075, 2014.

\bibitem[Xu et~al.(2018)Xu, Rong, and Yang]{xu2018first}
Y.~Xu, J.~Rong, and T.~Yang.
\newblock First-order stochastic algorithms for escaping from saddle points in
  almost linear time.
\newblock In \emph{Advances in Neural Information Processing Systems}, pages
  5535--5545, 2018.

\bibitem[Yu et~al.(2017)Yu, Xu, and Gu]{yu2017third}
Y.~Yu, P.~Xu, and Q.~Gu.
\newblock Third-order smoothness helps: Even faster stochastic optimization
  algorithms for finding local minima.
\newblock \emph{arXiv preprint arXiv:1712.06585}, 2017.

\bibitem[Zhao et~al.(2021{\natexlab{a}})Zhao, Burlachenko, Li, and
  Richt{\'a}rik]{zhao2021faster}
H.~Zhao, K.~Burlachenko, Z.~Li, and P.~Richt{\'a}rik.
\newblock Faster rates for compressed federated learning with client-variance
  reduction.
\newblock \emph{arXiv preprint arXiv:2112.13097}, 2021{\natexlab{a}}.

\bibitem[Zhao et~al.(2021{\natexlab{b}})Zhao, Li, and
  Richt{\'a}rik]{zhao2021fedpage}
H.~Zhao, Z.~Li, and P.~Richt{\'a}rik.
\newblock {FedPAGE}: A fast local stochastic gradient method for
  communication-efficient federated learning.
\newblock \emph{arXiv preprint arXiv:2108.04755}, 2021{\natexlab{b}}.

\bibitem[Zhao et~al.(2022)Zhao, Li, Li, Richt{\'a}rik, and Chi]{zhao2022beer}
H.~Zhao, B.~Li, Z.~Li, P.~Richt{\'a}rik, and Y.~Chi.
\newblock {BEER}: Fast {$O(1/T)$} rate for decentralized nonconvex optimization
  with communication compression.
\newblock \emph{arXiv preprint arXiv:2201.13320}, 2022.

\bibitem[Zhou et~al.(2018{\natexlab{a}})Zhou, Xu, and Gu]{zhou2018finding}
D.~Zhou, P.~Xu, and Q.~Gu.
\newblock Finding local minima via stochastic nested variance reduction.
\newblock \emph{arXiv preprint arXiv:1806.08782}, 2018{\natexlab{a}}.

\bibitem[Zhou et~al.(2018{\natexlab{b}})Zhou, Xu, and Gu]{zhou2018stochastic}
D.~Zhou, P.~Xu, and Q.~Gu.
\newblock Stochastic nested variance reduction for nonconvex optimization.
\newblock In \emph{Advances in Neural Information Processing Systems}, pages
  3921--3932, 2018{\natexlab{b}}.

\end{thebibliography}

\end{document}